\documentclass[fleqn,10pt]{wlscirep}
\usepackage{cite}

\usepackage{caption}
\usepackage{bm}
\usepackage{subcaption}
\usepackage[font=small,labelfont=bf]{caption}
\definecolor{darkyellow2}{RGB}{245, 239, 231}
\definecolor{salmon2}{RGB}{150, 112, 126}
\definecolor{Gray}{gray}{0.9}
\definecolor{darkyellow}{RGB}{219, 181, 181} 
\definecolor{salmon}{RGB}{237, 230, 200}
\definecolor{darksalmon}{RGB}{201, 193, 160}
\usepackage{multirow}
\definecolor{darkgreen}{RGB}{180, 230, 198}
\definecolor{lightgreen}{RGB}{209, 233, 201}

\definecolor{lightred}{RGB}{255,211,222}
\definecolor{darkred}{RGB}{252,150,167}
 \definecolor{LightBlue}{rgb}{0.8,0.89,1} %%% Ali 1
\definecolor{DarkBlue}{rgb}{0.57,0.71,0.82} %%% Ali 3
\definecolor{MagLight}{rgb}{1, 0.89, 0.8}
\usepackage{enumitem}
  \makeatletter
\newcommand\notsotiny{\@setfontsize\notsotiny\@vipt\@viipt}
\makeatother
\usepackage[most]{tcolorbox}
\usepackage[font=footnotesize]{caption}
\newcommand{\semihuge}{\fontsize{18pt}{24pt}\selectfont}
\usepackage{bm}
\usepackage{amsmath}
\usepackage{amsthm}
\usepackage{ amssymb }
\usepackage{mathrsfs}

\usepackage{amsmath,amssymb,amsthm,mathtools,mathrsfs}

\allowdisplaybreaks
\newtheorem{theorem}{Theorem}
\newtheorem{lemma}{Lemma}

\newtheorem{definition}{Definition}

\newtheorem{corollary}{Corollary}

\newtheorem{assumption}{Assumption}

\usepackage[]{algorithm2e}

\addto\captionsenglish{}
\usepackage{bbm}

\usepackage{multicol}
\usepackage{mathtools}

\usepackage{lipsum,graphicx,subcaption}
\makeatletter
\newcommand{\vast}{\bBigg@{4}}

\newcommand{\Vast}{\bBigg@{5}}
\begin{abstract}
 We provide our perspective on $\mathbb{X}$-Learning ($\mathbb{X}$L), a novel distributed learning architecture that generalizes and extends the concept of \textit{decentralization}. 
  {Our goal is to present a vision for $\mathbb{X}$L, introducing its unexplored design considerations and  degrees of freedom.} 
  To this end, we shed light on the intuitive yet non-trivial connections between $\mathbb{X}$L, graph theory, and Markov chains. We also present a series of open research directions to stimulate further research.
\end{abstract}

\begin{document}

\title{\semihuge From Federated Learning to $\mathbb{X}$-Learning: Breaking the Barriers of Decentrality Through Random Walks}

\author[1,+]{Allan Salihovic}
\author[1,+]{Payam Abdisarabshali}
\author[1]{Michael Langberg}
\author[1,*]{\\ Seyyedali Hosseinalipour}
\affil[1]{University at Buffalo--SUNY, Department of Electrical Engineering, Buffalo, NY, USA}
\affil[+]{The authors contributed equally to this work.}
\affil[*]{Corresponding Author: S. Hosseinalipour (email: alipour@buffalo.edu)}
% \affil[2]{New Jersey Institute of Technology (NJIT), Newark, New Jersey, USA}
\maketitle
% \maketitle

\vspace{-3.95mm}
% Introduction - 1.5 Pages

% \ali{Changed FedL to FedL and FogL to FogL}
\section{Introduction}\label{sec:intro}
\noindent

\noindent  Rapid advancement of Internet-of-Things (IoT) and widespread adoption of wireless connectivity  (e.g., smart cars) have led to a surge in data-driven applications for 6G-and-beyond networks~\cite{lv2021big,khare2019big,sasaki2021survey}. Collectively, modern devices generate vast amount of data, enabling the development of powerful machine learning (ML) models~\cite{ahammed2023vision,mishra2022role}. Traditionally, these ML models are obtained using a \textit{centralized} approach, requiring devices to transfer their local data to a central server. However, centralizing data can result in excessive energy consumption, high latency, and privacy concerns~\cite{9060868}. 
These challenges  have fueled interest  in \textit{distributed ML} techniques.
% \vspace{-3mm}
\subsection{Distributed ML: Federated Learning \& Fog Learning}
Federated learning (FedL)~\cite{kairouz2021advances,konevcny2016federated,niknam2020federated,savazzi2020federated,abdisarabshali2023synergies, abdisarabshali2024dynamic, borazjani2024multi} is the most prominent distributed ML technique that shifts model training to data-collecting devices and uses the central servers to aggregate the locally trained models of the devices. Traditional FedL begins with a central server broadcasting a global model to devices and is characterized by two sequential steps repeated until model convergence: \textit{(i) Local training}, where each device updates its model using its dataset (e.g., via stochastic gradient descent (SGD)); (ii) \textit{Global model aggregations}, where the devices send their models to the central server, which aggregates the received models into a new global model (e.g., via weighted averaging) and broadcasts it to all devices. 
% Subsequently, traditional FedL is an ideal fit for the device-to-edge server architecture commonly found in edge networks.
% Nevertheless, when deploying FedL in large-scale networks (e.g., across multiple countries and continents), its learning framework should be adapted to fog networks architecture. Fog networks feature hierarchical architectures that encompass all the nodes in the cloud-to-things continuum, such as the cloud, core, metro, edge, clients, and things. 
Traditional FedL fits well with the device-to-edge architecture in edge networks. However, for large-scale networks (e.g., across countries/continents), FedL must be adapted to the \textit{fog network}, including nodes across the cloud-to-things continuum (e.g., the cloud, core, metro, edge, and things). This has led to the development of \textit{fog learning} (FogL)~\cite{hosseinalipour2020federated}, which addresses the extreme heterogeneity in communication  and computation capabilities of fog/edge network elements for effective implementation of distributed ML.

\subsection{Notion of Decentralization in FedL/FogL}
\begin{figure*}[t]
\vspace{-9mm}
\centering
{\includegraphics[width=1\textwidth, trim=3 3 3 3, clip]{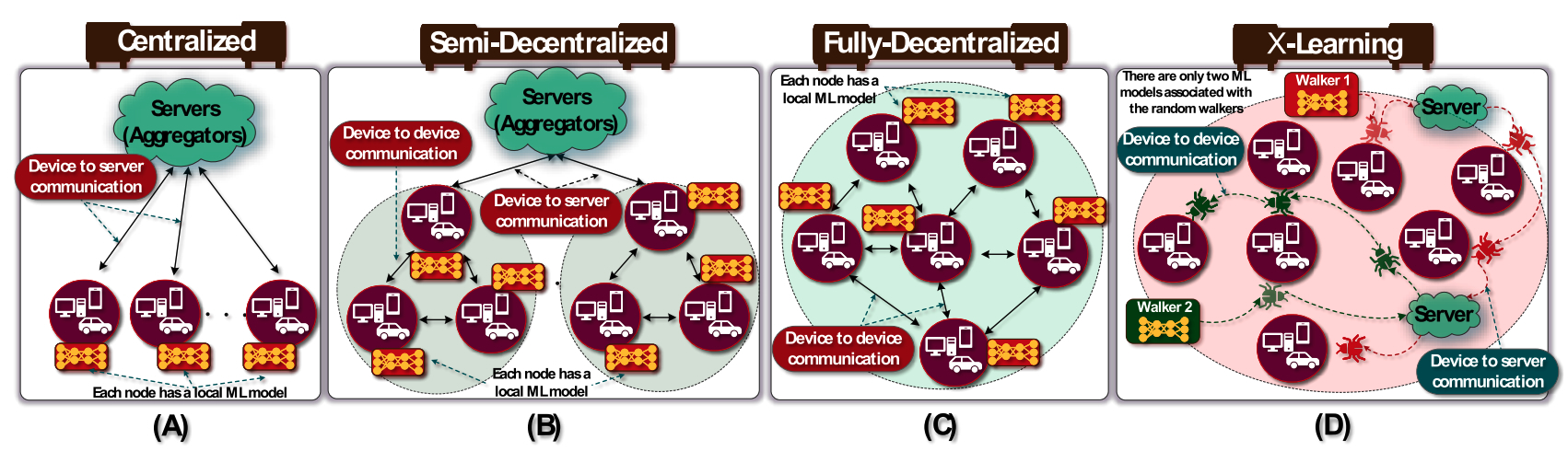}}
\vspace{-7mm}
\caption{Centralized to semi/fully-decentralized FedL/FogL architectures (A)-(C), and the extension presented by the $\mathbb{X}$L (D). In centralized to semi/fully-decentralized FedL/FogL, each node has its own local ML model, whereas in $\mathbb{X}$L there are only two ML models associated with the random walkers, inducing two active sessions per training round.}
\label{fig:arch}
\vspace{-4.5mm}
\end{figure*}

The \textit{notion of decentralization} has three major interpretations in distributed ML.
% There are three interpretations  of the \textit{notion of decentralization} in distributed ML.
\textit{(i) Centralized FedL~\cite{qi2023model,nguyen2020fast},} where ML training is conducted at devices and the model aggregations are conducted at a central server through device-to-server communications (Fig.~\ref{fig:arch}(A)). \textit{(ii) Semi-decentralized FogL/FedL~\cite{lin2021timescale,parasnis2023connectivity,yemini2022semi,yemini2023robust},} which involves ML training at devices followed by decentralized model aggregations within local clusters of devices through one-hop device-to-device (D2D) communications and centralized aggregations across the clusters (Fig.~\ref{fig:arch}(B)). 
\textit{(iii) Fully-decentralized FogL/FedL~\cite{beltran2023decentralized,roy2019braintorrent,9293091, savazzi2020federated},} where ML training is conducted at devices and model aggregations are executed solely via one-hop D2D communications (Fig.~\ref{fig:arch}(C)). 

% Both FedL and FogL envision three interpretations  of the \textit{notion of decentralization} in distributed ML. \textit{(i) Decentralized training, centralized aggregations,} where ML training is carried out at the devices and the model aggregations are conducted at a central server through device-to-server communications (Fig.~\ref{fig:arch}(A)). \textit{(ii) Decentralized training, semi-decentralized aggregations,} which is a hybrid approach that involves ML training at devices followed by decentralized model aggregation within local clusters of devices through one-hop D2D communications and centralized aggregation across the clusters \cite{hosseinalipour2020multistage} (Fig.~\ref{fig:arch}(B)). 
% \textit{(iii) Decentralized training, decentralized aggregations,} where ML training is carried out at devices and model aggregations are conducted through one-hop decentralized device-to-device (D2D) communications \cite{9293091, savazzi2020federated} (Fig.~\ref{fig:arch}(C)). 

% This architecture is shown to be particularly advantageous for FogL over large scale fog networks \cite{hosseinalipour2020multistage}.
% Schematics of these architectures are depicted in Fig.~\ref{fig:arch}(A)-(C). 

% \vspace{-3.5mm}
\subsection{From FedL/FogL to $\mathbb{X}$-Learning}
In this work, we are motivated to migrate away from the above three rigid interpretations of decentralization in distributed ML, and unveil a new flexible/generic notion of decentralization through $\mathbb{X}$-Learning ($\mathbb{X}$L) that extends FedL and FogL. We thus adopted the letter $\mathbb{X}$ in its name to represent a generic family of decentralization techniques, encompassing the aforementioned notions as its special cases.

% $\mathbb{X}$L at its core is inspired by the web crawlers that are used to index the contents of websites across the internet. Analogously, the fundamental principle of $\mathbb{X}$L is to consider ML models as \textit{autonomous agents or walkers} that crawl across devices within a network via D2D communications. Within $\mathbb{X}$L, independent ML models deploy \textit{a multi-hop, exploratory, random walk-based strategy} to traverse the network and learning from the data of the nodes. The mobility of a single walker allows it to be trained on diverse datasets which may yield more generalized models.

At its core, $\mathbb{X}$L is inspired by web crawlers used to index websites. The principle of $\mathbb{X}$L is to treat ML models as \textit{autonomous agents/walkers} that traverse devices/nodes through D2D communications. 
In $\mathbb{X}$L, independent ML models use \textit{a multi-hop exploratory strategy based on random walks} to traverse networks for distributed ML training. 
% The mobility of each walker enables it exposure to diverse datasets, potentially leading to more generalized models compared to the model trained on only one node.

\textbf{Generalized Notion of Decentralization in $\mathbb{X}$L:}
$\mathbb{X}$L takes a new perspective on two core threads in FedL/FogL architectures and proposes fresh alterations to them: 
\textit{(i) Predetermined/Rigid Model Transfers:} The interpretations of decentralization in FedL/FogL consider only \textit{predetermined/rigid} one-hop device-to-server or one-hop D2D communications, where all nodes are presumed to have a unified model exchange path/trajectory. Nevertheless, $\mathbb{X}$L extends this to \textit{flexible multi-hop D2D jumps} between the nodes under the presence or absence of edge servers in the network, paving the way for flexible paths/trajectories for model exchanges in complex networks.
\textit{(ii) Leashed/Tied Local Model-to-Node Logic:}
In FedL/FogL, each device is treated as an \textit{identity} with a local model (i.e., a neural network (NN)).
% The conventional FedL/FogL models confine one local ML model (i.e., a neural network (NN)) to each node, which resembles tying/leashing a local NN to each participating client. 
Nevertheless, $\mathbb{X}$L shifts this identity to the ML models, treating them as autonomous agents that can jump between nodes and exploit their datasets and computation/communication resources.  $\mathbb{X}$L further empowers ML models to \textit{pull} each other or engage in \textit{collisions} to replicate model aggregations as we will describe. 
% In other words, in FedL/FogL models, \textit{the focus is on nodes} that each have a local ML model and train it locally using their predetermined decisions (e.g., number of SGD iterations) and model exchange. 
% In particular, $\mathbb{X}$L makes the ML models to break free from the nodes and puts the \textit{emphasis on the ML models} themselves, treating them as autonomous agents that move between nodes and perform ML model training at their discretion and have the ability to \textit{pull} each other or engage in \textit{collisions} to replicate model aggregations as later discussed.

% $\mathbb{X}$L takes this new notion one step further to enable personalized training. In particular, \ali{XXXXXXXX}.

\textbf{Reduction of $\mathbb{X}$L to FedL/FogL:}
$\mathbb{X}$L, visualized in Fig.~\ref{fig:arch}(D), is a foundational distributed ML architecture that can be reduced to FedL/FogL. 
For example, when one walker/NN is placed at each  node and set to follow a predetermined path between the node and a server, $\mathbb{X}$L reduces to  FedL with centralized aggregations. Also, when one walker/NN is placed at each node and set to follow a predetermined one-hop path between the hosting node and its neighboring nodes, $\mathbb{X}$L can replicate decentralized/semi-decentralized FedL/FogL. Giving autonomy to the walkers and removing the assumption on their set/fixed paths, which in turn allows them to conduct multi-hop jumps across the nodes, opens unique opportunities for the development of flexible and generalized distributed ML methods under the umbrella of $\mathbb{X}$L.

\vspace{-0.5mm}
\section{Related Work,  Applications, and Challenges}
% \vspace{-0.5mm}
\subsection{Related Work}
$\mathbb{X}$L shares some architectural similarities to two distributed ML methods, yet possesses key distinguishing features.

\textbf{(1) Gossip Learning:}
Gossip learning involves nodes selecting peers to exchange ML models via D2D links~\cite{1498447,yang2022decentralized,5545370,hu2019decentralized,9440060,9954055,nadiradze2021asynchronous,jin2023privacy,7165636,10286430,assran2019gossip,4545250,10266757}, where peer selection is often blind/random. In contrast, $\mathbb{X}$L features a \textit{purposeful/intelligent} approach, where autonomous walkers/NNs traverse nodes over \textit{optimizable/deliberate trajectories} and \textit{model training durations}. 
% Subsequently, $\mathbb{X}$L allows for configurable parameters such as node sampling and  . 
% Additionally, $\mathbb{X}$L focuses on optimizing the navigation and learning strategies of these walkers/NNs. 
While gossip learning is typically suited where blind/random D2D model transfers are sufficient (e.g., when nodes' data is homogeneous), intelligent navigation across nodes combined with adaptable local ML model training make $\mathbb{X}$L applicable to heterogeneous/non-iid (non-independent and identically distributed) data across irregular network topologies.

\textbf{(2) Random Walk-based Learning:}
Research has explored random-walk-based distributed learning, focusing on graph-specific tasks such as \textit{graph learning} \cite{perozzi2014deepwalk,grover2016node2vec,liu2020towards} (e.g., extracting structures/communities from a graph).
% Random walkers have also been used as components in other learning methods, including graph convolutional networks \cite{kipf2016semi}, addressing node-level tasks on graph-structured data like node classification and link prediction. 
Migrating from graph-specific tasks, $\mathbb{X}$L \textit{offers a fresh perspective to random walk-based learning} by widening its scope to distributed ML tasks involving the training of single/multiple NNs over non-iid data scattered across a network of nodes.
The closest work to $\mathbb{X}$L is a recent study that uses a single random walker in a graph, while balancing the exploration and exploitation of distributed data in FedL \cite{ayache2023walk}. Compared to this, $\mathbb{X}$L is designed to be a generic distributed ML framework and introduces several degrees of freedom structured into two major innovations.
\textbf{(Innovation~1):} Starting with a single walker (Sec.~\ref{sec_3}), we 
highlight the inadequacy of conventional graph theoretic centrality measures as well as Markov chain analysis for $\mathbb{X}$L, which triggers a set of novel concepts: (i) integration of \textit{node' data and spatial qualities}; (ii) \textit{elastic learning}; (iii) \textit{perception of walkers}; (iv)  \textit{memory} of walkers. 
\textbf{(Innovation~2):} We further envision multiple interacting walkers in $\mathbb{X}$L (Sec.~\ref{sec_4}),  unveiling the following novel concepts: (i) optimal \textit{interactions/collaborations} between walkers;  (ii) \textit{attraction/repulsion} between  walkers. These innovations further trigger a variety of future directions that we will highlight.

% \textbf{(Innovation  3):} A set of complementary techniques for random walk-based learning, including   memory, progressive/conservative walkers, and expanding/contracting walkers.

\vspace{-1mm}
\subsection{Applications and Scenarios of Interest}

 Our choice of a random walk-based coordination mechanism in $\mathbb{X}$L is driven by both architectural motivations and its versatility across diverse deployment environments. Compared to conventional centralized or decentralized FedL/FogL paradigms, the use of mobile learning agents (i.e., random walkers) offers the following key advantages:

\textit{1. Minimizing Concurrent Active Sessions for Resource Efficiency:}
In typical FedL/FogL setups, large portions of the network nodes may be active simultaneously, creating considerable demand on bandwidth, energy, and processing as highlighted in the existing literature~\cite{tao2024federated,letaief2021edge}. In contrast, $\mathbb{X}$L decouples training activity from network size: with $R$ walkers in the system, only $R$ nodes are active at a time (only the current node being visited by each walker is active in model training and communication). In the limiting case of $R=1$, $\mathbb{X}$L enables distributed learning with only one active learning sessions at a time across the network nodes. This makes $\mathbb{X}$L especially suitable for \textit{resource-constrained networks} (e.g., low-power sensor networks or interference-prone wireless mesh setups), where controlling the number of concurrent training sessions is critical.

\textit{2. Implicit Topology-Aware Exploration without Global Coordination:}
The walker's localized traversal enables structure-aware, decentralized exploration of the network. Without requiring global scheduling or aggregation, the walker naturally adapts to partial, dynamic, or intermittently connected topologies. This is particularly beneficial in \textit{opportunistic and volatile networks}, such as vehicular and aerial swarms, where D2D communication links are ephemeral and nodes may only be momentarily adjacent.

\textit{3. Scalability, Autonomy, and Lightweight Design:}
$\mathbb{X}$L scales gracefully as the walker’s activity is localized and independent of overall network size. The absence of a centralized server or full-mesh communication model allows learning to proceed in autonomous \textit{networks lacking cellular infrastructure}, such as remote/rural areas. Here, a single walker  initialized at one node can propagate across the network, training on local datasets and adapting the model as it goes.

\textit{4. Support for Large-Scale and Social Networks:}
The exploratory nature of the $\mathbb{X}$L's random walk mechanism makes it naturally applicable to large peer-to-peer networks, such as \textit{social network graphs}. In such networks (e.g., Facebook, Twitter), walkers can traverse along edges (friendship/follow relations) to perform decentralized and privacy-preserving learning, where centralized coordination is infeasible or undesirable.

In summary, $\mathbb{X}$L extends FedL/FogL paradigms by providing a unified mechanism that is communication/computation-efficient, topology-aware, and adaptable to dynamic environments, while ensuring minimal active node footprint and natural support for scalability.

% \vspace{-4mm}
\subsection{Challenges and Design Considerations}

% \section{Fundamentals of $\mathbb{X}$-Learning over Heterogeneous Fog/Edge Networks}
% A keystone feature of $\mathbb{X}$L is its implementation of autonomous random walkers (i.e., NNs) as mobile learners. Nevertheless, 

$\mathbb{X}$L's novel and flexible nature, in which autonomous random walkers (i.e., NNs) are deployed as mobile learners, presents several strategic questions and challenges.

\begin{itemize}[leftmargin=4.5mm]
    \item \textbf{Which Node to Choose and How to Interact? (Spatial Challenge):} In $\mathbb{X}$L, after finishing the ML model training at a node, walkers face the major challenge of selecting the next node in the network/graph to jump to. Additionally, in scenarios with multiple active walkers, they must determine `if' and `how' to interact with each other.
    % to improve their overall learning efficiency.
    \item \textbf{How Long to Stay? (Temporal Challenge):} In $\mathbb{X}$L, after arriving at a node, walkers face the major challenge of how long to train at that node (e.g., how many SGD iterations should be conducted).
    \item \textbf{How to Integrate the Learning Result? (Learning Challenge):} In $\mathbb{X}$L, after training its model on a node's data, the walker faces a challenge of how to use this new model (e.g., whether to use it directly, replace its previous model with it, or conduct a weighted average between the two).
\end{itemize}
% \begin{tcolorbox}[colback=darkyellow2, colframe=salmon2, title=\textbf{The Vision of $\mathbb{X}$L}]
\begin{tcolorbox}[colback=white,colframe=black,colbacktitle=black,
title=\centering \small Box 1: The Vision of $\mathbb{X}$L,fonttitle=\bfseries]
Finding an overarching strategy addressing the above questions, combining both network- and data-dependent considerations, is the major challenge in $\mathbb{X}$L.
% Finding  an overarching strategy that addresses all the above questions is non-trivial and network/data-dependent.
At first glance, it may seem that answering these questions can be informed by a mixture of concepts drawn from (i) Markov chains in non-Euclidean spaces (graph topologies), (ii) graph theory, and (iii) distributed ML. However, in the subsequent discussions, we unveil that the existing frameworks/interpretations from these fields are insufficient to fully capture the dynamics inherent to $\mathbb{X}$L. This is mainly because these frameworks are developed to focus on either the \textit{graph topology} (e.g., Markov chains in non-Euclidean spaces~\cite{bloch2023centrality,borgatti2006graph}) or the \textit{data quality} of the nodes (e.g., majority of distributed ML literature~\cite{9060868}). \textbf{Consequently, the efficient design and implementation of $\mathbb{X}$L is an untapped area of research, which calls for a series of follow-up works.}

\end{tcolorbox}

\subsection{Overarching Formulation of $\mathbb{X}$L}\label{sec:Formulation}
 In a nutshell, the central goal of $\mathbb{X}$L is to optimize the movement and behavior of the walkers to achieve two key objectives simultaneously: (i) maximizing the \textit{accuracy} (i.e., minimizing the \textit{loss}) of their learned models and (ii) minimizing the \textit{operation cost/overhead} of their jumps and interactions (e.g., in terms of communication, latency, and storage).
To formalize this goal, consider $\mathbb{X}$L operating with a set of random walkers denoted by $\mathcal{R}$, where each walker is indexed by $r \in \mathcal{R}$.
Also, let $\mathcal{U}$ denote the set of nodes (e.g., edge devices) in the system, and let $\mathcal{D}_u$ represent the local dataset stored at node $u \in \mathcal{U}$. The global dataset is defined as $\mathcal{D} = \bigcup_{u \in \mathcal{U}} \mathcal{D}_u$, representing the union of all local datasets across the network. For an arbitrary ML model $\bm{\omega}$, we define the \textit{global loss function} of $\mathbb{X}$L as
$
\mathfrak{L}(\bm{\omega}) = \sum_{u \in \mathcal{U}} \frac{|\mathcal{D}_u|}{|\mathcal{D}|} \mathfrak{L}_u(\bm{\omega}),
$
where $\mathfrak{L}_u(\bm{\omega})$ is the \textit{local loss function} at node $u$, given by
$
\mathfrak{L}_u(\bm{\omega}) = \frac{1}{|\mathcal{D}_u|} \sum_{d \in \mathcal{D}_u} f(\bm{\omega}, d),
$
with $f(\bm{\omega}, d)$ denoting the loss (e.g., cross-entropy) incurred by model $\bm{\omega}$ on data point $d$.
We presume an episodic view of the network evolution, where each episode, indexed by $k\in\mathcal{K}$ corresponds to a global jump event, i.e., a full round in which all random walkers in 
$\mathcal{R}$ complete one jump. At the end of each episode 
$k$, we define the average model parameter across all walkers (e.g., obtained via FedAvg~\cite{mcmahan2017communication}) as $\bm{\omega}^{(k)}$. This average serves as a (hypothetical) global model that would result if training were to halt at episode 
$k$ and all walkers' models were synchronized through aggregation for the downstream ML task of interest. The high-level objective of $\mathbb{X}$L can then be formulated as the following optimization problem:

\begin{equation}
\hspace{-8mm}
\begin{aligned}
\min_{\{\bm{P}^{(k)},\bm{\phi}^{(k)}\}_{k\in\mathcal{K},r\in\mathcal{R}}} & \Gamma \times \underbrace{\frac{1}{|\mathcal{K}|}\sum_{k\in\mathcal{K}}\Big\Vert \nabla\mathfrak{L}\big(\bm{\omega}^{(k)}\big)\Big\Vert_{_2}^2}_{\text{ML Performance Metric}} + (1-\Gamma) \times \underbrace{\frac{1}{|\mathcal{K}|}\sum_{k\in\mathcal{K}}\mathfrak{C}\Big(\{\bm{P}^{(k)}, \bm{\phi}^{(k)}\}_{r\in\mathcal{R}}\Big)}_{\text{Operation Cost/Overhead}} \\
\text{s.t.} \quad & \bm{P}^{(k)} \in \mathscr{P}, ~ \bm{\phi}^{(k)} \in \bm{\Phi},
\end{aligned}
\end{equation}
where we have considered the 2-norm of the gradients, capturing the proximity to a stationary point, as the performance metric of the ML model within the objective function. The components of this optimization problem are defined below.

\begin{itemize}[leftmargin=3.5mm]
    \item $\bm{P}^{(k)}$: Transition probability matrix of each walker $r$ at episode $k$, defining the walker's jump behaviors across nodes.
    \item $\bm{\phi}^{(k)}$: Generic set of tunable parameters of each walker $r$ at episode $k$ that influence walkers' operations, such as number of SGD iterations, and memory length.
    \item $\nabla \mathfrak{L}(\bm{\omega}^{(k)})$: The gradient of the global ML loss of $\mathbb{X}$L evaluated at $\bm{\omega}^{(k)}$. 
    
    \item $\mathfrak{C}\Big(\{\bm{P}^{(k)}, \bm{\phi}^{(k)}\}_{r\in\mathcal{R}}\Big)$: The overall operation cost/overhead of $\mathbb{X}$L, comprising all the walkers at episode $k$, which is context-specific and can comprise:
    \begin{itemize}
        \item Jump overhead: Communication and latency costs linked to walkers' jumps (captured by $\bm{P}^{(k)}$ across the walkers $r\in\mathcal{R}$).
        \item Memory overhead: Walkers' memory footprint (affected by $\bm{\phi}^{(k)}$ across the walkers $r\in\mathcal{R}$).
        \item Computation overhead: Cost of local training and aggregations (affected by $\bm{\phi}^{(k)}$ across the walkers $r\in\mathcal{R}$).
    \end{itemize}
    \item $\Gamma\in[0,1]$: Trade-off coefficient balancing the importance of objective terms.
    \item $\mathscr{P}$: Feasible set ensuring that each $\bm{P}^{(k)}$ is a valid stochastic transition probability matrix (e.g., ensuring that it is row stochastic).
    \item $\bm{\Phi}$: Feasible set of each $\bm{\phi}^{(k)}$ (e.g., imposing communication/computation constraints).
\end{itemize}

$\star$~\textbf{Scope of This Work:} We note this perspective paper does not aim to provide a formal mathematical treatment of the above optimization problem (e.g., obtaining the closed-form expressions of the objective terms and constraints for various network configurations), left as an open research direction. Instead, \textbf{our focus is to (i) unveil the fundamental degrees of freedom (DoF) in $\mathbb{X}$L that affect the aforementioned optimization, (ii) articulate them intuitively, and (iii) propose simple heuristics that can achieve notable performance gains even without solving the above problem.} Moreover, to keep our discussions simple, we abstract the cost function in the objective as \textit{the number of jumps} performed by the random walkers, which is captured by the x-axis of the subsequent figures. Henceforth, we first elaborate on the DoF in $\mathbb{X}$L and then consider the case of $\mathbb{X}$L with a single random walker as a gateway to our discussions.

% In this magazine-style paper, our goal is \underline{\textbf{not}} further concretization of this problem, which we leave it as an open direction (e.g., obtaining convergence analysis to characterize the first term of the objective); instead, our goal is to 
% show the fundamental degrees of freedom in $\mathbb{X}$-Learning and articulate them in an intuitive way and further propose some simple heuristics to yield notable performance gains even without solving the above problem in its entirety. Further, to simplify our discussions, the cost function in the objective is simply abstracted as the number of jumps the random walker conducts, reflected in the x-axis of the subsequent figures.

\subsection{Degrees of Freedom (DoF) in $\mathbb{X}$L}
In the above formulation, variables $\{\bm{\phi}^{(k)}\}_{r\in\mathcal{R},k\in\mathcal{K}}$ capture the DoF in $\mathbb{X}$L. In particular, DoF of a learning framework refer to the tunable design parameters or control variables that determine the system’s behavior, adaptability, and performance. In the context of $\mathbb{X}$L, the degrees of freedom arise from its modular and agent-based architecture, where learning and communication are guided by mobile agents (walkers) traversing a communication graph. This architecture allows several DoF, which are:

\textit{\textbf{1. Number of Walkers:}} This DoF controls the level of parallelism in the system. Multiple walkers can be deployed simultaneously, enabling concurrent training across (potentially disjoint) graph regions or communities. Increasing the number of walkers improves coverage and speeds up convergence but raises concerns on communication/computation overhead.

\textit{\textbf{2. Walker Transition Policy:}}
This DoF specifies how walkers move across nodes through their dedicated transition probability matrices. This can range from uniform random walks to biased movements guided by node utility (e.g., loss, uncertainty), structural features (e.g., centrality), or even exogenous constraints (e.g., mobility patterns in vehicular networks). 

\textit{\textbf{3. Local Update Strategy per Visit:}}
This DoF defines how many local training steps are executed at each node per visit, which optimizer is used, and how learning rate is adjusted. This affects the balance between the walker's model local accuracy (evaluated at the visiting node) and  generalization (evaluated at the global dataset distributed across the nodes).

\textit{\textbf{4. Walker Memory or Statefulness:}}
This DoF enables walkers to perform more informed and context-aware navigation across the network. In particular, a walker can be:
(i) stateless, where its next move depends only on the current node and local transition probabilities, or
(ii) stateful, where it maintains a memory of past visits, cumulative training outcomes (e.g., loss or model), or trust-related metadata (e.g., reputation scores of visited nodes).
In this setting, stateful walkers can implement intelligent routing policies; for example, by avoiding: nodes that have already been sufficiently trained, nodes previously flagged as unreliable or adversarial, or regions of the graph that are currently oversampled. Such statefulness introduces a new layer of autonomous decision-making within each walker.

\textit{\textbf{5. Acceptance and Aggregation Criteria at Nodes:}}
This DoF enables each network node to define local rules for accepting, rejecting, or modifying incoming models carried by walkers. Such rules enhance node-level autonomy and allow for flexible trust management and model validation. Examples include:
(i) validation filtering (e.g., using a small local validation set to evaluate the performance of the incoming model and accepting it only if it meets a quality threshold for the local dataset),
(ii) update averaging (e.g., integrating the incoming model with the node’s existing model via weighted averaging, potentially based on recency or reputation),
and (iii) rejection thresholds (e.g., discarding updates if the local model performance drops or divergences). 

\textit{\textbf{6. Scheduling and Lifespan of Walkers:}}
Walkers may be persistent (long-lived agents) or ephemeral (restarted periodically), where this DoF defines \textit{when} and \textit{how} new walkers are launched.

In our subsequent discussions, we will shed light on some of these DoF and their impacts on the performance of $\mathbb{X}$L.

\section{$\mathbb{X}$-Learning with a Single Random Walker}\label{sec_3}

\noindent Considering a single walker traversing a network, the learning strategy of the walker consists of \textit{jumping/traversing} between nodes and \textit{training its model} on their local data using SGD.
% Within $\mathbb{X}$L, a random walker dynamically and intelligently navigates and trains across the network, and the number of active nodes in the training process is equal to the number of random walkers. This is unlike in traditional FedL methods where large portions of the nodes in a graph may be active during each training step. 
Through both conceptual reasoning and simulations, we next show that this learning strategy is tied to  \textit{node importance}.
% and \textit{node temperature}.

% \begin{tcolorbox}[colback=darkyellow2, colframe=salmon2, title=\textbf{Data Distribution and Network Topology}]
\begin{tcolorbox}[colback=white,colframe=black,colbacktitle=black,
title=\centering \small Box 2: Data Distribution and Network Topology,fonttitle=\bfseries]
All simulations use a four-layer convolutional neural network trained on CIFAR-10 \cite{CIFAR10} and SVHN \cite{SVHN} datasets, which are widely used in FedL/FogL studies. Each simulation will specify its own non-iid data partitioning. Two network topologies are considered:
\begin{itemize}[leftmargin=3.6mm]
    \item \textbf{Caveman Graph:} Used to model social/sensor networks with tight regional communities/connectivity.
    \item \textbf{Random Geometric Graph (RGG):} Used to model D2D links over wireless networks~\cite{hosseinalipour2020multistage}.
\end{itemize}
$\star$ These graphs were intentionally chosen to examine $\mathbb{X}$L in \textit{network topologies with highly unbalanced (caveman) and semi-unbalanced (RGG) connectivity}.

% , much like how FedL/FogL is often examined under different \textit{data heterogeneity} levels (i.e., extreme vs. mild non-iid data distributions).

% various and yet challenging network topologies, much like how FedL is examined under different data heterogeneity settings (i.e., IID vs. non-IID distributions).
\end{tcolorbox}

\subsection{Node Importance/Centrality}\label{sec:temp}

In both $\mathbb{X}$L and FedL/FogL, the quality of the local data at a node is a key to determining its importance: the more representative the local data is of the \textit{global dataset} (i.e., the union of the nodes' local data), the higher its quality. Although this metric alone may suffice for FedL/FogL, it is \textit{insufficient} for $\mathbb{X}$L. To demonstrate this, consider a node with poor data quality serving as a connective bridge or hub between multiple cliques of nodes, each containing high-quality data. 
Here, while the node itself may not offer high-quality data for training, its \textit{network centrality} (i.e., spatial quality) can make it highly valuable for the walker to visit as a gateway for reaching nodes with higher-quality data.
As a result, in $\mathbb{X}$L, if we aim to model the jumps/traversal between the nodes for a walker through a probability transition matrix (similar to Markov chains over graphs), this matrix should be a function of \textit{both} the nodes' data qualities and their network centralities.
% Motivated by this, for determining the quality of a node, we envision combining the node's spatial quality with that of its data quality. 
To achieve this goal, the concept of centrality in graph/network theory (e.g., measured via degree, betweenness, eigenvalue, and closeness metrics~\cite{grando2016analysis,bloch2023centrality,borgatti2006graph}) should be integrated to the concept of data quality (e.g., measured via the disparity of gradients generated on the local data of a node versus the global dataset~\cite{9060868}).
By jointly considering spatial and data quality metrics, \textit{walkers in $\mathbb{X}$L environments follow trajectories that are adaptive to nodes' data distribution and network topology}. However, selecting a suitable realization for each metric and determining how to combine them (e.g., weighted sum of data quality and spatial quality) remains a non-trivial and open research challenge.

% determining the relative weight of data quality vs. spatial importance in a node's overall desirability, remains a non-trivial and open research problem.

\begin{figure}[t]
\vspace{-5.75mm}
\centering
{\includegraphics[width=0.7\textwidth]{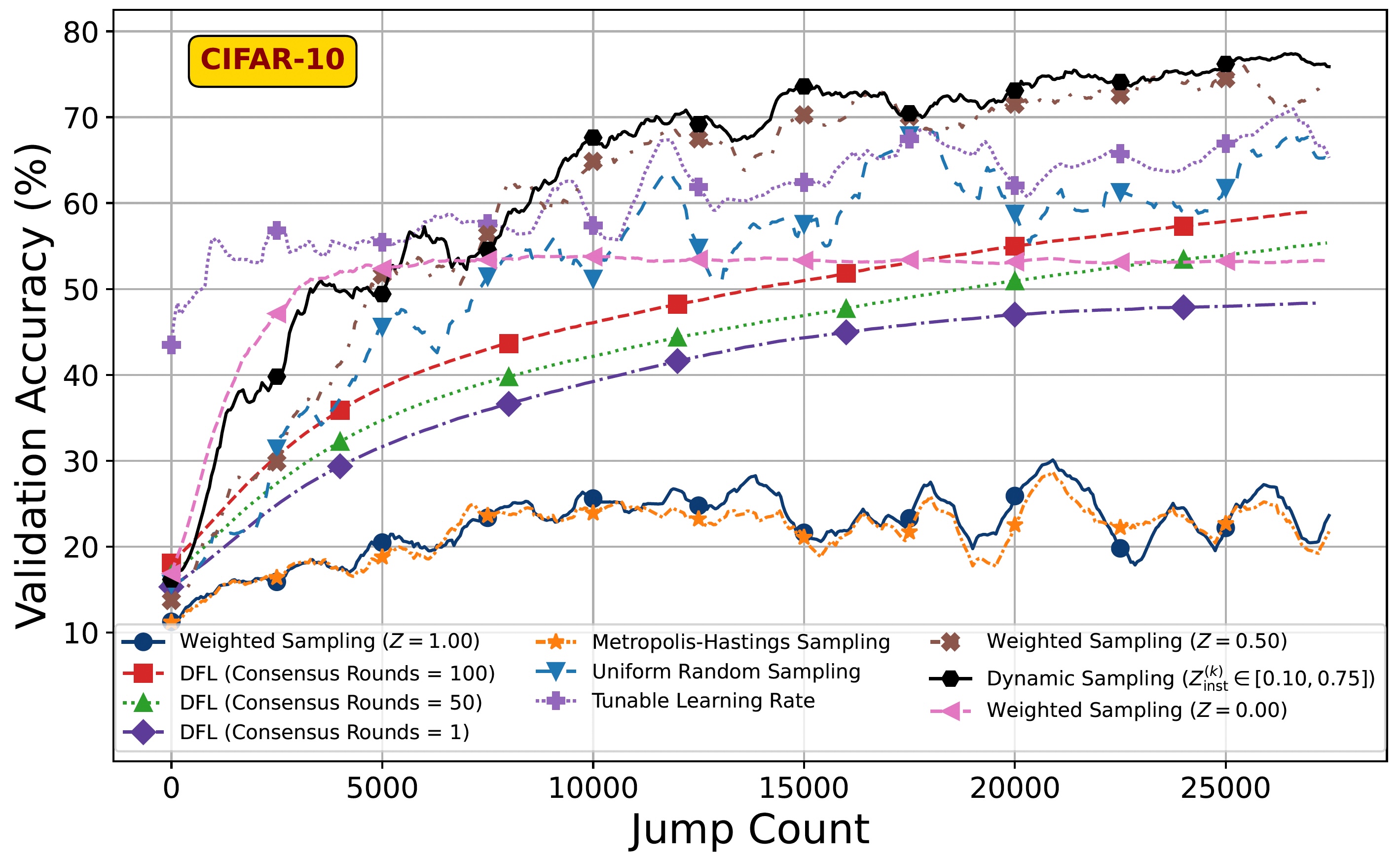}}\\[1mm]
\centering
{\includegraphics[width=0.7\textwidth]{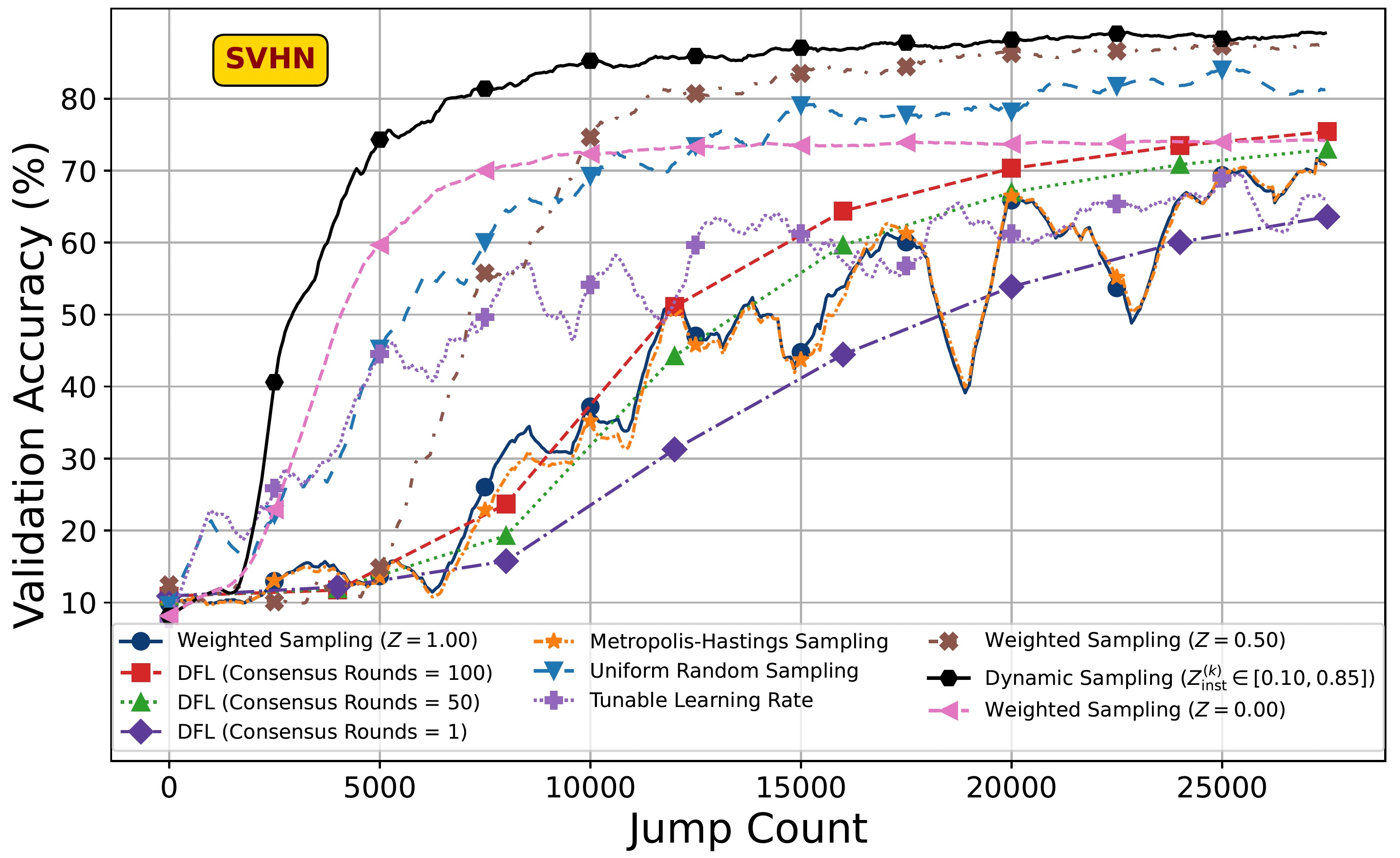}}\\[1mm]
\vspace{-3.5mm}
\caption{Performance of random walkers under different node traversal strategies for CIFAR-10 and SVHN datasets (smoothed using a moving average with a window size of 10). Referring to \textbf{Sec.~\ref{sec:temp}}, among methods with \textit{fixed/static} node importance values (red, blue, green, magenta, and yellow curves using static sampling or fixed $Z \in \{0, 0.5, 1\}$), our approach with non-binary weighting of data and spatial quality metrics (yellow curve, $Z = 0.5$) yields the best performance. Furthermore, referring to~\textbf{Sec.~\ref{sec:perception}}, our method with \textit{dynamic} node importance (black curve using a time-varying  $Z^{(k)}_{\mathsf{Inst}}$ based on current model accuracy) outperforms all the static strategies.}
\label{fig:singlecrawl}
\vspace{-1.5mm}
\end{figure}

% The design of the transition matrix, therefore, becomes a non-trivial challenge, as the degree to which data quality and spatial quality each contribute to the importance of a node in a graph may vary greatly depending on the type of graph and distribution of data across all the nodes. Through these design decisions, walkers in $\mathbb{X}$L environments have trajectories that are responsive to the data distribution and structure of the network graph.

To validate the above-described approach to node importance, we conduct a simulation and present it in Fig.~\ref{fig:singlecrawl}. Here, we consider a caveman graph comprising $25$ nodes clustered into $8$ cliques. The nodes' data distribution is non-iid with $80\%$ of nodes only containing data samples from $1$ or $2$ of the $10$ labels in CIFAR-10/SVHN dataset. 
% We assume that the walker traverses the nodes according to various strategies, including uniform random sampling (blue curve) and Metropolis-Hastings sampling~\cite{chib1995understanding} (green curve) --- a well-known technique in Markov chains over graphs. 
% We consider two cases of our node importance method, where node $i$'s importance is measured through the following weighted-sum linear modeling  $\widetilde{\mathfrak{c}}_u=\alpha\times d_u\times \ell_u + (1-\alpha) \times c_u$.
% in which we model the walkers' affinity towards \textit{data} and \textit{spatial} quality via a  weighting parameter $\alpha \in [0, 1]$, where node $i$'s importance is measured through the following weighted-sum linear modeling  $\widetilde{\mathfrak{c}}_u=\alpha\times d_u\times \ell_u + (1-\alpha) \times c_u$. 
We consider a \textit{simple} realization of our node importance approach, where \textit{the importance of node} $u$ is computed as a weighted sum: $\widetilde{\mathfrak{c}}_u = Z \cdot d_u \cdot \xi_u + (1 - Z) \cdot c_u$. Here, the weighting parameter $Z \in [0, 1]$ controls the walker’s preference between data and spatial quality.
In this formulation, $d_u$ and $\xi_u$ represent the fractions ($d_u,\xi_u\in [0, 1]$) of global dataset and labels present at node $u$ (i.e., capturing \textit{data} quality), respectively, and $c_u$ is a conventional measure of the node's centrality (chosen to be betweenness centrality~\cite{bloch2023centrality,borgatti2006graph}, capturing the \textit{spatial} quality). At each node, the transition probability of the walker to the neighboring nodes is defined proportional to their importance values ($\{\widetilde{\mathfrak{c}}_u\}$). 
% From this formulation, lower values of $\alpha$ make the walker favor spatial quality while higher values assign preference to data quality. 
%\ali{bring the exact formula and add discussions!}
We compare our proposed method with four relevant baseline methods from the literature: 

\textit{1) Pure Gossip-based Learning with Uniform Random Sampling (denoted by `Uniform Random Sampling' in the legend):} Inspired by~\cite{liu2024technical}, this baseline represents a classical decentralized learning approach in which the random walker selects its next node uniformly at random from its current location’s neighborhood. Specifically, if the walker is currently at node $u$ with neighbors $\mathcal{N}(u)$, it chooses the next node with equal probability $1/|\mathcal{N}(u)|$ from this set. 
% This serves as a representative of memoryless, structure-agnostic random walks.

\textit{2) Random Walk with Metropolis-Hastings Sampling (denoted by `Metropolis-Hastings Sampling' in the legend):} Following the Metropolis-Hastings methodology from Markov Chain Monte Carlo (MCMC) literature~\cite{chib1995understanding}, this method selects the next node based on an acceptance probability that depends on a target stationary distribution (i.e., node degrees). This allows the walker to quickly explore the network with minimal information about the node degrees while maintaining ergodicity and avoiding local traps.

\textit{3) Decentralized Federated Learning (denoted by `DFL' in the legend):} This baseline reflects traditional decentralized FL (DFL) approaches~\cite{beltran2023decentralized} in which each node collaborates directly with its neighbors through model averaging. Unlike our framework, which features an active learner navigating the graph, DFL distributes the training workload across the network via periodic inter-node  D2D communications (i.e., it has a rigid one-hop model communication among the nodes). During these D2D communications, also referred to as \textit{consensus rounds}, nodes compute the average of their neighbors’ models: as the number of consensus rounds increases, the estimation of the average model across neighbors becomes more accurate. In our plots, we depict the performance of DFL under varying numbers of consensus rounds, where the consensus matrix is constructed following the method in~\cite{hosseinalipour2020multistage}.
To ensure a fair comparison in our experiments, we treat each model update at a node in DFL as equivalent to a single ``jump" of the random walker, since in our framework, the walker also trains on one node per jump when reporting performance across figures.

\textit{4) Random Walk with Tunable Learning Rate (denoted by `Tunable Learning Rate' in the legend):} Inspired by one of the advanced techniques in random walk-based FedL~\cite{ayache2023walk}, this method adjusts the transition probabilities of the walker adaptively based on the observed loss values of visited nodes, allowing the walker to prioritize undertrained nodes over time. Nevertheless, the method does not explicitly take into account the centrality (i.e., spatial importance) of the nodes.

As shown in Fig.~\ref{fig:singlecrawl}, our scheme  that obtains node importance as a balanced average ($Z=0.5$) between data and spatial quality, significantly outperforms other strategies, including the ones that only consider spatial quality ($Z=0$) and data quality ($Z=1$).
% This even value of $\alpha$ would \textit{not} hold for all environments, but the results convey the \textit{importance of considering the network topology in conjunction with the nodes' data quality}, suggesting a departure from the literature on both Markov chains over graphs and FedL/FogL.
This balanced value of $Z$ may not hold across all environments, but the results highlight the \textit{significance of considering network topology alongside the nodes' data quality}. 
We later (Sec.~\ref{sec:perception}) show that we can further obtain dynamic transition matrices to achieve a better performance (i.e., the black curve in Fig.~\ref{fig:singlecrawl}).

Motivated by the above results, we conjecture that optimal node importance measurement and node traversal in $\mathbb{X}$L demands moving beyond conventional graph exploration and distributed ML approaches, such as Markov chain-based methods that focus solely on network topology while ignoring node data quality, and FedL/FogL techniques that emphasize only data quality while overlooking dynamic multi-hop graph exploration via random walks. A formal treatment of this conjecture is left for future work.  In what follows, we present a set of informal yet insightful notes to guide the design of  effective node importance measurement for $\mathbb{X}$L.

\textbf{Node Importance Estimation (Fusing Graph Centrality and Data Utility in $\mathbb{X}$L):} 
As explained above, to assess the importance of a node, $\mathbb{X}$L must consider a combination of:
\textit{(i) spatial quality} (i.e., the topological significance or connectivity of the node in the communication graph) and
\textit{(ii) data quality} (i.e., the informativeness or uniqueness of the local dataset). An effective node selection strategy should this fuse these two dimensions (i.e., spatial importance and data quality) to determine the utility of a visit for the walker. 
To estimate these node importance metrics within $\mathbb{X}$L, several strategies can be used: 
\\

\textit{\textbf{1. Estimating Spatial Quality via Graph-Theoretic Centrality Metrics (e.g., Degree, Betweenness, Closeness, and Eigenvector Centrality):}} These metrics capture the structural significance of nodes within the communication graph, helping identify nodes that are well-connected, serve as bottlenecks, or are topologically influential.
\begin{itemize}
    \item \textit{Associated Overhead:} Some centrality metrics, such as degree centrality, can be computed locally using only the node’s immediate neighborhood. Others, including betweenness, closeness, and eigenvector centrality, generally require global or multi-hop information and are more suitable for static or slowly-changing topologies where such computations or approximations are feasible.
\end{itemize}

\textit{\textbf{2. Estimating Data Quality via Local Loss or Validation Error:}} Walkers or nodes maintain and update estimates of local training loss or validation error, which are then used to prioritize undertrained or high-error  nodes (i.e., nodes with high validation loss) during walkers' traversals.
\begin{itemize}
    \item \textit{Associated Overhead:} This approach requires lightweight local model evaluation and a small validation dataset at each node or walker, with no need for inter-node communications.
    \end{itemize}
    
\textit{\textbf{3. Estimating Data Quality via Data Uniqueness or Class Rarity:}} Nodes indicate whether they possess rare or underrepresented data samples (e.g., they distribute the information about their held local labels to their neighboring nodes and walkers), such as minority classes or unique modalities, thereby signaling high training value.
\begin{itemize}
    \item \textit{Associated Overhead:} Requires occasional sharing of class distributions among nodes/walkers, which may incur communication or privacy overhead depending on the deployment.
    \end{itemize}
    
\textit{\textbf{4. Estimating Data Quality via Reputation or Trust Scores:}} Walkers accumulate and share scores that reflect the quality, consistency, or past usefulness of nodes' local model updates, helping walkers avoid unreliable or malicious participants.
\begin{itemize}
    \item \textit{Associated Overhead:} Involves maintaining historical feedback or trust logs at the walkers, and may require  propagation of information among the walkers upon collision (i.e., when they visit the same nodes).
   \end{itemize} 

  \textit{\textbf{5. Estimating Data Importance via Global Distribution Statistics (e.g., Gradient Magnitude and Variability):}}
Node data importance can also be estimated by evaluating patterns in gradient statistics, such as the magnitude/variability of gradients generated at each node compared to the overall cumulative gradient generated across all the nodes. This allows identification of nodes that contribute significantly to model updates (i.e., those that their local data distribution is a better representative of the global data distribution) or represent underfit regions of the data space.
Such statistics can be aggregated in two ways:
(i) Server-based aggregation, where a central coordinator collects and analyzes per-node gradients across rounds.
(ii) Walker-based progressive aggregation, where the walker accumulates gradient snapshots as it visits nodes, gradually constructing an approximate view of global gradient behavior across the nodes.
\begin{itemize}
\item \textit{Associated Overhead:}
Server-based methods offer higher fidelity but require bidirectional device-to-server communication and may violate full decentralization of the learning process. Walker-based approaches maintain decentralization but rely on slow accumulation and may suffer from delayed estimates, especially in large or dynamic graphs.
\end{itemize}
   
\textit{\textbf{6. Estimating Data and Spatial Quality via Walker-Collected Statistics:}} Walkers track statistics such as visit frequency and encountered loss values. These statistics allow the walker to adapt its traversal path on-the-fly based on local observations, enabling fully self-contained and decentralized routing/traversal without the need for external metadata or inter-node coordination.

\begin{itemize}
    \item \textit{Associated Overhead:} Adds minor memory and computation overhead to the walker.
    \end{itemize}

As a result, the choice of importance estimation method (i.e., selecting between the aforementioned methods) introduces trade-offs in terms of: (i) \textit{Computation} (e.g., exact computation of centrality metrics vs. lightweight approximations), (ii) \textit{Communication} (e.g., whether information needs to be exchanged between nodes/walkers), (iii) \textit{Storage} (e.g., the size of the saved information at the nodes/walkers), and (iv)
\textit{Privacy} (e.g., whether label distributions or local performance of nodes/walkers needs to be shared).
In particular, some of the aforementioned strategies can be implemented in a fully local or passive manner (e.g., using the walker's collected statistics), while others may require lightweight communication or coordination between neighboring nodes.
Subsequently, in real-world deployments, the choice of node importance mechanism should align with the level of decentralization, communication budget, and trust assumptions of the environment. For example, in \textit{fully decentralized and privacy-sensitive networks}, local metrics (e.g., local model loss, walkers' visit counters and degree-based centrality) are preferred, 
in \textit{cooperative settings with limited communication}, sharing simple statistics (e.g., class histograms/distributions and validation error) among nodes and walkers may be feasible, and in \textit{infrastructure-assisted networks} (e.g., with periodic access to edge servers or relays), more global metrics (e.g., eigenvector centrality and global information about the heterogeneity of node's datasets) can be periodically obtained (especially for time-varying networks) to guide walkers.

% Motivated by the above results, we conjecture that optimal node traversal in $\mathbb{X}$L requires moving beyond traditional graph exploration and learning approaches—such as Markov chain-based methods that consider only network topology while ignoring node data quality, and FedL/FogL techniques that focus solely on data quality while overlooking dynamic, multi-hop exploration via random walks. A formal treatment of this conjecture is left for future work.

% , we conjecture that optimal node traversal in $\mathbb{X}$L requires a shift from the techniques that are developed in isolation for Markov chains over graphs (i.e.,  only consider network topology) and FedL/FogL (i.e., only consider data quality). Further formalization of this conjecture is left as future work.

% This balanced value of $\alpha$ may not generalize across all environments, but the results underscore the \textit{importance of accounting for both network topology and node data quality}. We therefore conjecture that optimal node traversal in $\mathbb{X}$L demands moving beyond conventional approaches—Markov chain-based methods that focus solely on topology and FedL/FogL techniques that emphasize only data quality.

% \vspace{-3mm}
\subsection{Elastic Learning}
In $\mathbb{X}$L, the walker trains on the data of the node it visits. 
Seemingly counter-intuitive, remaining at every node for extensive training and model refinement (i.e., a greater number of SGD iterations) is \textit{not} always optimal.
% Instead, upon arriving at a node, the walker must determine how long it should train on the node's local dataset.
% Instead, the durations of stay at each node (i.e., the number of SGD iterations) should be a function of its data quality and walker's model state. For example, a walker with a high performing model may conduct only a few SGD iterations on the nodes with low quality data to avoid degrading the performance of its model, while conducting more SGD iterations on the nodes with high-quality data. This simple analogy unveils the importance of a major degree of freedom in $\mathbb{X}$L, which we name \textit{elasticity of local training}.
Instead, the decision on the model training duration should depend on both the quality of the node’s data and the current state of the walker's model. For instance, if the walker's model is already well-trained, performing many updates/SGDs on a node with low-quality (e.g., biased) data can degrade its performance, while more updates/SGDs can be conducted on nodes with high-quality data. 
The need for such dynamic range of update/SGD durations across nodes reveals a key degree of freedom in $\mathbb{X}$L, which we call \textit{elasticity of local training}.

\begin{figure}[t]
\vspace{-6mm}
\centering
{\includegraphics[width=0.7\textwidth]{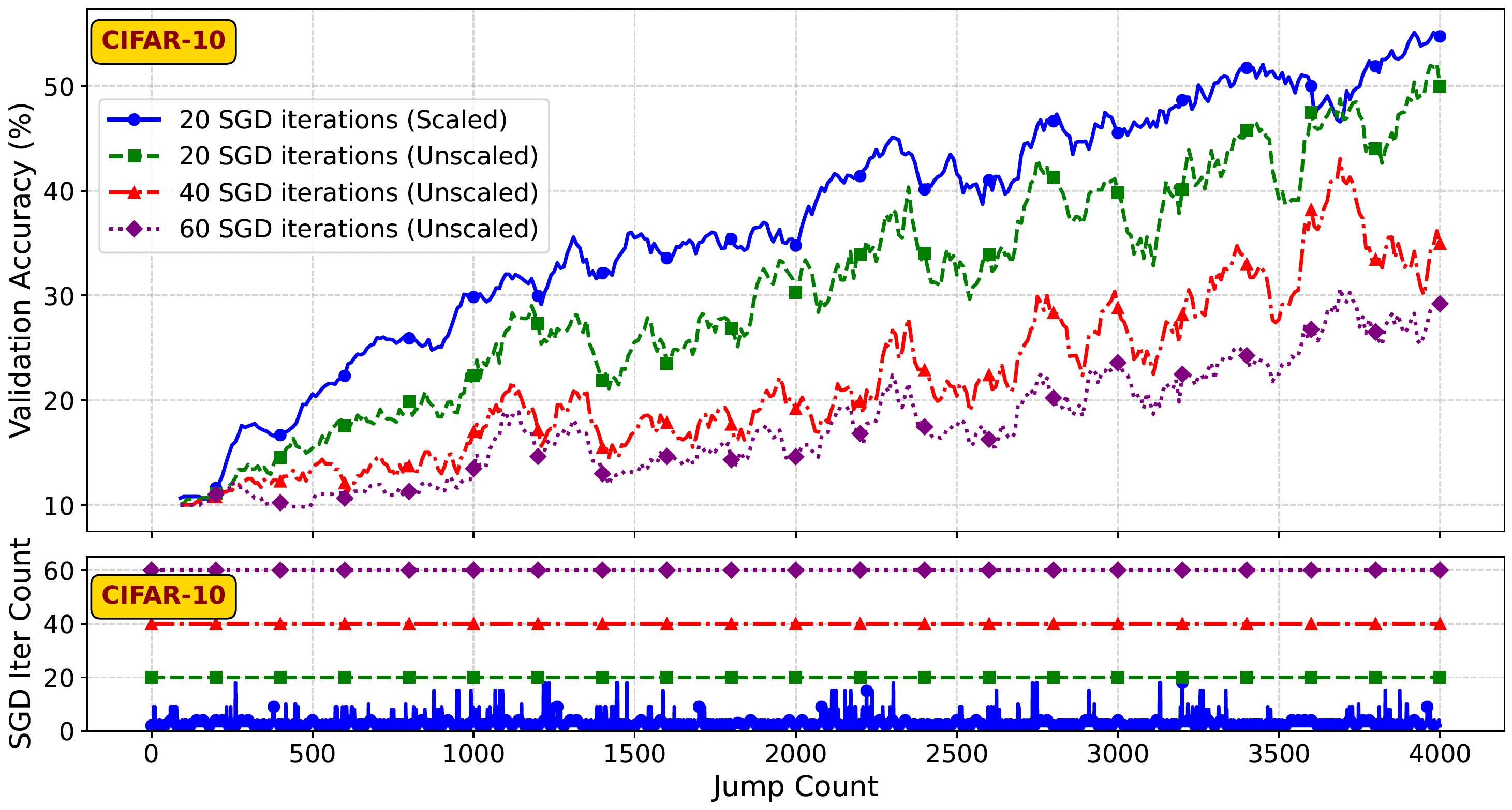}}\\[1mm]
\centering
{\includegraphics[width=0.7\textwidth]{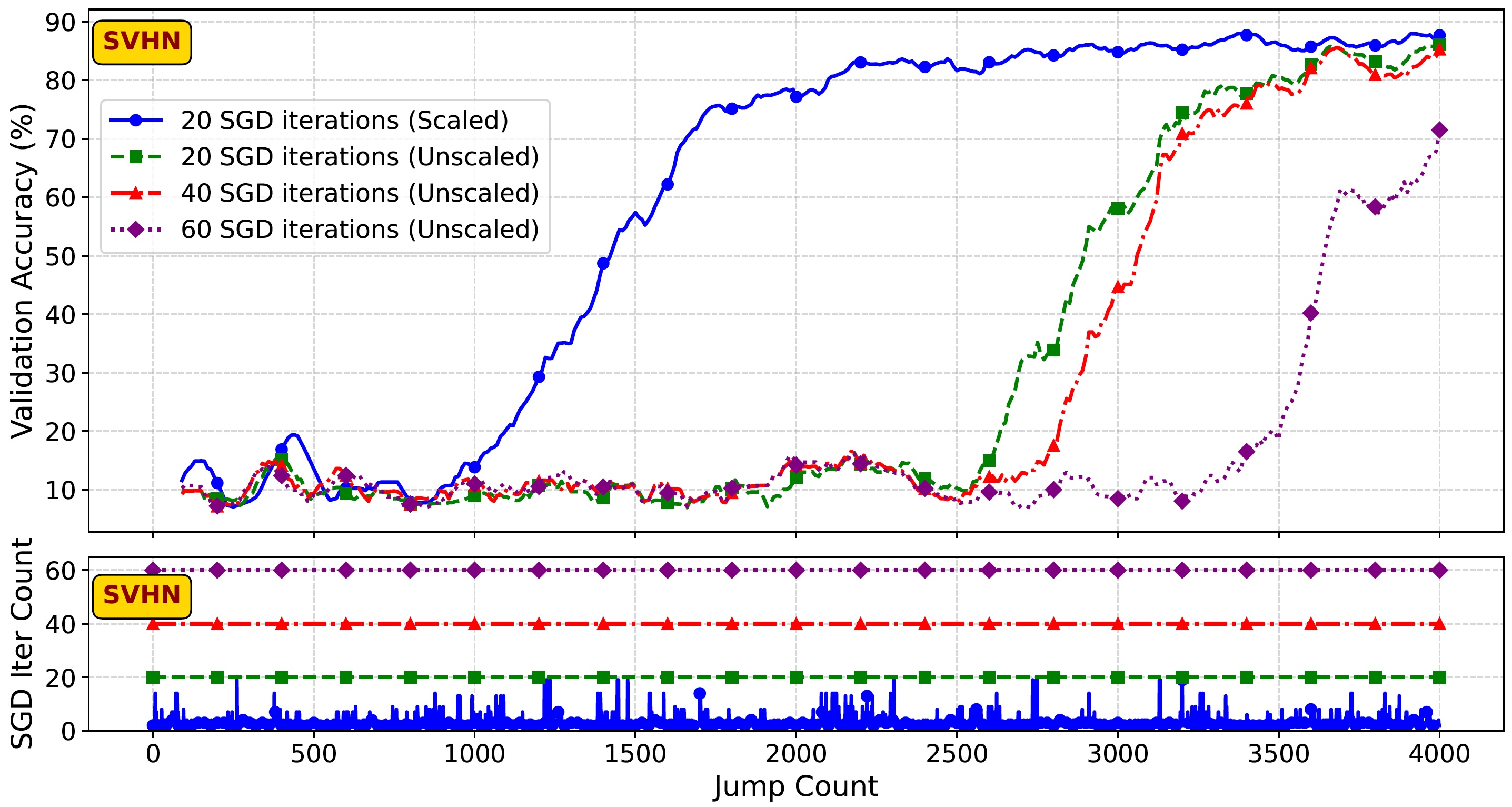}}\\[1mm]
\vspace{-3.5mm}
\caption{Performance comparison between our elastic random walker, where the number of SGD iterations is adaptively scaled based on node data quality, and methods with fixed SGD iteration numbers for CIFAR-10 and SVHN datasets (smoothed using a moving average with a window size of 10). Our elastic walker (blue curves) outperforms all baselines with static SGD iteration numbers (as depicted in the top plot for each dataset) while performing fewer SGD updates (as depicted in the bottom plot for each dataset).}
\label{fig:sgdscale}
\vspace{-1.5mm}
\end{figure}

In Fig.~\ref{fig:sgdscale}, we show the elasticity of local training through a simulation over an RGG with $100$ nodes, $90\%$ of which have only $1$ or $2$  labels of CIFAR-10/SVHN dataset. 
In the three baselines, the walker is assumed to be inelastic and performing a fixed number of SGD iterations at each node ($20$, $40$, and $60$ iterations). Also, we consider an elastic walker (blue curve), where its number of SGD iterations $\ell_u$ at each node $u$ is tuned according to the node's data quality, up to a maximum of $\ell_{\textrm{Max}}=20$ using a sigmoid-like function $\ell_u=\frac{\ell_{\textrm{Max}}}{1 + \exp({-\tau_1\times Q_u})}$, where $\tau_1=10$ is a temperature parameter, and $Q_u$ is a relative measure of the node $u$'s data quality defined by $Q_u=\xi_u \times {d}_u^{\tau_2\times(1-{d_u})}$ with $\tau_2=0.4$ and the rest of the variables ($d_u$,$\xi_u$) are the same as Sec.~\ref{sec:temp}. 
% The formula for assigning the SGD iteration count follows the form of a sigmoid function with its parameters obtained through a line search.
The form of this function allows for a sharp reduction in the number of SGD iterations executed at nodes with low-quality data, while its hyperparameters were chosen experimentally. The results unveil that the elastic walker performs markedly better than the inelastic ones, all while training on significantly fewer data samples. Other formulations for tuning SGD iterations according to node data quality are open for exploration.

\vspace{-1mm}
\subsection{Perception-Aware Random Walk Design}\label{sec:perception}
% In the above example we demonstrated the use of node temperature in tuning the SGD iterations.
% We next unveil the use of the walker's model state (i.e., its accuracy) in tuning the transitions between nodes, where we propose the concept of \textit{random walk perception resolution} in $\mathbb{X}$L. This concept describes how a walker perceives the importance of nodes over time. In the early stages of training, a walker will have a low model accuracy itself and thus a low perception resolution. This means that early on, most nodes will be seen suitable/important for training, as the walker benefits from exploring a broad range of local datasets to improve its initial performance. As the walker's performance/accuracy rises, its perceived resolution of the network will improve. In this situation, the walker becomes more \textit{picky} about selecting the nodes, favoring those with higher data quality and making fewer visits to those nodes with low-quality data to avoid experiencing a performance drop.
% We next introduce the concept of \textit{random walk perception resolution} in $\mathbb{X}$L, which leverages the walker's model state (i.e., accuracy) to further tune its node transitions, migrating from transitioning solely according to node's static importance values explained in Sec. III-A.
We next introduce \textit{random walk perception resolution} in $\mathbb{X}$L, where the walker adjusts its node traversal based on its model accuracy, shifting from static node importance-based traversals (Sec.~\ref{sec:temp}) to accuracy-aware dynamic exploration. Specifically, early in training, with low accuracy, the walker has low perception resolution and treats most nodes as important for broad exploration. As its accuracy improves, the walker’s resolution sharpens, making it more selective, favoring  nodes with high-quality data and avoiding low-quality ones to prevent performance degradation.

% As the walker temperature rises, so does its perceived resolution of the network, and thus its trajectory will be altered such that the walker becomes \textit{picky} about selecting the nodes: it will make fewer visits to cold nodes and always prefer nodes with higher temperatures so that its own temperature does not drop.

% We next shed light on a scenario in which the walkers dynamically alters its one-step probability transition matrices as a function of their own temperatures along with the nodes' temperatures. This may be viewed as an extension to the principles relating to node importance, as this strategy yields a sampling scheme that factors node data quality, spatial quality, and the walker's own temperature.

This further unveils a perception-aware trajectory design, where the walker's trajectory changes over time, suggesting a departure from time-homogeneous to time-inhomogeneous Markov Chains over graphs for $\mathbb{X}$L. To show this,
we redefine the \textit{node importance} as a function of the walker's model accuracy at jump/time $k$ as {\small$\widetilde{\mathfrak{c}}_u^{(k)}=Z_{\mathsf{Inst}}^{(k)} \cdot d_u \cdot \xi_u + (1-Z_{\mathsf{Inst}}^{(k)}) \cdot c_u$}, where compared to the fixed value of $Z$ used in Sec.~\ref{sec:temp}, here $Z_{\mathsf{Inst}}^{(k)}$ (chosen as {\small$Z_{\mathsf{Inst}}^{(k)} \in [Z_{\mathsf{Min}} {=} 0.10~,~ Z_{\mathsf{Max}} {=} 0.75]$} for CIFAR-10 and {\small$Z_{\mathsf{Inst}}^{(k)} \in [Z_{\mathsf{Min}} {=} 0.10~,~ Z_{\mathsf{Max}} {=} 0.85]$} for SVHN) is a dynamic/instantaneous weighting coefficient. We model {\small$Z_{\mathsf{Inst}}^{(k)}$} as a simple linear function of the walker's instantaneous accuracy {\small$\text{A}_{\mathsf{Inst}}^{(k)}$} as 
{\small $Z_{\mathsf{Inst}}^{(k)} = Z_{\mathsf{Min}} + \big((\text{A}_{\mathsf{Inst}}^{(k)} - \text{A}_{\mathsf{Min}}) \cdot (Z_{\mathsf{Max}} - Z_{\mathsf{Min}})\big) \big/ \big(\text{A}_{\mathsf{Max}} - \text{A}_{\mathsf{Min}}\big)$},
where {\small$\text{A}_{\mathsf{Min}}$} and {\small$\text{A}_{\mathsf{Max}}$} (chosen as {\small$\text{A}_{\mathsf{Min}}=0.1$} and {\small$\text{A}_{\mathsf{Max}}=0.85$} for CIFAR-10 and {\small$\text{A}_{\mathsf{Min}}=0.1$} and {\small$\text{A}_{\mathsf{Max}}=0.85$} for SVHN) capture the \textit{estimated} minimum and maximum achievable accuracy of the walker. 
% the value of $\alpha^{\mathsf{Inst}}$ follows 
% $\alpha^{\mathsf{Inst}} = Z_{\mathsf{Min}} + \big((\text{Acc}^{\mathsf{Inst}} - \text{Acc}^{\mathsf{Min}}) \times (Z_{\mathsf{Max}} - Z_{\mathsf{Min}})\big) \big/ \big(\text{Acc}^{\mathsf{Max}} - \text{Acc}^{\mathsf{Min}}\big)$, where 
% $Z_{\mathsf{Min}} = 0.10 $, $Z_{\mathsf{Max}} = 0.85$, $\text{Acc}^{\mathsf{Inst}}$ is the walker's current validation accuracy. Also, $\text{Acc}^{\mathsf{Min}}=0.1$, and $\text{Acc}^{\mathsf{Max}}=0.8$ capture the \textit{estimated} minimum and maximum achievable accuracy of the walker. 
% This simple linearly adjusted choice of tuning $\alpha^{\mathsf{Inst}} $ leads to jumping to only those nodes with high quality datasets as the walker's own performance increases.  
% Tuning the selection of node $i$ based on the above $\widetilde{\mathfrak{c}}_u$, which by itself depends on the accuracy of the random walker's model, makes the transition probability matrix of the walker time-varying, leading to a time-inhomogeneous Markov Chain over the network graph.
Subsequently, at each node, the transition probability of the walker to the neighboring nodes is defined proportional to their \textit{time-varying} importance values ($\{\widetilde{\mathfrak{c}}_u^{(k)}\}$), resulting in a time-inhomogeneous Markov Chain over the network graph.
Inspecting Fig. \ref{fig:singlecrawl}, the above-described perception-aware walker (black curve) exhibits a superior performance. We leave exploration of more advanced formulations of perception-aware trajectory design as future work.

% Within $\mathbb{X}$L, we propose the concept of random walk perception resolution. This describes how a single walker perceives the quality of its local neighboring nodes over time. In the early stages of training, a walker will have a cold temperature and thus a low perception resolution. This means that most nodes will be viewed as viable for training early on such that the walker will benefit from visiting a broad range of local datasets to enhance its early performance trajectory. As the walker temperature rises, so does its perceived resolution of the network, and thus its trajectory will be altered such that fewer visits to cold nodes will occur and its own temperature may be better preserved.
\begin{figure}[t]
\vspace{-2mm}
\centering
\includegraphics[width=0.8\textwidth]{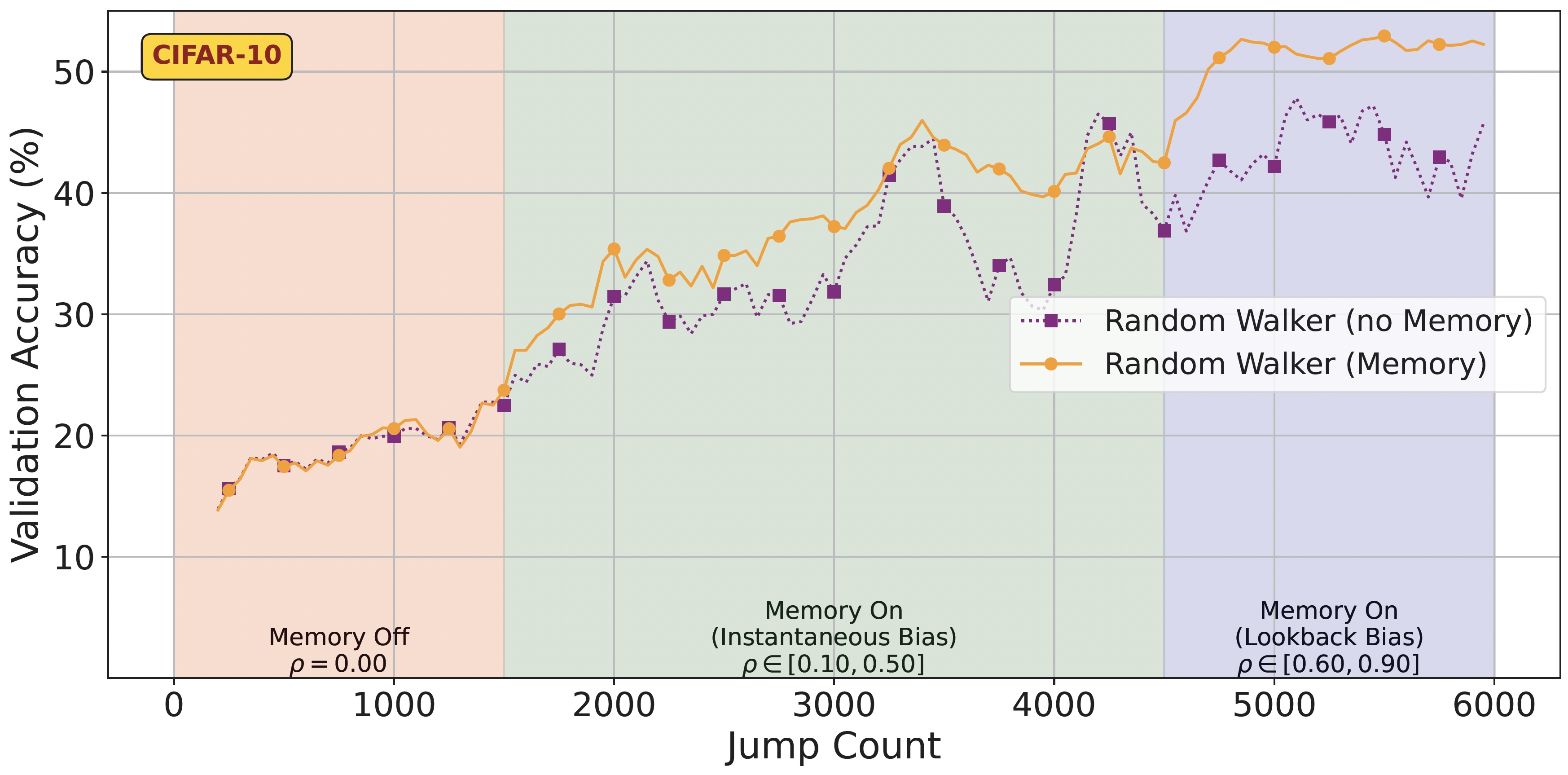} \\[1mm]
\includegraphics[width=0.8\textwidth]{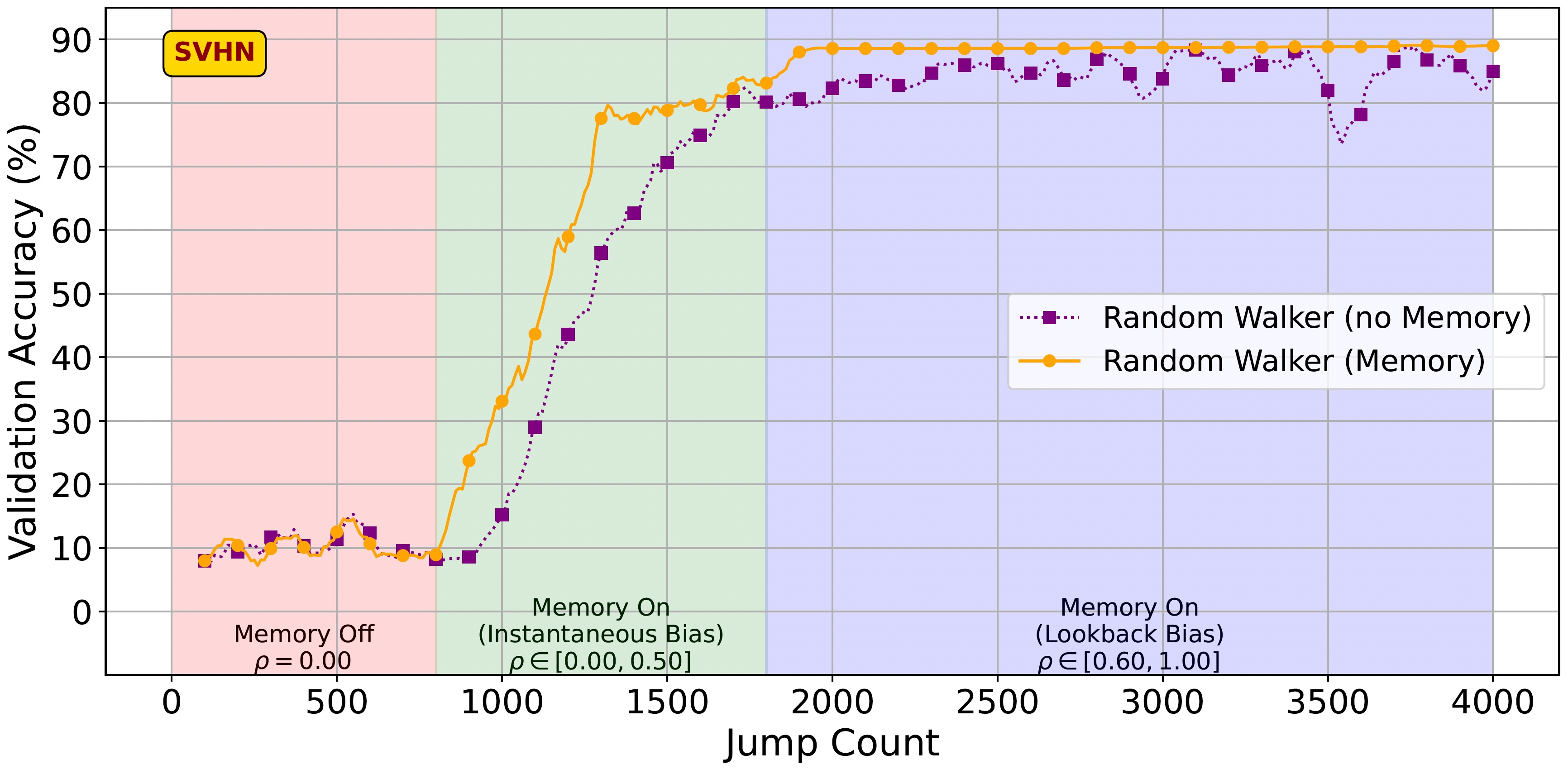}
\vspace{-3.5mm}
\caption{Performance comparisons (smoothed using a moving average with a window size of 10) between a memory-enabled random walker and a memoryless walker, both using the same node traversal strategy for CIFAR-10 and SVHN datasets. Initially, memory is disabled (red region) to accelerate early training. As training progresses, memory influence gradually increases (green and purple regions), helping mitigate model bias and yielding notable gains -- especially in the mid to late stages of training.}
\label{fig:memory}
\vspace{-1.5mm}
\end{figure}

\vspace{-1mm}
\subsection{Random Walks with Flexible Memory}

A key challenge for a random walker is forgetting patterns learned from previously visited nodes as it continues updating its model during new node visits, mirroring the \textit{catastrophic forgetting} phenomenon in ML.
% To mitigate this, we propose flexible-memory walkers carrying two model instances: a stale model (SM) and an instantaneous model (IM). Initially, the walker synchronizes the SM with the IM. Upon visiting a node, the IM is trained on the node’s data. After several jumps, the walker averages the SM and IM, ensuring that the IM does not forget the learned patterns in the previously visited nodes.
To address this, we propose flexible-memory walkers that maintain two model instances: a stale model (SM) and an instantaneous model (IM). Initially, both models are synchronized. Upon visiting a node, only the IM is updated using its local data. 
% After a number of jumps, the walker averages the SM and IM into a model that is used to synchronize both SM and IM, allowing the IM to retain knowledge from earlier visits and mitigating catastrophic forgetting.
Afterwards, the walker averages the SM and IM to produce an aggregated model and uses it to synchronize both models. This process helps the IM retain knowledge from earlier visits.
% , thereby mitigating catastrophic forgetting.

To verify this strategy, we conduct a simulation on an RGG with $500$ nodes, $90\%$ of which possessing $1$ or $2$ of the CIFAR-10/SVHN labels. 
The walker is presumed to have an IM, {\small $\bm{w}^{\mathsf{IM}}$}, and an SM, {\small$\bm{w}^{\mathsf{SM}}$}. After arriving at each node, {\small$\bm{w}^{\mathsf{IM}}$} is first updated via conducting $5$ SGDs on the node's data. 
Then, the IM is updated via aggregation with the SM as {\small$\bm{w}^{\mathsf{IM}}\leftarrow  (1-\rho)\bm{w}^{\mathsf{IM}}+\rho \bm{w}^{\mathsf{SM}}$}; the updated IM is further used to refine the SM as {\small$\bm{w}^{\mathsf{SM}} \leftarrow\bm{w}^{\mathsf{IM}}$}. Here, $\rho\in[0,1]$ captures the importance of the SM. The performance improvements for the memory-enabled walker are notable in Fig.~\ref{fig:memory}, where in the early stages (red area), memory was disabled to accelerate the initial learning and its effect was gradually increased in stages (green and purple areas). For demonstrations, the value of $\rho$ at each jump was obtained through a line search across the interval  written in each shaded region in Fig.~\ref{fig:memory}. A more generalized formulation for memory-enabled walker is left for future work. 
% Values at the low end of the spectrum have a preference for the IN-ML during aggregation, and the higher end of the range favors the ST-ML. As the ST-ML becomes more stable in the later stages, a higher preference is given to it during the periodic recall process such that learning is stabilized.

% Building on this, we plan to optimize the aggregation of the IN-ML and ST-ML by adjusting the frequency and methodology based on data heterogeneity. Additionally, we will explore extending the memory size with multiple ST-MLs, each representing different levels of staleness, to further stabilize learning across varying data quality nodes.

% \begin{tcolorbox}[colback=darkyellow2, colframe=salmon2, title=\textbf{Future Research Directions}]
\begin{tcolorbox}[colback=white,colframe=black,colbacktitle=black,
title=\centering \small Box 3: \textbf{Future Research Directions},fonttitle=\bfseries]
The exploration of $\mathbb{X}$L with a single walker opens up various future directions, discussed below.

\begin{itemize}[leftmargin=2.2mm]
\item \textit{\textbf{Connection to Graph Signal Processing (GSP):}}  We suggest exploring node importance metrics for $\mathbb{X}$L inspired by core signal processing methods, such as GSP~\cite{leus2023graph,ortega2018graph,wang2021gene,li2021graph,8931013}.  GSP incorporates graph topology into node signal representation through eigen-decomposition of the graph Laplacian. Developing a unified metric for node importance in $\mathbb{X}$L via GSP calls for (i) defining appropriate signals for the nodes based on their data, (ii) studying interpretations of eigen-decomposition for $\mathbb{X}$L, and (iii) leveraging GSP-inspired learning mechanisms (e.g., graph neural networks) for $\mathbb{X}$L.
\item \textit{\textbf{ML Convergence under Time-Varying Transition Matrices:}} Characterizing the ML convergence under time-varying transition matrices is crucial and will involve obtaining ML convergence bounds for convex and non-convex loss functions, considering the impacts of random walk trajectory on its model convergence.  To motivate this line of analysis, we will later present a set of preliminary convergence results for $\mathbb{X}$L in the subsequent sections of this paper.

\item \textbf{\textit{Comparison with Traditional Markov Chain Designs:}} 
Analyzing how the dynamic design of transition matrices differs from time-homogeneous Markov chains considering traditional  metrics (e.g., covering time and mixing time~\cite{levin2017markov,avin2018cover}) is interesting. This involves demonstrating `how' and `why' traditional random walk visiting patterns may lead to inferior performance when compared to $\mathbb{X}$L-specific strategies.
\item \textbf{\textit{Resource-Aware $\mathbb{X}$L:}} To derive the optimal walker trajectory and training durations at  nodes, future work can extend our generic formulation in Section~\ref{sec:Formulation} and formulate optimization problems with competing objective functions, comprising the walker's ML performance, communication/traversal latency, and energy consumption, under realistic transition costs (e.g., energy and delay) between nodes.

\item 
\textit{\textbf{Trust and Reliability in D2D/P2P Settings:}}  
   The issue of trustworthiness becomes complex in decentralized and D2D/P2P networks, as highlighted in~\cite{tao2024federated,letaief2021edge}. Since $\mathbb{X}$L relies on local training and opportunistic model transfers, it is similarly vulnerable to:
(i) adversarial nodes (i.e., devices that inject misleading data or models to manipulate the walker's model convergence or induce bias), and
 (ii) unreliable model updates (i.e., model updates from nodes with noisy, low-quality, or stale data that can harm the walker's model stability). To mitigate this, research can explore $\mathbb{X}$L with \textit{local validation filters} (i.e., mechanisms that verify the utility or correctness of  model updates of each node  before accepting them using local validation sets held at the random walker's memory),
\textit{lightweight reputation mechanisms} (i.e., scoring systems that track the trustworthiness of nodes based on their historical model contributions), and \textit{secure transition protocols} (i.e., cryptographic or noise-injection schemes that ensure robustness and privacy of the walker's model during traversal between nodes).

% To derive optimal walker trajectories and training durations at different nodes, future work can formalize this challenge as an optimization problem balancing competing objectives—such as the walker's ML performance, jump latency, and energy consumption—while accounting for realistic transition costs (e.g., energy and delay) between nodes.
\end{itemize}

\end{tcolorbox}

\begin{figure}[t]
\vspace{-3mm}
\centering
{\includegraphics[width=0.6\textwidth]{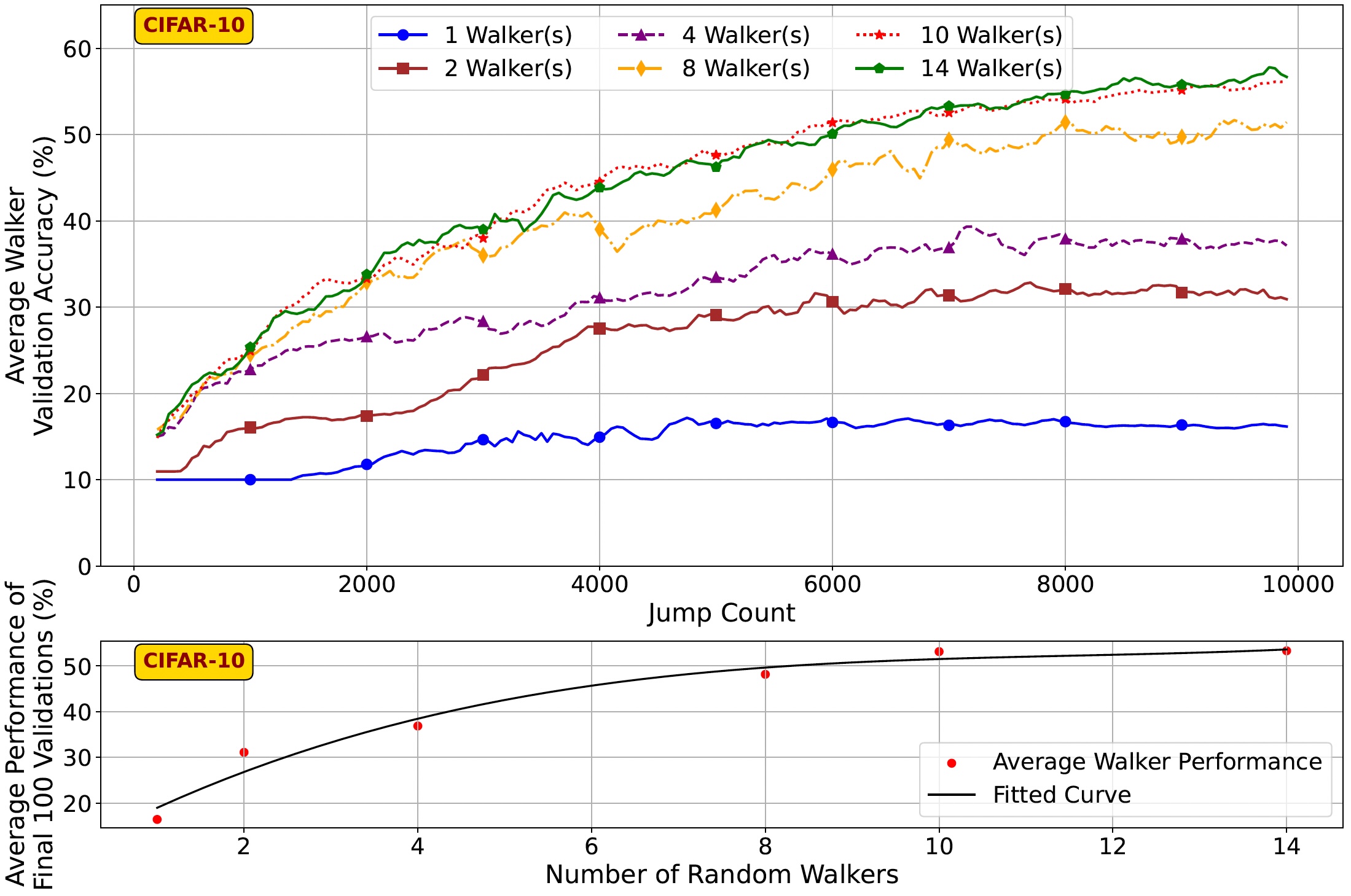}}\\[1mm]
\includegraphics[width=0.6\textwidth]{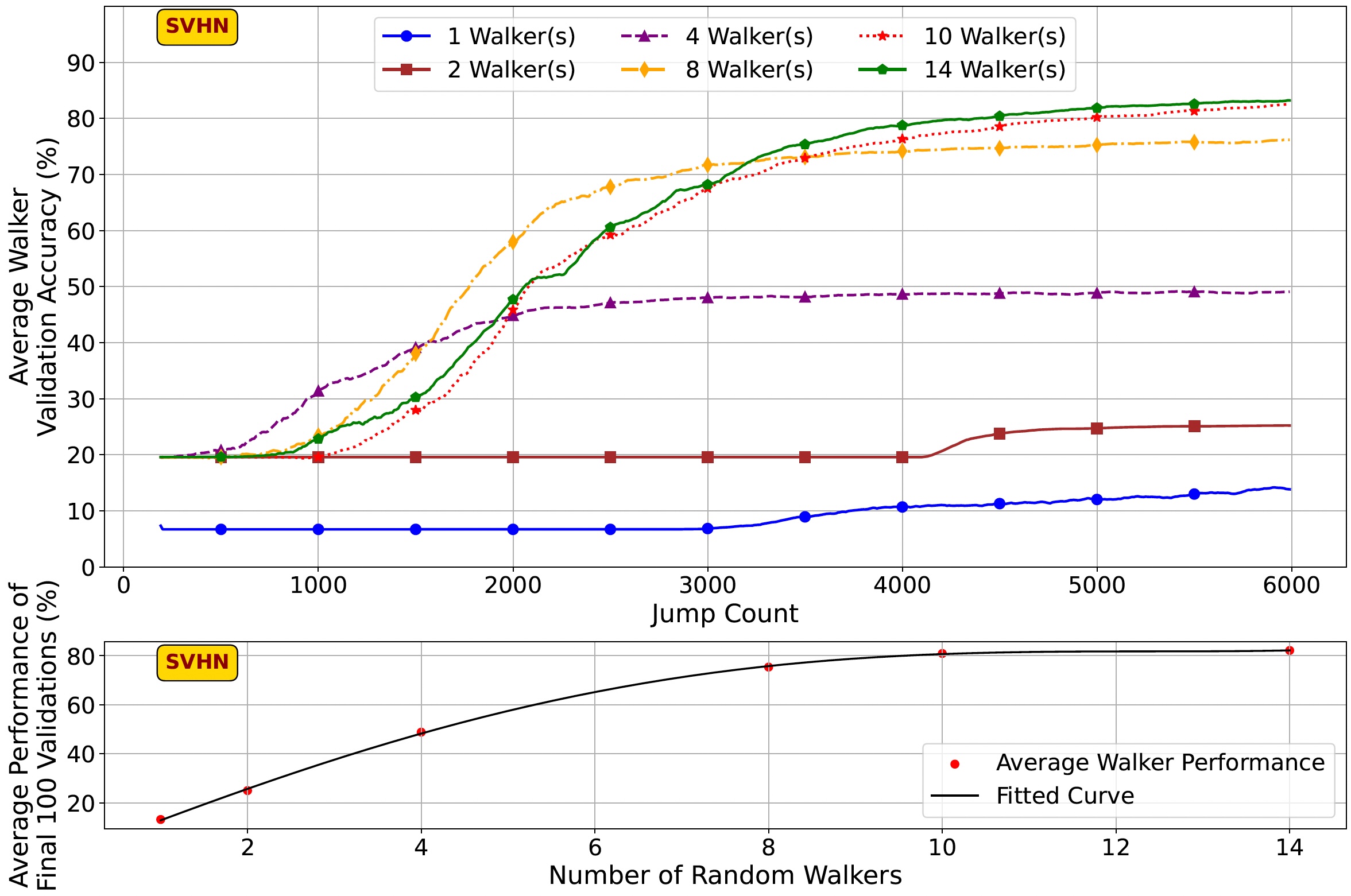}
\vspace{-2.5mm}
\caption{Impact of the number of random walkers on model convergence for CIFAR-10 and SVHN datasets. The top plot of each  dataset shows the instantaneous average performance across all walkers, where increasing the number of walkers initially boosts performance; however, beyond a certain point (e.g., from 10 to 14 walkers), the improvements become marginal. The bottom plot of each  dataset presents the walkers' final accuracies, highlighting an initial sharp gain followed by diminishing gains as more walkers are deployed.
}
\label{fig:basic_collision}
% \vspace{-2.5mm}
\end{figure}

\vspace{-1mm}
\section{$\mathbb{X}$-Learning with Multiple Random Walkers}\label{sec_4}
\noindent Expanding $\mathbb{X}$L, we consider scenarios involving multiple random walkers, opening up new research opportunities. Here, the dynamics of learning and the interaction between walkers add new layers of complexity. For instance, multiple walkers can have overlapping traversal patterns and potential collaborative or competitive dynamics as detailed next. 
% In the following, we first discuss collaborations among walkers, exploring whether walkers should communicate with each other, and if so, what information should they exchange. 
% We then introduce a series of open research directions.

% and propose treating the walkers as a collective to optimize the learning process as a whole.
\vspace{-1mm}
\subsection{Collaboration Among Walkers}
 In deployments of $\mathbb{X}$L involving multiple walkers, collaboration between walkers becomes a key degree of freedom that affects convergence speed, redundancy avoidance, and model robustness. Due to the decentralized and asynchronous nature of the framework, cooperation can be designed with a high degree of flexibility. Below, we outline the major considerations across three dimensions that leave a pronounced impact on $\mathbb{X}$L: \textit{how to cooperate?, when to cooperate?, and whom to cooperate with?}
\paragraph*{How to Cooperate?}
Walkers can engage in various forms of interaction when they encounter one another or operate within nearby regions of the communication graph. Cooperation mechanisms include:
\textit{(i) Model Averaging:} When two walkers meet at the same node or adjacent nodes, they may exchange and average their carried models to synchronize their models.
\textit{(ii) Role Specialization:} Walkers can be assigned complementary roles (e.g., one walker may focus on exploring under-visited nodes, while another explores high-loss areas/nodes). In this setting., metadata such as walkers' visit counters can be exchanged among them to coordinate their trajectories.
\textit{(iii) Meta-Information Sharing:} Walkers may share statistics among each other, such as observed loss values at nodes and nodes' trustworthiness, to improve each others' decision-making in a decentralized manner.

\paragraph*{When to Cooperate?}
The timing of collaboration events may be:
\textit{(i) Opportunistic:} Cooperation occurs when walkers are co-located at the same node or at neighboring nodes.
\textit{(ii) Periodic:} Walkers synchronize their states every $S$ steps by visiting a set of predefined nodes, where the value of $S$ is also predefined.
\textit{(iii) Conditional:} Cooperation is triggered by conditions such as the duration of time in which the walkers have not met each other.

\paragraph*{Whom to Cooperate With?}
Walkers may select collaboration partners based on:
\textit{(i) Proximity:} Physical co-location or when the walkers are only one-hop away and can exchange information is a natural basis for interaction.
\textit{(ii) Statistical Similarity:} Walkers exploring regions/nodes with dissimilar or non-overlapping data distributions may benefit from model aggregation, as it can help debias their local models and accelerate convergence in environments with non-iid data across the nodes. 
\textit{(iii) Trustworthiness:} Cooperation can be restricted to walkers with a verified history of trust (e.g., walkers instantiated by the same entiriy/company across the nodes), enabling trust-aware interactions in adversarial settings.

We take the first step towards addressing the above dimensions through studying the interactions between the walkers. 
Drawing from physics concepts on interacting particles~\cite{liggett1985interacting,de1986reaction,kipnis2013scaling,jin2020random,schinazi1996interacting} and epidemic spreading~\cite{colizza2008epidemic,hasegawa2016outbreaks,prakash2012spotting,salehi2015spreading,brodka2020interacting,pare2018analysis}, we explore cooperation between random walkers through information transfer.
% Mirroring concepts in physics relating to interacting particle systems \cite{aldous2013interacting} and the spread of epidemics \cite{colizza2008epidemic}, we consider cooperation between random walkers through the transfer of information between them.
Specifically, when two walkers meet at a node, they ``collide" and aggregate their models (e.g., via weighted averaging their models based on the number of seen data points). To show this, we simulate  $\mathbb{X}$L over a caveman graph with $10$ cliques, each containing $100$ nodes, where each clique is dominated by the presence of a single label of CIFAR-10/SVHN. 
All walkers select their next nodes to jump uniformly at random; however, every $10$ jumps, each pair of walkers aggregate by visiting a predetermined node.
% We improve on this (naive) form of aggregation shortly.
%sampling, at visit each other once a wall-clock hits (after 10 jumps) at a predetermined node in the network, which is the most naïve version of aggregation that we will later improve.
% Our findings in  Fig.~\ref{fig:basic_collision} suggest that collisions, and information exchange, can be highly beneficial. Further, they 
% show the diminishing reward of the number of deployed walkers. Understanding this point of diminishing rewards and the network-specific factors that affect it are exciting research directions.
Fig.~\ref{fig:basic_collision} shows that collisions and information exchange are beneficial. Further, it shows the diminishing reward of the number of deployed walkers. Investigating the factors influencing this diminishing reward is an exciting research direction.

% \allan{Our preliminary findings in Fig.~\ref{fig:basic_collision} suggest that walker collisions and the resultant information exchange can be highly beneficial in improving the rate of convergence for the overall system. The network-wide rate may be increased through the introduction of additional interacting walkers, though larger amounts of walkers yield diminishing performance returns. Understanding this point of saturation and the network-specific factors that affect it are an open research direction.}

% However, on large scale networks with a small amount of walkers, these types of collisions would occur rarely, and the benefits of information transfer would be diminished as walkers continue to learn independently for the most part. In these situations, it is worth considering methods of increasing the degree of interactivity between walkers.

% \subsection{Future Research Directions}
\begin{figure}[t]
% \vspace{-6mm}
\centering
{\includegraphics[width=0.7\textwidth]{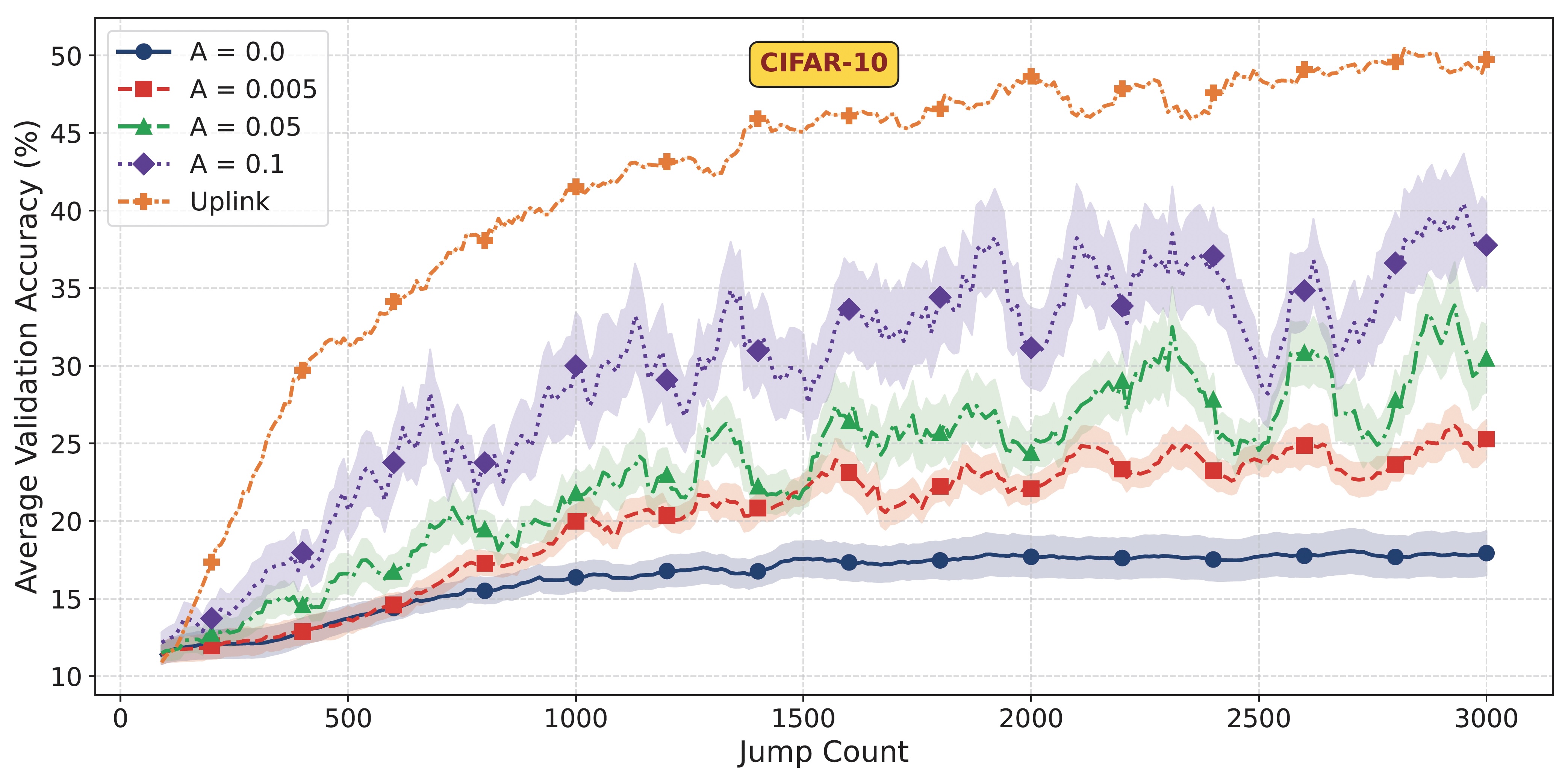}}\\[1mm]
{\includegraphics[width=0.7\textwidth]{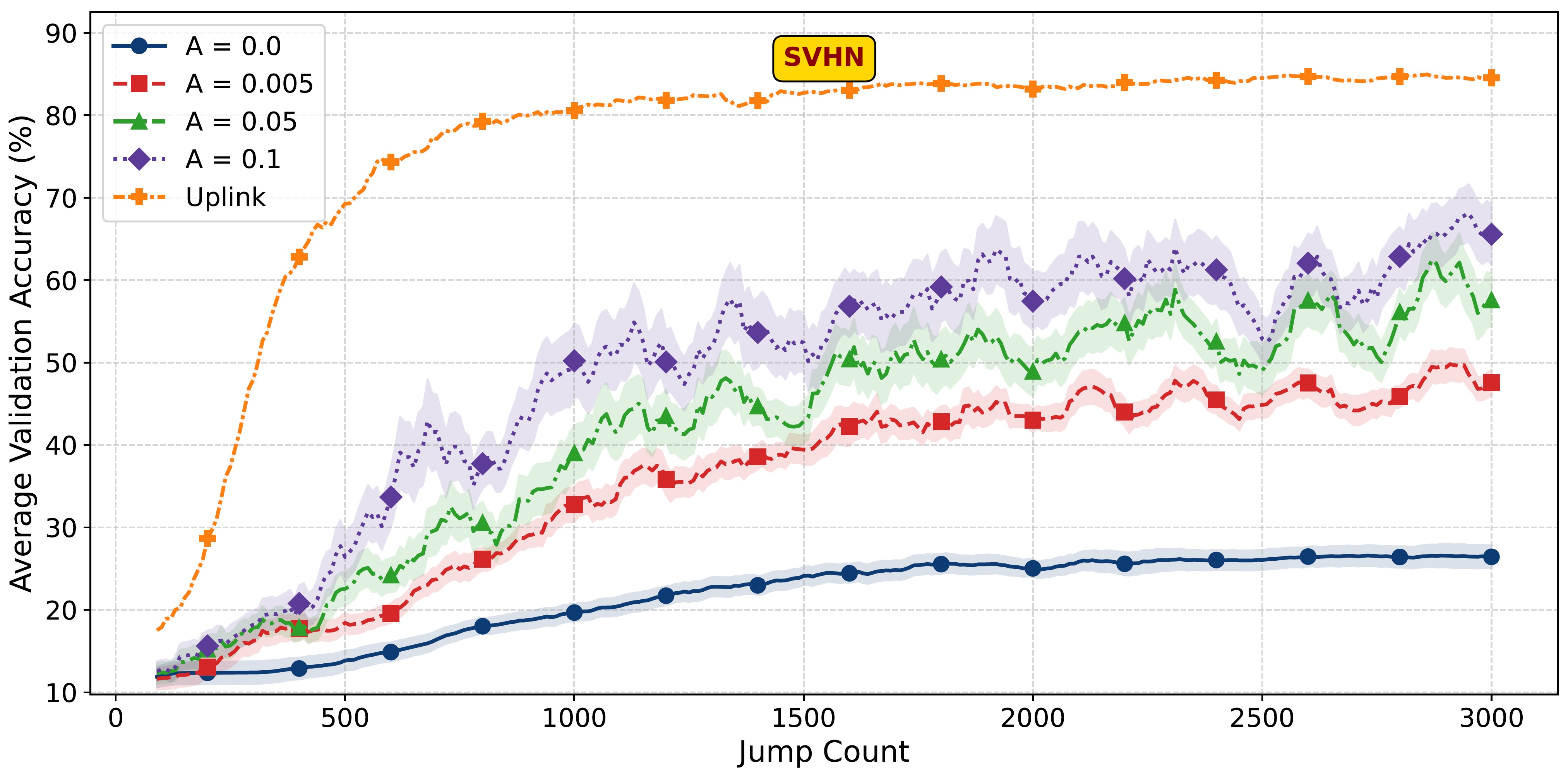}}
\vspace{-3.5mm}
\caption{Impact of inter-walker attraction on model performance (depicted using a moving average with a window size of 10) for CIFAR-10 and SVHN datasets (shaded regions depict $\pm$25\% of standard deviation of performance across the walkers). As the attraction between walkers increases (i.e., higher values of $A$, leading to shorter inter-collision intervals), the walkers interact more frequently, which results in mitigating model bias and an improved performance. For comparison, the orange curve illustrates the (hypothetical) upper bound achieved by centrally aggregating the walkers' models after every jump.}
\label{fig:attraction}
\vspace{-1.5mm}
\end{figure}
\vspace{-1mm}
\subsection{Attraction/Repulsion Between Multiple Random Walkers}

We next aim to control \textit{the frequency of collisions between walkers}.
% Consider a scenario in which two walkers starting at the same node embark on divergent paths; as they continue sampling and training at nodes along each of their paths, they may begin to develop unique biases and distinct models. 
Consider a scenario in which two walkers start at the same node and follow divergent paths. These walkers may develop  distinct (biased) models as they traverse and train across different nodes.
% When they do eventually collide, the aggregation of their models may suffer worsened performance. Inspired by the physics of charged particles, we consider mechanisms to periodically \textit{pull or attract} walkers to each other, enabling more frequent aggregations and reducing model bias.
When walkers eventually collide, the aggregated model can potentially exhibit a low performance if their models are highly biased/separated.
Drawing inspiration from charged particles, we propose mechanisms to periodically \textit{pull/attract} walkers, enabling more frequent aggregations and reducing bias.
% The longer two models traverse the network without a collision, the stronger their polarity and thus attraction. This increasing attraction between the ``charged" walkers will influence their node traversal processes such that they are drawn towards each other for an eventual collision. We can also conceptualize a mechanism where walkers lose their attraction momentarily upon collision, akin to charged particles discharging --- this avoids redundant aggregations between them. 
Essentially, the longer two walkers travel without colliding, the stronger their \textit{polarity} and thus \textit{attraction}, pulling them towards each other for eventual aggregation. After a collision, walkers briefly lose attraction, preventing redundant aggregations.

% This process will repeat as the walkers rebuild their mutual attraction.

%Using centralized FedL uplink as an upper bound, 

To demonstrate, we set up a caveman graph comprising $64$ nodes grouped into $8$ cliques. In this scenario, $8$ random walkers are deployed, one in each clique and their transition matrices are tuned such that they solely explore their own cliques. The level of attraction  between the walkers was dictated by a configurable parameter $A$,  higher values of which lead to more frequent collisions (i.e., shorter inter-collision intervals). In particular, we defined the attraction between each pair of walkers $r$ and $r'$ at jump instant $k$ via the probability {\small $p_{r\leftrightarrow r'}^{(k)}=\min\{1,\exp{(A \cdot \mathcal{T}_{r,r'}^{(k)})}\}$}, where {\small $\mathcal{T}_{r,r'}^{(k)}$} represents the  elapsed time since the walker pair collided. 
For each walker $r$, its transition matrix is extended such that attractions between $r$ and all other walkers $r'$ are included, i.e., {\small$p_{r\leftrightarrow r'}^{(k)}, \forall r'$}. 
 The results are presented in Fig.~\ref{fig:attraction}: as parameter $A$ increases and collisions become more frequent, the walkers' performance improves, approaching a hypothetical best-case scenario with centralized aggregation of walkers after every jump (labeled as `uplink').
% These intuitions are verified via the results in Fig.~\ref{fig:attraction}. 

% \ali{bring the exact formulations!} \allan{Added formulation / discussion here}

% \subsection{Redefining the Notion of Centrality and Perception}

% \begin{tcolorbox}[colback=darkyellow2, colframe=salmon2, title=\textbf{Future Research Directions}]
\begin{tcolorbox}[colback=white,colframe=black,colbacktitle=black,
title=\centering \small Box 4: \textbf{Future Research Directions},fonttitle=\bfseries]
% \subsection{}
The above exploration of $\mathbb{X}$L with multiple random walkers opens up a series of future directions:
% In particular, we solely focus on introducing a series of novel concepts, the formalization of which is left as future work.

\begin{itemize}[leftmargin=2.5mm]
\item \textbf{\textit{Incorporating Caching:}} The need for physical collisions between walkers can be reduced by assuming that walkers leave a \textit{trace}/\textit{footprint} of their ML models on nodes. This trace gradually fades over time to signify the staleness of the left ML model. If a new walker arrives at a node where a trace is still fresh/reliable, it can aggregate its ML model with this trace, while accounting for the age of the trace. 
\item \textit{\textbf{From Single-Task to Multi-Task Learning:}}
The walkers can potentially train different ML tasks, introducing multi-task learning~\cite{chen2023fedbone,zhong2022over,mills2021multi,marfoq2021federated,zeng2021multi} into $\mathbb{X}$L. This opens up exploration of (i) aggregation between the walkers based on their task similarity, (ii) optimal walker-to-task allocation, and (iii) resource optimization (e.g., number of SGD iterations of walkers).    
% \item \textbf{\textit{Introducing Entangled Random Walker Swarms:}} We envision the concept of ``entangled random walker swarms", where walkers are organized into swarms that move closely, following a \textit{macro trajectory}, with frequent intra-swarm aggregations and less frequent inter-swarm aggregations. This hierarchical scheme aims to optimize the balance between frequent aggregations and resource overhead.

\item \textbf{\textit{Security of Random Walkers and their Byzantine Resilience:}} We suggest studying how exploiting the parameters of random walkers during node traversal (e.g., via implementing an inference attack) in different network neighborhoods (e.g., areas with high vs. low data quality) can reveal information about the data of their previously visited nodes. Additionally, exploring the walkers' robustness against Byzantine attackers (e.g., malicious random walkers that intentionally aggregate with legitimate walkers to degrade their performance) is an interesting direction.

% \item \textbf{\textit{Random Walkers with Shaded Visible Regions:}}
%     % In our initial simulations above, walkers can attract each other from any location in the network, which is challenging to implement in decentralized systems. To address this, 
% We suggest studying the ``visibility radius", within which a walker can attract others, e.g., via broadcasting a beacon. 
% If the visibility radius is set to one, walkers can only attract others from neighboring nodes. Increasing the visibility radius can enhance the walkers' ML models; however, a larger visible region will increase the communication overhead (e.g., beacon transmissions). 
\item \textbf{\textit{Redefining the Notion of Centrality and Perception:}} Incorporating multiple random walkers necessitates reevaluating node centrality and walkers’ perceptions defined in Sec.~\ref{sec_3}. These revamps should be done to avoid redundancy in exploration by assigning different centrality values to a node across walkers. 

\item \textbf{\textit{Dynamic Communication Graphs and Mobility Models:}} A key direction for advancing $\mathbb{X}$L lies in extending its applicability to time-varying communication graphs induced by realistic mobility and connectivity patterns as many potential deployment environments of $\mathbb{X}$L, such as vehicular networks, drone swarms, and peer-to-peer mobile systems, feature highly dynamic network topologies. For example, in  \textit{vehicle-to-vehicle (V2V)} networks~\cite{ngo2023cooperative,abdisarabshali2023decomposition}, graph edges are formed opportunistically based on proximity, road layout, and node communication range.
To support such settings, future extensions of $\mathbb{X}$L can integrate mobility-aware random walk strategies over evolving graphs, where the walker's transitions are shaped by both spatial-temporal constraints and ML task utility (e.g., model loss, uncertainty). To this end, mobility models, such as Manhattan Grid, highway flow, or SUMO-based simulation traces, can be used to guide the construction of these dynamic graphs~\cite{tang2021comprehensive,liu2024dnn}. Additionally, investigation of the deployment of \textit{multiple walkers} to explore disjoint mobility corridors (e.g., north-south and east-west roadways in Manhattan models) is of particular interest as such deployments enable parallelized data processing across the nodes.
%     \item \textbf{\textit{Expanding and Contracting Random Walkers:}}
% We suggest \textit{the adaptive expansion and contraction of random walkers}. In homogeneous network segments, a walker can expand by creating multiple replicas to explore non-overlapping paths rapidly. Once exploration is complete, these replicas contract back into a single walker. This approach leverages parallelization for expediting the learning. Determining optimal timing and conditions for expansion/contraction, as well as the number of replicas are of particular interest.
\end{itemize}
\end{tcolorbox}

% \section{Generic Techniques for $\mathbb{X}$-Learning}

% \noindent To address the challenges posed by data heterogeneity and network topology irregularities with single and multiple random walkers, we propose enhancing the models with several generic and complementary features.

\section{Convergence Analysis}
 In this section, we provide preliminary convergence results for special cases of $\mathbb{X}$L with single random walker, aiming to shed light on its convergence behavior and to catalyze future research on convergence properties across various extensions (e.g., $\mathbb{X}$L with multiple random walkers, $\mathbb{X}$L with memory, attraction-based interactions, etc.). 
To describe the learning behavior of the random walker, presume that the random walker is at an arbitrary node $u$ at jump/step $k$. At this step, the walker performs $\ell^{(k)}$  SGD iterations on the local dataset of node $u\in\mathcal{U}$, denoted by $\mathcal{D}_u$ with size ${D}_u=|\mathcal{D}_u|$ (the size of the global dataset defined in Sec.~\ref{sec:Formulation} will then be given by ${D}=\sum_{u\in\mathcal{U}}D_u$). Let $\bm{\omega}^{(k)}$ denote the model held by the walker at the beginning of step $k$, and let $\bm{\omega}_{u}^{(k),\ell}$ denote the local model at node $u$ after $\ell$ local SGD iterations at step $k$ (initially, $\bm{\omega}_{u}^{(k),0}=\bm{\omega}^{(k)}$). The evolution of this local model for $\ell=1,\dots,\ell^{(k)}$ is then given by:
\begin{equation}\label{eq:updaterule}
    \bm{\omega}_u^{(k),\ell} = \bm{\omega}_u^{(k),\ell-1}-\eta_k \sum_{d\in \mathcal{B}^{(k),\ell}_{u}} \hspace{-3mm} {\frac{\nabla  f(\bm{\omega}^{(k),\ell-1}_{u},d)}{{B}_{u}}},
\end{equation}
where $\mathcal{B}^{(k),\ell}_{u}$ denotes the mini-batch of data with size ${B}_{u}$, randomly sampled at SGD iteration $\ell$ from the local dataset $\mathcal{D}_u$, and $\eta_{k}>0$ represents the SGD step-size/learning-rate. Also, as defined in Sec.~\ref{sec:Formulation}, $f(\bm{\omega},d)$ is the \textit{loss function} that quantifies the error of parameter $\bm{\omega}\in\mathbb{R}^{M}$ for each data point $d \in \mathcal{B}^{(k),\ell}_{u}$. Let $\widehat{\lambda}_{u}^{(k)}\in \{0,1\}$ denote a binary random variable indicating whether the random walker visits node $u$ at step $k$ (i.e., $\widehat{\lambda}_{u}^{(k)}=1$ indicates the visit; otherwise $\widehat{\lambda}_{u}^{(k)}=0$), where we have  $\sum_{u\in\mathcal{U}}\widehat{\lambda}_{u}^{(k)}= 1$,~$\forall k$.
After completing the $\ell^{(k)}$ local SGD iterations, the walker computes its cumulative gradient as $\widetilde{\nabla \mathfrak{L}}_{u}^{(k)} = \sum_{\ell=1}^{\ell^{(k)}} \sum_{d\in \mathcal{B}^{(k),\ell}_{u}}  {\frac{\nabla  f(\bm{\omega}^{(k),\ell-1}_{u},d)}{{B}_{u}}}$ and then obtains its next model as
\begin{equation}
    \bm{\omega}^{(k+1)} = \bm{\omega}^{(k)} -\eta_k \sum_{u\in \mathcal{U}}\widehat{\lambda}_{u}^{(k)} \widetilde{\nabla \mathfrak{L}}_{u}^{(k)}.
\end{equation}

Considering the above setup, we make a set of standard assumptions~\cite{8737464,8664630,dinh2019federated,wang2021network,wang2020tackling} below to conduct our analysis. Henceforth, $\Vert.\Vert$ denotes the 2-norm.

\begin{assumption}[Smoothness of the Loss Functions]\label{Assup:lossFun}
   The local loss function (defined in Sec.~\ref{sec:Formulation}) of each node $u\in\mathcal{U}$  is  $\beta$-smooth. That is, for any two model parameters $\bm{\omega}$ and $\widehat{\bm{\omega}}$, carried by the random walker, we have:
     \vspace{-2.mm}
    \begin{equation}
        \Vert \nabla \mathfrak{L}_u(\bm{\omega})- \nabla \mathfrak{L}_u(\widehat{\bm{\omega}}) \Vert \hspace{-.4mm}\leq \hspace{-.4mm} \beta \Vert \bm{\omega}- \widehat{\bm{\omega}} \Vert, 
    \end{equation}
    which implies the $\beta$-smoothness of the overall/global loss function $\mathfrak{L}$ (defined in Sec.~\ref{sec:Formulation}). 
    %   \begin{equation}
    %     \Vert \nabla f_u(\mathbf{w};x)- \nabla f_u(\mathbf{w}';x) \Vert \leq L \Vert \mathbf{w}-\mathbf{w}' \Vert,~\forall \mathbf{w},\mathbf{w}',x.
    % \end{equation}
\end{assumption}
 \vspace{-3mm}
\begin{assumption}[Bounded Dissimilarity of Local Loss Functions]\label{Assup:Dissimilarity}
For each node $u$, there exist a finite constant $\mathfrak{X}_u$, such that
%   For any set of coefficients $\{a_n\}_{n\in \mathcal{N}}$, where $\sum_{n\in \mathcal{N}} a_n=1$, there exist two finite constants $\zeta_1 \geq 1$, $\zeta_2 \geq 0$, such that
    \vspace{-1mm}
   \begin{equation}
    \Vert \nabla \mathfrak{L}_u(\bm{\omega})- \nabla \mathfrak{L}({\bm{\omega}}) \Vert \hspace{-.4mm}\leq \mathfrak{X}_u,~~\forall \bm{\omega}.
   \end{equation}
\end{assumption}
The parameter $\mathfrak{X}_u$ introduced in the assumption above quantifies the degree of data heterogeneity (i.e., the non-iid nature of data) across the nodes in the network. Specifically, larger values of $\mathfrak{X}_u$ indicate that the local dataset at node $u$ deviates more significantly from the global data distribution, thereby increasing the potential for model bias in the traversing walker. 
Building on this assumption, we now introduce an additional condition on the instantaneous model carried by the walker, which allows us to characterize the walker's instantaneous bias as a function of its current model parameters during traversal.

\begin{assumption}[Bounded Dissimilarity of Local Loss Functions under Instantaneous Walkers' Models]\label{Assup:Dissimilarity2}
Let $\{\bm{\omega}^{(k)}\}_{k=0}^{K}$ denote the sequence of the instantaneous models carried by the walker during its traversal. 
For each node $u\in\mathcal{U}$, there exist a finite constant $\mathfrak{X}^{(k)}_{u}$, such that
%   For any set of coefficients $\{a_n\}_{n\in \mathcal{N}}$, where $\sum_{n\in \mathcal{N}} a_n=1$, there exist two finite constants $\zeta_1 \geq 1$, $\zeta_2 \geq 0$, such that
    \vspace{-1mm}
   \begin{equation}
    \Vert \nabla \mathfrak{L}_u(\bm{\omega}^{(k)})- \nabla \mathfrak{L}({\bm{\omega}}^{(k)}) \Vert \hspace{-.4mm}\leq \mathfrak{X}^{(k)}_{u},~~\forall \bm{\omega}^{(k)}.
   \end{equation}
   It is trivial to verify that $\mathfrak{X}^{(k)}_{u}\leq \mathfrak{X}_u$, where $\mathfrak{X}_u$ is defined in Assumption~\ref{Assup:Dissimilarity}.
\end{assumption}

We next define \textit{local data variability}, which provides a finer-grained measure of the non-uniformity of data points inherent in each node’s local dataset.
 \vspace{-1mm}
\begin{definition}[Local Data Variability]\label{Assump:DataVariabilit}
 The local data variability at each node $u$ is quantified by a non-negative parameter $\Theta_u \geq 0$, which satisfies the following condition:
    % \nm{isnt this just a smoothness of f wrt to the data points?}
    \vspace{-2mm}
    \begin{equation}
    \hspace{-5mm}
   \Vert \nabla f\hspace{-.4mm}(\bm{\omega},d) \hspace{-.4mm}-\hspace{-.4mm} \nabla f\hspace{-.4mm}(\bm{\omega},d')\Vert  \hspace{-.4mm} \leq \hspace{-.4mm} \Theta_u \Vert d\hspace{-.4mm}-\hspace{-.4mm}d' \Vert,\hspace{-1.2mm}~~~\forall d,d'\hspace{-.7mm}\in\hspace{-.7mm}\mathcal{D}_u,~~\forall \bm{\omega}.
    \end{equation}
    % \nm{what norm are you using here? (and earlier)
    We further define $\Theta = \max_{u\in\mathcal{U}}\{\Theta_u \}$.
\end{definition}

We now present our first convergence result, which establishes the general convergence behavior of $\mathbb{X}$L. Below, we use $\mathbb{E}_s$ to denote expectation with respect to random data sampling incurred by conducting SGD, $\mathbb{E}_{\hat{\lambda}}$ to denote expectation with respect to node selection of the random walker, and  $\mathbb{E}_{s,\hat{\lambda}}$ to denote expectation with respect to both random data sampling and  node selection.

\begin{theorem}[Convergence Behavior of $\mathbb{X}$L]\label{th:main} Assume that the learning-rate satisfies {\small$ \eta_k   < \min\left\{\frac{1}{2 \beta}\sqrt{\frac{\zeta^{(k)}}{\left(1+ \zeta^{(k)}\right)  \left(\ell^{(k)}\right) \left(\ell^{(k)}-1\right)}}, \frac{1}{2\beta} \right\}$}, where {\small $\zeta^{(k)}\in (0,1/4)$} is a constant. Also, let $\left(\sigma_u\right)^2$
denote the variance of data points (i.e., their feature vectors) inside the dataset of user $u$, 
{\small $\Delta_{D_u}^{(k)}\triangleq \left(\frac{\widehat{\lambda}_{u}^{(k)}D-D_{u}}{D}\right)^2=\left(\widehat{\lambda}_{u}^{(k)} - \frac{D_{u}}{D}\right)^2$} denote a parameter that is a function of node sampling, and $G^2$ denote an upper-bound on the norm of generated gradients during model training, i.e., $\Vert\nabla \mathfrak{L}(\bm{\omega}^{(k)})\Vert^2 \leq G^2$,~~$\forall k$. Then, the cumulative average of the gradient of the walker's loss under $\mathbb{X}$L, which quantifies its convergence, satisfies the bound in~\eqref{eq:gen_conv}.
\end{theorem}
\begin{proof}
    Please refer to Appendix A for the details of the proof.
\end{proof}

\textbf{Interpretation of the Result:} Inspecting the bound in \eqref{eq:gen_conv}, several key insights can be drawn. In particular, the bound in~\eqref{eq:gen_conv} captures the effect of the ML-related parameters on the performance of $\mathbb{X}$L. For example, term $(a)$ captures the effect of consecutive loss function gains during ML training across consecutive jumps/steps. Also, terms $(b)$, $(e)$, and $(f)$ demonstrate the impact of  the size of SGD mini-batches $B_u$ on the model convergence: larger mini-batch sizes lead to a smaller bound and thus faster convergence.
Moreover, terms $(b)$, $(c)$, and $(f)$ capture the effect of the number of SGD iterations (via $\ell^{(k)}$) on the convergence. Terms $(c)$ and $(d)$ quantify the impact of data heterogeneity across nodes (as they get multiplied by $\mathfrak{X}^{(k)}_u$): term $(c)$ captures the interplay between the data heterogeneity of each node and the number of local SGD iterations, whereas $(d)$ reflects the interplay between the data heterogeneity of each node and the node-selection strategy. In particular, these terms suggest that greater data heterogeneity at a node $u$ (i.e., larger values of $\mathfrak{X}^{(k)}_u$) must be accompanied by fewer local SGD iterations (reflected via $\ell^{(k)}$ in term $(c)$) and a lower probability that the random walker visits node $u$ at step $k$ (reflected via $\Delta_{D_u}^{(k)}$ in term $(d)$), i.e., the bound suggests that the random walker may select those nodes with lower data heterogeneity which are the ones with closer data distribution to the global dataset. Furthermore, larger local data variability implies a larger bound ($\Theta$ in $(b)$, $(e)$, and $(f)$).
Additionally, term $(g)$ captures the interplay between gradient magnitude (represented by $G^2$) and node selection (represented by $\Delta_{D_u}^{(k)}$).

\begin{table*}[t!]
\vspace{-1mm}
\begin{minipage}{0.99\textwidth}
{\footnotesize
\begin{equation}
\hspace{-6mm}
\begin{aligned}\label{eq:gen_conv}
    &\frac{1}{K} \sum_{k=0}^{K-1}\mathbb{E}_s \left[\Bigg\Vert  \nabla \mathfrak{L}(\bm{\omega}^{(k)}) \Bigg\Vert^2\right]  \leq \frac{1}{K} \sum_{k=0}^{K-1} \underbrace{\frac{\mathbb{E}_s \left[\mathfrak{L}(\bm{\omega}^{(k)})\right]- \mathbb{E}_s \left[\mathfrak{L}(\bm{\omega}^{(k+1)})\right]}{\eta_{_k}\ell^{(k)}  \left(\frac{1}{4} - \zeta^{(k)}\right)}}_{(a)} +   \frac{1}{K} \sum_{k=0}^{K-1}\vast[  \frac{\left(\zeta^{(k)}+1\right)}{ \left(\frac{1}{4} - \zeta^{(k)}\right)}\underbrace{\sum_{u\in \mathcal{U}}\frac{D_{u}}{D }{  2 \beta^2\Theta^2 \eta_k^2} \left(\ell^{(k)}-1\right)  \left(1-\frac{{B}_{u}}{D_{u}} \right)  \frac{{(D_{u}-1)}
    \left(\sigma_{u}\right)^2}{D_{u}{B}_{u}}}_{(b)}  \\
    %%%%%%%%%
    & \hspace{24mm}+    \frac{\left(\zeta^{(k)}+1\right)}{\left(\frac{1}{4} - \zeta^{(k)}\right)}\sum_{u\in \mathcal{U}}  \left( \underbrace{\frac{D_{u}}{D} {4 \eta_k^2  \beta^2 }\left(\ell^{(k)}\right) \left(\ell^{(k)}-1\right)}_{(c)}+\underbrace{|\mathcal{U}|6  \Delta_{D_u}^{(k)}}_{(d)}\right)\mathfrak{X}^{(k)}_{u} + \frac{\beta\eta_{_k}}{\left(\frac{1}{4} - \zeta^{(k)}\right)}  \underbrace{ \sum_{u\in \mathcal{U}} \widehat{\lambda}_{u}^{(k)} \left(1-\frac{{B}_{u}}{D_{u}} \right) \frac{(D_{u}-1)\Theta^2\left(\sigma_{u}\right)^2}{D_{u}{B}_{u}}}_{(e)}\\
    &\hspace{24mm}+    \frac{|\mathcal{U}| \left(\zeta^{(k)}+1\right)}{ \left(\frac{1}{4} - \zeta^{(k)}\right)} \underbrace{\sum_{u\in \mathcal{U}}\Bigg( 12 \Theta^2 \beta^2 \Delta_{D_u}^{(k)} \eta_k^2  \left(\ell^{(k)}-1\right) \left(1-\frac{{B}_{u}}{D_{u}} \right)  \frac{{(D_{u}-1)}\left(\sigma_{u}\right)^2}{D_{u}{B}_{u}}\Bigg)}_{(f)} +   \underbrace{\frac{|\mathcal{U}|\left(\zeta^{(k)}+1\right)}{ \left(\frac{1}{4} - \zeta^{(k)}\right)} \sum_{u\in \mathcal{U}}\left(6  \Delta_{D_u}^{(k)}\right)G^2}_{(g)}\vast]
 \end{aligned}
 \hspace{-8mm}
 \end{equation}
  \vspace{-.1mm}
 \hrule
 }
 \end{minipage}
 % \vspace{-6.5mm}
\end{table*}

\begin{table*}[t!]
\vspace{-1mm}
\begin{minipage}{0.99\textwidth}
{\footnotesize
\begin{equation}
\hspace{-7mm}
\begin{aligned}\label{eq:cor1}
     \frac{1}{K} \sum_{k=0}^{K-1}\mathbb{E}_{s,\hat{\lambda}} \left[\Bigg\Vert  \nabla \mathfrak{L}(\bm{\omega}^{(k)}) \Bigg\Vert^2\right]  &\leq \frac{ \mathfrak{L}(\bm{\omega}^{(0)})-  \mathfrak{L}^{\star}}{ \sqrt{K}\sqrt{|\mathcal{U}|} \alpha\ell_{\mathsf{min}}   \left(\frac{1}{4} - \zeta_{\mathsf{max}}\right)} + \frac{{  2 \beta^2\Theta^2  \alpha^2 }\left(\ell_{\mathsf{max}}-1\right)\left(\zeta_{\mathsf{max}}+1\right)}{K\left(\frac{1}{4} - \zeta_{\mathsf{max}}\right) } \sigma_{\mathsf{max}}\\
     &+ \frac{ \left({4  \alpha^2 \beta^2 }\left(\ell_{\mathsf{max}}\right) \left(\ell_{\mathsf{max}}-1\right) \mathfrak{X}_{\mathsf{max}} +2|\mathcal{U}| \alpha^2 \mathfrak{X}_{\mathsf{max}}\right)\left(\zeta_{\mathsf{max}}+1\right)}{K\left(\frac{1}{4} - \zeta_{\mathsf{max}}\right) } +  \frac{4\beta \ell_{\mathsf{max}} |\mathcal{U}| \alpha \mathfrak{X}_{\mathsf{max}}\left(\zeta_{\mathsf{max}}+1\right)}{\sqrt{K} \left(\frac{1}{4} - \zeta_{\mathsf{max}}\right) }\\
     &+ \frac{ (16 \ell_{\mathsf{max}} \Theta^2 \beta^3 ( \alpha^3 |\mathcal{U}|  \left(\ell_{\mathsf{max}}-1\right)\sigma_{\mathsf{max}}\left(\zeta_{\mathsf{max}}+1\right)}{K^{3/2}\left(\frac{1}{4} -  \zeta_{\mathsf{max}}\right)} +\frac{ 8   \Theta^2 \beta^2 (\alpha^4 |\mathcal{U}| \left(\ell_{\mathsf{max}}-1\right)) \sigma_{\mathsf{max}}\left(\zeta_{\mathsf{max}}+1\right)}{K^2\left(\frac{1}{4} -  \zeta_{\mathsf{max}}\right)} \\
     &+ \frac{ 4\beta \ell_{\mathsf{max}}|\mathcal{U}| \alpha  G^2 \left(\zeta_{\mathsf{max}}+1\right)}{\sqrt{K}\left(\frac{1}{4} -  \zeta_{\mathsf{max}}\right)} +\frac{2|\mathcal{U}|  \alpha^2   G^2 \left(\zeta_{\mathsf{max}}+1\right)}{K\left(\frac{1}{4} -  \zeta_{\mathsf{max}}\right)}+ \frac{\beta\alpha  2\Theta^2}{\sqrt{K} 2  \left(\frac{1}{4} - \zeta_{\mathsf{max}}\right)}   \sigma_{\mathsf{max}}\\
     & \overset{(a)}{=}  O\left(\frac{1}{\sqrt{K}}\right)+ O\left(\frac{1}{{K}}\right)+O\left(\frac{1}{{K}}\right)+O\left(\frac{1}{\sqrt{K}}\right)+O\left(\frac{1}{{K}^{3/2}}\right)+O\left(\frac{1}{{K}^2}\right)+O\left(\frac{1}{\sqrt{K}}\right)+O\left(\frac{1}{{K}}\right)+O\left(\frac{1}{\sqrt{K}}\right) = O\left(\frac{1}{\sqrt{K}}\right)
\end{aligned}
\end{equation}
  \vspace{-.1mm}
 }
 \hrule
 \end{minipage}
 % \vspace{-6.5mm}
\end{table*}

Building upon the result of Theorem~\ref{th:main}, we next derive a set of \textit{sufficient} conditions under which $\mathbb{X}$L exhibits guaranteed convergence with a rate of ${O}(1/\sqrt{K})$ (i.e., matching the convergence rate of SGD over non-convex smooth loss functions):
\begin{corollary}[Guaranteed Convergence of $\mathbb{X}$L]\label{cor:main}
 In addition to the conditions in Theorem~\ref{th:main}, further assume that (i) {\small $\eta_k = \alpha \big /{\sqrt{K}}$} with a finite positive constant $\alpha$ chosen to satisfy the condition on {\small $\eta_k$} in Theorem~\ref{th:main}, (ii) {\small $\max_{u}\left\{\left(1-\frac{{B}_{u}}{D_{u}} \right)  \frac{(D_{u}-1) \left(\sigma_{u}\right)^2}{D_{u}{B}_{u}} \right\} \leq \sigma_{\mathsf{max}}$}, (iii)  {\small $\max_k \left\{\zeta^{(k)}\right\} \leq \zeta_{\mathsf{max}}< 1$}, (iv)  {\small$q_{u}^{(k)}= \left(\frac{\beta \ell^{(k)}\left(\mathbb{E}_{\hat{\lambda}}\left[\widehat{\lambda}_{u}^{(k)}\right]D^2-2\mathbb{E}_{\hat{\lambda}}\left[\widehat{\lambda}_{u}^{(k)}\right]D D_{u}+ D^2_{u}\right)+\eta_k D^2}{D^2}\right)$}, (v) {\small $\ell_{\mathsf{min}} \le \ell^{(k)}\leq \ell_{\mathsf{max}}$}, and (vi) {\small$\max_{k,u}\left\{\mathfrak{X}_{u}^{(k)}\right\}\le \mathfrak{X}_{\mathsf{max}}$}. Also, assume the the node sampling by the walker satisfies $\mathbb{E}_{\hat{\lambda}}\left[\widehat{\lambda}_{u}^{(k)}\right]- \frac{D_{u}}{D} \le \eta_k$
and presume that the random walker's movements form an ergodic Markov chain where the walker's instantiation location (i.e., the node from which the walker starts its journey) follows its stationary distribution, meaning its initial location is drawn according to its stationary distribution of visit of the nodes.
% and $\max_{k}\left\{\left(\sum_{n\in \mathcal{N}}\frac{{D}^{(k)}_{n}e^{(k)}_{n}}{D^{(k)} }\right)^{-1}\right\}\leq \overline{e}_{\mathsf{max}}$. 
Then, the cumulative average of the gradient of the walker's loss under $\mathbb{X}$L satisfies~\eqref{eq:cor1}, implying
$\frac{1}{K} \sum_{k=0}^{K-1}\mathbb{E}_{s,\hat{\lambda}} \left[\left\Vert  \nabla \mathfrak{L}(\bm{\omega}^{(k)}) \right\Vert^2\right]{=}O\left(\frac{1}{\sqrt{K}}\right)$.
\end{corollary}
\begin{proof}
    Please refer to Appendix B for the details of the proof.
\end{proof}

\textbf{Interpretation of the Result:} Bound~\eqref{eq:cor1} no longer depends on consecutive loss gains that was captured by term $(a)$ in~\eqref{eq:gen_conv}; instead, it depends only on the initial model error $\mathfrak{L}(\bm{\omega}^{(0)})-  \mathfrak{L}^{\star}$, where $\mathfrak{L}^{\star}$ denotes the minimum value of the loss function. Further, each term in the bound is in the order of 
either $O\left(\frac{1}{\sqrt{K}}\right)$, $O\left(\frac{1}{{K}}\right)$, $O\left(\frac{1}{{K}^2}\right)$, or $O\left(\frac{1}{{K}^{3/2}}\right)$ as written in the penultimate equality in~\eqref{eq:cor1} with super-index (a), which are collectively in the order of $O\left(\frac{1}{\sqrt{K}}\right)$.  In other words, the bound implies that {\small $\frac{1}{K} \sum_{k=0}^{K-1}\mathbb{E}_{s,\hat{\lambda}} \left[\left\Vert  \nabla \mathfrak{L}(\bm{\omega}^{(k)}) \right\Vert^2\right]{=}O\left(\frac{1}{\sqrt{K}}\right)$}, which in turn results in {\small $\lim_{K\rightarrow \infty}\frac{1}{K} \sum_{k=0}^{K-1}\mathbb{E}_{s,\hat{\lambda}} \left[\left\Vert  \nabla \mathfrak{L}(\bm{\omega}^{(k)}) \right\Vert^2\right] {=} 0$} demonstrating the convergence of the walker's model to a stationary point of the loss function (as in conventional centralized SGD). 
Inspecting the additional conditions imposed in the above corollary to guarantee convergence, the requirement on \textit{node selection} is particularly noteworthy:
the condition $
\mathbb{E}_{\hat{\lambda}}\left[\widehat{\lambda}_{u}^{(k)}\right] - \frac{D_{u}}{D} \le \eta_k
$ implies that, to ensure convergence, the \textit{stationary distribution} of node visits (i.e., the long-run fraction of times each node $u$ is visited, which corresponds to $\mathbb{E}_{\hat{\lambda}}\left[\widehat{\lambda}_{u}^{(k)}\right]$ since the Markov chains of node visits were presumed to be ergodic) must be close to the \textit{data-size-based sampling ratio} $\frac{D_u}{D}$, where $D_u$ is the size of the local dataset and $D$ is the total/global dataset size. In particular, as $k \to \infty$, the step-size condition in the corollary ensures that $\eta_k \to 0$, leading to
$
\mathbb{E}_{\hat{\lambda}}\left[\widehat{\lambda}_{u}^{(k)}\right] \to \frac{D_{u}}{D},
$
which means that the \textit{asymptotic visitation frequency of each node must match its data contribution proportion}.

\section{Conclusion}\label{conclusion}
\noindent We introduced $\mathbb{X}$L, a novel framework that expands distributed ML by treating ML models as autonomous agents capable of traversing the network through flexible, multi-hop strategies. The presented formulations and empirical results demonstrated the significant potential of $\mathbb{X}$L. Also, a set of convergence analysis was performed to inform the model convergence behavior of $\mathbb{X}$L. Further, a series of future research directions were discussed, broadening the horizon of $\mathbb{X}$L.

\bibliography{OFedRefs}

\section*{Figure Legends}

\textbf{Figure 1.} Centralized to semi/fully-decentralized FedL/FogL architectures (A)-(C), and the extension presented by the $\mathbb{X}$L (D). In centralized to semi/fully-decentralized FedL/FogL, each node has its own local ML model, whereas in $\mathbb{X}$L there are only two ML models associated with the random walkers, inducing two active sessions per training round.
\\
\\
\textbf{Figure 2.} Performance of random walkers under different node traversal strategies for CIFAR-10 and SVHN datasets (smoothed using a moving average with a window size of 10). Referring to \textbf{Sec.~\ref{sec:temp}}, among methods with \textit{fixed/static} node importance values (red, blue, green, magenta, and yellow curves using static sampling or fixed $Z \in \{0, 0.5, 1\}$), our approach with non-binary weighting of data and spatial quality metrics (yellow curve, $Z = 0.5$) yields the best performance. Furthermore, referring to~\textbf{Sec.~\ref{sec:perception}}, our method with \textit{dynamic} node importance (black curve using a time-varying  $Z^{(k)}_{\mathsf{Inst}}$ based on current model accuracy) outperforms all the static strategies.
\\
\\
\textbf{Figure 3.} Performance comparison between our elastic random walker, where the number of SGD iterations is adaptively scaled based on node data quality, and methods with fixed SGD iteration numbers for CIFAR-10 and SVHN datasets (smoothed using a moving average with a window size of 10). Our elastic walker (blue curves) outperforms all baselines with static SGD iteration numbers (as depicted in the top plot for each dataset) while performing fewer SGD updates (as depicted in the bottom plot for each dataset).
\\
\\
\textbf{Figure 4.} Performance comparisons (smoothed using a moving average with a window size of 10) between a memory-enabled random walker and a memoryless walker, both using the same node traversal strategy for CIFAR-10 and SVHN datasets. Initially, memory is disabled (red region) to accelerate early training. As training progresses, memory influence gradually increases (green and purple regions), helping mitigate model bias and yielding notable gains -- especially in the mid to late stages of training.
\\
\\
\textbf{Figure 5.} Impact of the number of random walkers on model convergence for CIFAR-10 and SVHN datasets. The top plot of each  dataset shows the instantaneous average performance across all walkers, where increasing the number of walkers initially boosts performance; however, beyond a certain point (e.g., from 10 to 14 walkers), the improvements become marginal. The bottom plot of each  dataset presents the walkers' final accuracies, highlighting an initial sharp gain followed by diminishing gains as more walkers are deployed.
\\
\\
\textbf{Figure 6.} Impact of inter-walker attraction on model performance (depicted using a moving average with a window size of 10) for CIFAR-10 and SVHN datasets (shaded regions depict $\pm$25\% of standard deviation of performance across the walkers). As the attraction between walkers increases (i.e., higher values of $A$, leading to shorter inter-collision intervals), the walkers interact more frequently, which results in mitigating model bias and an improved performance. For comparison, the orange curve illustrates the (hypothetical) upper bound achieved by centrally aggregating the walkers' models after every jump.
\\
\\

\section*{Author contributions}
A. Salihovic, P. Abdisarabshali, M. Langberg, and
S. Hosseinalipour conceptualized the paper, prepared the first draft of the paper, and refined the paper.

\noindent  The final version of the paper was critically
reviewed and approved by all authors.

\section*{Acknowledgements}
The authors acknowledge support from the National Science Foundation (NSF) under Grant ECCS 2512911.

\section*{Competing interests}
The authors declare no competing interests.
\section*{Additional information}
Correspondence and requests for materials should be addressed to
S. Hosseinalipour.

\vfill
\begingroup

\appendix
\newpage
\section{Proof of Theorem \ref{th:main}}\label{app:th:main}

\noindent Consider the following lemma which is a direct result of $\beta$-smoothness of a function.
\begin{lemma}[Smooth Function Characteristics~\cite{2200000050}]\label{lemma:smooth}
    Let $f$ be a $\beta$-smooth function on $\mathbb{R}^n$. Then for any $\bm{x},\bm{y}\in \mathbb{R}^n$, we have
    \begin{equation}
        f(\bm{x})\le f(\bm{y})+\left\langle\nabla f(\bm{y}), (\bm{x}-\bm{y})\right\rangle +\frac{\beta}{2}\left\Vert \bm{x}-\bm{y} \right\Vert^2.
    \end{equation}
\end{lemma}

\noindent Using the $\beta$-smoothness of the global loss function (Assumption~\ref{Assup:lossFun}) and considering Lemma~\ref{lemma:smooth}, we have 
\begin{equation}
 \mathfrak{L}(\bm{\omega}^{(k+1)}) \leq \mathfrak{L}(\bm{\omega}^{(k)}) +  \left\langle\nabla{\mathfrak{L}(\bm{\omega}^{(k)})},\left( \bm{\omega}^{(k+1)} - \bm{\omega}^{(k)}\right)\right\rangle+ \frac{\beta}{2} \left\Vert \bm{\omega}^{(k+1)} - \bm{\omega}^{(k)}\right\Vert^2.
\end{equation}
Replacing the updating rule for $\bm{\omega}^{(k+1)}$ and taking the conditional expectation (with respect to randomized data sampling at the last aggregation, referred to by $\mathbb{E}_s$) from both hand sides yields
\begin{align}\label{ineq:main1}
 \mathbb{E}_s\left[\mathfrak{L}(\bm{\omega}^{(k+1)})\right] &\leq   \mathfrak{L}(\bm{\omega}^{(k)}) - \eta_{_k}\ell^{(k)}\mathbb{E}_s\left[\left\langle\nabla{\mathfrak{L}(\bm{\omega}^{(k)})}, \sum_{u\in \mathcal{U}}\frac{\widehat{\lambda}_{u}^{(k)}}{ \ell^{(k)}}\widetilde{\nabla \mathfrak{L}}_{u}^{(k)}\right\rangle\right] + \frac{\beta\eta_{_k}^2\left(\ell^{(k)}\right)^2}{2} \mathbb{E}_s\left[\left\Vert \sum_{u\in \mathcal{U}}\frac{\widehat{\lambda}_{u}^{(k)}}{ \ell^{(k)}}\widetilde{\nabla \mathfrak{L}}_{u}^{(k)}\right\Vert^2\right],
\end{align}
where $\widetilde{\nabla \mathfrak{L}}_{u}^{(k)} = \frac{1}{\eta_{_k}}\left(\bm{\omega}^{(k)}-\bm{\omega}_{u}^{(k),\ell^{(k)}}\right)$. Via recursive expansion of the update rule in~\eqref{eq:updaterule}, we get
\begin{equation}\label{eq:nablabarF}
\widetilde{\nabla \mathfrak{L}}_{u}^{(k)} = \sum_{\ell=1}^{\ell^{(k)}} \sum_{d\in \mathcal{B}^{(k),\ell}_{u}} \hspace{-3mm} {\frac{\nabla  f(\bm{\omega}^{(k),\ell-1}_{u},d)}{{B}_{u}}}.
\end{equation}
Replacing the above result back in \eqref{ineq:main1} leads to
\begin{equation}
\begin{aligned}
    \mathbb{E}_s\left[\mathfrak{L}(\bm{\omega}^{(k+1)})\right] &\leq  \mathfrak{L}(\bm{\omega}^{(k)}) - \eta_{_k} \ell^{(k)}\mathbb{E}_s\left[\left\langle\nabla{\mathfrak{L}(\bm{\omega}^{(k)})}, \sum_{u\in \mathcal{U}}\frac{\widehat{\lambda}_{u}^{(k)}}{ \ell^{(k)}}\sum_{\ell=1}^{\ell^{(k)}} \sum_{d\in \mathcal{B}^{(k),\ell}_{u}} \hspace{-3mm} {\frac{\nabla  f(\bm{\omega}^{(k),\ell-1}_{u},d)}{{B}_{u}}}\right\rangle\right] + \frac{\beta\eta_{_k}^2\left(\ell^{(k)}\right)^2}{2} \mathbb{E}_s\left[\left\Vert  \sum_{u\in \mathcal{U}}\frac{\widehat{\lambda}_{u}^{(k)}}{ \ell^{(k)}}\widetilde{\nabla \mathfrak{L}}_{u}^{(k)}\right\Vert^2\right].
\end{aligned}
\end{equation}
Let $\mathscr{N}_u^{\mathsf{G},(k),\ell}=\sum_{d\in \mathcal{B}^{(k),\ell}_{u}}{\frac{\nabla  f(\bm{\omega}^{(k),\ell-1}_{u},d)}{{B}_{u}}}-{\nabla  \mathfrak{L}_u(\bm{\omega}^{(k),\ell-1}_{u})}$ denote the noise of SGD of node $u$, where ${\nabla  \mathfrak{L}_u(\bm{\omega}^{(k),\ell-1}_{u})}=\sum_{d\in\mathcal{D}_{u}}\frac{\nabla  f(\bm{\omega}^{(k),\ell-1}_{u},d)}{D_{u}}$. The above inequality can be written as follows:
\begin{align}\label{ineq:main2}
    \mathbb{E}_s&\left[\mathfrak{L}(\bm{\omega}^{(k+1)})\right] \leq  \mathfrak{L}(\bm{\omega}^{(k)})- \eta_{_k}\ell^{(k)} \mathbb{E}_s\left[\left\langle\nabla{\mathfrak{L}(\bm{\omega}^{(k)})}, \sum_{u\in \mathcal{U}}\frac{\widehat{\lambda}_{u}^{(k)}}{ \ell^{(k)}}\sum_{\ell=1}^{\ell^{(k)}}  \left({\nabla  \mathfrak{L}_u(\bm{\omega}^{(k),\ell-1}_{u})}+\mathscr{N}_u^{\mathsf{G},(k),\ell}\right)\right\rangle\right]\nonumber\\
    &~~~~~~~~+ \frac{\beta\eta_{_k}^2\left(\ell^{(k)}\right)^2}{2} \mathbb{E}_s\left[\left\Vert  \sum_{u\in \mathcal{U}}\frac{\widehat{\lambda}_{u}^{(k)}}{ \ell^{(k)}}\widetilde{\nabla \mathfrak{L}}_{u}^{(k)}\right\Vert^2\right]\nonumber\\
    &=\mathfrak{L}(\bm{\omega}^{(k)}) - \eta_{_k}\ell^{(k)} \mathbb{E}_s\left[\left\langle\nabla{\mathfrak{L}(\bm{\omega}^{(k)})}, \sum_{u\in \mathcal{U}}\frac{\widehat{\lambda}_{u}^{(k)}}{ \ell^{(k)}}\sum_{\ell=1}^{\ell^{(k)}}  {\nabla  \mathfrak{L}_u(\bm{\omega}^{(k),\ell-1}_{u})}\right\rangle\right]\nonumber\\
    &~~~~~~~-\underbrace{\eta_{_k}\ell^{(k)} \mathbb{E}_s\left[\left\langle\nabla{\mathfrak{L}(\bm{\omega}^{(k)})}, \sum_{u\in \mathcal{U}}\frac{\widehat{\lambda}_{u}^{(k)}}{ \ell^{(k)}}\sum_{\ell=1}^{\ell^{(k)}}\mathscr{N}_u^{\mathsf{G},(k),\ell} \right\rangle\right]}_{(x_1)}+ \frac{\beta\eta_{_k}^2\left(\ell^{(k)}\right)^2}{2} \mathbb{E}_s\left[\left\Vert  \sum_{u\in \mathcal{U}}\frac{\widehat{\lambda}_{u}^{(k)}}{ \ell^{(k)}}\widetilde{\nabla \mathfrak{L}}_{u}^{(k)}\right\Vert^2\right]\nonumber\\
    &\overset{(i)}{=}\mathfrak{L}(\bm{\omega}^{(k)}) - \eta_{_k}\ell^{(k)} \mathbb{E}_s\left[\left\langle\nabla{\mathfrak{L}(\bm{\omega}^{(k)})}, \sum_{u\in \mathcal{U}}\frac{\widehat{\lambda}_{u}^{(k)}}{ \ell^{(k)}}\sum_{\ell=1}^{\ell^{(k)}}  {\nabla  \mathfrak{L}_u(\bm{\omega}^{(k),\ell-1}_{u})}\right\rangle\right]\nonumber\\
    &~~~~~~~+ \frac{\beta \eta_{_k}^2\left(\ell^{(k)}\right)^2}{2} \mathbb{E}_s\left[\left\Vert  \sum_{u\in \mathcal{U}}\frac{\widehat{\lambda}_{u}^{(k)}}{ \ell^{(k)}}\widetilde{\nabla \mathfrak{L}}_{u}^{(k)}\right\Vert^2\right],
\end{align}
where $(i)$ uses the fact that $(x_1)$ is zero since the noise of gradient estimation across the mini-batches are independent and zero mean. Further, let 
\begin{equation}\label{eq:node_selection_noise}
    \mathscr{U}^{(k)}= \sum_{u\in \mathcal{U}}\frac{\widehat{\lambda}_{u}^{(k)}}{ \ell^{(k)}}\sum_{\ell=1}^{\ell^{(k)}} {\nabla  \mathfrak{L}_u(\bm{\omega}^{(k),\ell-1}_{u})}- \sum_{u\in \mathcal{U}}\frac{D_{u}}{D \ell^{(k)}}\sum_{\ell=1}^{\ell^{(k)}} {\nabla  \mathfrak{L}_u(\bm{\omega}^{(k),\ell-1}_{u})} 
\end{equation}

denote the noise of node selection of the random walker. Inequality \eqref{ineq:main2} can be rewritten as follows:
\begin{align}
    \mathbb{E}_s&\left[\mathfrak{L}(\bm{\omega}^{(k+1)})\right] \leq \mathfrak{L}(\bm{\omega}^{(k)}) - \eta_{_k} \ell^{(k)}\mathbb{E}_s\left[\left\langle\nabla{\mathfrak{L}(\bm{\omega}^{(k)})}, \sum_{u\in \mathcal{U}}\frac{D_{u}}{D \ell^{(k)}}\sum_{\ell=1}^{\ell^{(k)}}  {\nabla  \mathfrak{L}_u(\bm{\omega}^{(k),\ell-1}_{u})}+ \mathscr{U}^{(k)} \right\rangle\right]\nonumber\\
    &~~~~~~~~~~~+ \frac{\beta\eta_{_k}^2\left(\ell^{(k)}\right)^2}{2} \mathbb{E}_s\left[\left\Vert  \sum_{u\in \mathcal{U}}\frac{\widehat{\lambda}_{u}^{(k)}}{ \ell^{(k)}}\widetilde{\nabla \mathfrak{L}}_{u}^{(k)}\right\Vert^2\right]\nonumber\\
    &=\mathfrak{L}(\bm{\omega}^{(k)}) - \eta_{_k} \ell^{(k)}\mathbb{E}_s\left[\left\langle\nabla{\mathfrak{L}(\bm{\omega}^{(k)})}, \sum_{u\in \mathcal{U}}\frac{D_{u}}{D \ell^{(k)}}\sum_{\ell=1}^{\ell^{(k)}}  {\nabla  \mathfrak{L}_u(\bm{\omega}^{(k),\ell-1}_{u})}\right\rangle\right]\nonumber\\
    &+ \eta_{_k}\ell^{(k)} \mathbb{E}_s\left[\left\langle\nabla{\mathfrak{L}(\bm{\omega}^{(k)})},-\mathscr{U}^{(k)} \right\rangle\right]+ \frac{\beta\eta_{_k}^2\left(\ell^{(k)}\right)^2}{2} \mathbb{E}_s\Bigg[\Bigg\Vert  \sum_{u\in \mathcal{U}}\frac{\widehat{\lambda}_{u}^{(k)}}{ \ell^{(k)}}\widetilde{\nabla \mathfrak{L}}_{u}^{(k)} \Bigg\Vert^2\Bigg].
\end{align}
Similarly, writing the last term of the above inequality in terms of $\mathscr{N}_u^{\mathsf{G},(k),\ell}$ gives us
\begin{align}
    %%%%%%%%%
    \mathbb{E}_s&\left[\mathfrak{L}(\bm{\omega}^{(k+1)})\right] \leq \mathfrak{L}(\bm{\omega}^{(k)}) - \eta_{_k} \ell^{(k)}\mathbb{E}_s\left[\left\langle\nabla{\mathfrak{L}(\bm{\omega}^{(k)})}, \sum_{u\in \mathcal{U}}\frac{D_{u}}{D \ell^{(k)}}\sum_{\ell=1}^{\ell^{(k)}}  {\nabla  \mathfrak{L}_u(\bm{\omega}^{(k),\ell-1}_{u})}\right\rangle\right]\nonumber\\
    &+ \eta_{_k}\ell^{(k)} \mathbb{E}_s\left[\left\langle\nabla{\mathfrak{L}(\bm{\omega}^{(k)})},-\mathscr{U}^{(k)} \right\rangle\right]+ \frac{\beta\eta_{_k}^2\left(\ell^{(k)}\right)^2}{2} \mathbb{E}_s\Bigg[\Bigg\Vert  \sum_{u\in \mathcal{U}}\frac{\widehat{\lambda}_{u}^{(k)}}{ \ell^{(k)}}\sum_{\ell=1}^{\ell^{(k)}} \sum_{d\in \mathcal{B}^{(k),\ell}_{u}} \hspace{-3mm} {\frac{\nabla  f(\bm{\omega}^{(k),\ell-1}_{u},d)}{{B}_{u}}} \Bigg\Vert^2\Bigg]\nonumber\\
    &=\mathfrak{L}(\bm{\omega}^{(k)}) - \eta_{_k} \ell^{(k)}\mathbb{E}_s\left[\left\langle\nabla{\mathfrak{L}(\bm{\omega}^{(k)})}, \sum_{u\in \mathcal{U}}\frac{D_{u}}{D \ell^{(k)}}\sum_{\ell=1}^{\ell^{(k)}}  {\nabla  \mathfrak{L}_u(\bm{\omega}^{(k),\ell-1}_{u})}\right\rangle\right]\nonumber\\
    &+ \eta_{_k}\ell^{(k)} \mathbb{E}_s\left[\left\langle\nabla{\mathfrak{L}(\bm{\omega}^{(k)})},-\mathscr{U}^{(k)} \right\rangle\right]+ \frac{\beta\eta_{_k}^2\left(\ell^{(k)}\right)^2}{2} \mathbb{E}_s\Bigg[\Bigg\Vert  \sum_{u\in \mathcal{U}}\frac{\widehat{\lambda}_{u}^{(k)}}{ \ell^{(k)}}\sum_{\ell=1}^{\ell^{(k)}} \left({\nabla  \mathfrak{L}_u(\bm{\omega}^{(k),\ell-1}_{u})}+\mathscr{N}_u^{\mathsf{G},(k),\ell}\right) \Bigg\Vert^2\Bigg]\nonumber\\
    %%%%%%%
    &=\mathfrak{L}(\bm{\omega}^{(k)}) - \eta_{_k}\ell^{(k)} \mathbb{E}_s\left[\left\langle\nabla{\mathfrak{L}(\bm{\omega}^{(k)})}, \sum_{u\in \mathcal{U}}\frac{D_{u}}{D \ell^{(k)}}\sum_{\ell=1}^{\ell^{(k)}}  {\nabla  \mathfrak{L}_u(\bm{\omega}^{(k),\ell-1}_{u})}\right\rangle\right]\nonumber\\
    &~~+ \eta_{_k}\ell^{(k)} \mathbb{E}_s\left[\left\langle\nabla{\mathfrak{L}(\bm{\omega}^{(k)})},-\mathscr{U}^{(k)} \right\rangle\right]\nonumber\\
    &~~+ \frac{\beta\eta_{_k}^2\left(\ell^{(k)}\right)^2}{2} \mathbb{E}_s\left[\left\Vert  \sum_{u\in \mathcal{U}}\frac{\widehat{\lambda}_{u}^{(k)}}{ \ell^{(k)}}\sum_{\ell=1}^{\ell^{(k)}} {\nabla  \mathfrak{L}_u(\bm{\omega}^{(k),\ell-1}_{u})}+\left( \sum_{u\in \mathcal{U}}\frac{\widehat{\lambda}_{u}^{(k)}}{ \ell^{(k)}}\sum_{\ell=1}^{\ell^{(k)}}\mathscr{N}_u^{\mathsf{G},(k),\ell}\right) \right\Vert^2\right]\nonumber\\
    %%%%%%%
    &\overset{(i)}{=}\mathfrak{L}(\bm{\omega}^{(k)}) - \eta_{_k}\ell^{(k)} \mathbb{E}_s\left[\left\langle\nabla{\mathfrak{L}(\bm{\omega}^{(k)})}, \sum_{u\in \mathcal{U}}\frac{D_{u}}{D \ell^{(k)}}\sum_{\ell=1}^{\ell^{(k)}}  {\nabla  \mathfrak{L}_u(\bm{\omega}^{(k),\ell-1}_{u})}\right\rangle\right]\nonumber\\
    &~~+ \eta_{_k}\ell^{(k)} \mathbb{E}_s\left[\left\langle\nabla{\mathfrak{L}(\bm{\omega}^{(k)})},-\mathscr{U}^{(k)} \right\rangle\right]\nonumber\\
    &~~+ \frac{\beta\eta_{_k}^2\left(\ell^{(k)}\right)^2}{2} \mathbb{E}_s\Bigg[\Bigg\Vert  \sum_{u\in \mathcal{U}}\frac{\widehat{\lambda}_{u}^{(k)}}{ \ell^{(k)}}\sum_{\ell=1}^{\ell^{(k)}} {\nabla  \mathfrak{L}_u(\bm{\omega}^{(k),\ell-1}_{u})}\Bigg\Vert^2\Bigg]+\frac{\beta\eta_{_k}^2\left(\ell^{(k)}\right)^2}{2}\mathbb{E}_s\Bigg[\Bigg\Vert \sum_{u\in \mathcal{U}}\frac{\widehat{\lambda}_{u}^{(k)}}{ \ell^{(k)}}\sum_{\ell=1}^{\ell^{(k)}}\mathscr{N}_u^{\mathsf{G},(k),\ell}\Bigg\Vert^2\Bigg]\nonumber\\
    &~~+\underbrace{\beta\eta_{_k}^2\left(\ell^{(k)}\right)^2 \mathbb{E}_s\Bigg[\left\langle  \sum_{u\in \mathcal{U}}\frac{\widehat{\lambda}_{u}^{(k)}}{ \ell^{(k)}}\sum_{\ell=1}^{\ell^{(k)}} {\nabla  \mathfrak{L}_u(\bm{\omega}^{(k),\ell-1}_{u})} , \sum_{u\in \mathcal{U}}\frac{\widehat{\lambda}_{u}^{(k)}}{ \ell^{(k)}}\sum_{\ell=1}^{\ell^{(k)}}\mathscr{N}_u^{\mathsf{G},(k),\ell}\right\rangle\Bigg]}_{(x_2)}\nonumber\\
    %%%%%%%
    &\overset{(ii)}{=}\mathfrak{L}(\bm{\omega}^{(k)}) - \eta_{_k}\ell^{(k)} \mathbb{E}_s\left[\left\langle\nabla{\mathfrak{L}(\bm{\omega}^{(k)})}, \sum_{u\in \mathcal{U}}\frac{D_{u}}{D \ell^{(k)}}\sum_{\ell=1}^{\ell^{(k)}}  {\nabla  \mathfrak{L}_u(\bm{\omega}^{(k),\ell-1}_{u})}\right\rangle\right]\nonumber\\
    &~+ \eta_{_k}\ell^{(k)} \mathbb{E}_s\left[\left\langle\nabla{\mathfrak{L}(\bm{\omega}^{(k)})},-\mathscr{U}^{(k)} \right\rangle\right]\nonumber\\
    &~+ \frac{\beta\eta_{_k}^2\left(\ell^{(k)}\right)^2}{2} \mathbb{E}_s\Bigg[\Bigg\Vert  \sum_{u\in \mathcal{U}}\frac{\widehat{\lambda}_{u}^{(k)}}{ \ell^{(k)}}\sum_{\ell=1}^{\ell^{(k)}} {\nabla  \mathfrak{L}_u(\bm{\omega}^{(k),\ell-1}_{u})}\Bigg\Vert^2\Bigg]+\frac{\beta\eta_{_k}^2\left(\ell^{(k)}\right)^2}{2}\mathbb{E}_s\Bigg[\Bigg\Vert \sum_{u\in \mathcal{U}}\frac{\widehat{\lambda}_{u}^{(k)}}{ \ell^{(k)}}\sum_{\ell=1}^{\ell^{(k)}}\mathscr{N}_u^{\mathsf{G},(k),\ell}\Bigg\Vert^2\Bigg],
\end{align}
where $(i)$ uses the fact that for any two real valued vectors $\bm{a}$ and $\bm{b}$ with the same length, we have $\Vert \bm{a}+\bm{b}\Vert^2=\Vert \bm{a}\Vert^2+\Vert \bm{b}\Vert^2+2\langle\bm{a},\bm{b}\rangle$. Further, $(ii)$ in the above inequality is due to the fact that $(x_2)$ is zero since the noise of gradient estimation of nodes are independent and zero mean. Using $\Vert \sum_{i=1}^{n}\bm{x}_i\Vert^2=\sum_{i=1}^{n}\Vert\bm{x}_i\Vert^2+\sum_{\substack{j=1\\j\neq i}}^{n}\left\langle\bm{x}_i,\bm{x}_j\right\rangle$, where $\bm{x}\in \mathbb{R}^m$, and performing some algebraic manipulation, we expand the last term of the above inequality as follows: 
\begin{align}\label{ineq:main3}
    %%%%%%%%%
    \mathbb{E}_s&\left[\mathfrak{L}(\bm{\omega}^{(k+1)})\right] \leq \mathfrak{L}(\bm{\omega}^{(k)}) - \eta_{_k}\ell^{(k)} \mathbb{E}_s\left[\left\langle\nabla{\mathfrak{L}(\bm{\omega}^{(k)})}, \sum_{u\in \mathcal{U}}\frac{D_{u}}{D \ell^{(k)}}\sum_{\ell=1}^{\ell^{(k)}}  {\nabla  \mathfrak{L}_u(\bm{\omega}^{(k),\ell-1}_{u})}\right\rangle\right]\nonumber\\
    &~~~~+ \eta_{_k}\ell^{(k)} \mathbb{E}_s\left[\left\langle\nabla{\mathfrak{L}(\bm{\omega}^{(k)})},-\mathscr{U}^{(k)} \right\rangle\right]+ \frac{\beta\eta_{_k}^2\left(\ell^{(k)}\right)^2}{2}  \sum_{u\in \mathcal{U}}\mathbb{E}_s\Bigg[\left(\frac{\widehat{\lambda}_{u}^{(k)}}{ \ell^{(k)}}\right)^2\Bigg\Vert \sum_{\ell=1}^{\ell^{(k)}}\mathscr{N}_u^{\mathsf{G},(k),\ell}\Bigg\Vert^2\Bigg]\nonumber\\
    &+\underbrace{ \sum_{u\in \mathcal{U}} \sum_{u'\in \mathcal{U}\setminus u}\frac{\beta\eta_{_k}^2\left(\ell^{(k)}\right)^2}{2}\mathbb{E}_s\Bigg[\left\langle \frac{\widehat{\lambda}_{u}^{(k)}}{ \ell^{(k)}}\sum_{\ell=1}^{\ell^{(k)}}\mathscr{N}_u^{\mathsf{G},(k),\ell}, \frac{\widehat{\lambda}_{u'}^{(k)}}{ \ell^{(k)}}\sum_{\ell=1}^{\ell^{(k)}}\mathscr{N}_{u'}^{\mathsf{G},(k),\ell}\right\rangle\Bigg]}_{(x_3)}\nonumber\\
    &~~~~+ \frac{\beta\eta_{_k}^2\left(\ell^{(k)}\right)^2}{2} \mathbb{E}_s\Bigg[\Bigg\Vert  \sum_{u\in \mathcal{U}}\frac{\widehat{\lambda}_{u}^{(k)}}{ \ell^{(k)}}\sum_{\ell=1}^{\ell^{(k)}} {\nabla  \mathfrak{L}_u(\bm{\omega}^{(k),\ell-1}_{u})}\Bigg\Vert^2\Bigg]\nonumber\\
    %%%%%%%
    &\overset{(i)}{=}\mathfrak{L}(\bm{\omega}^{(k)}) - \eta_{_k}\ell^{(k)} \mathbb{E}_s\left[\left\langle\nabla{\mathfrak{L}(\bm{\omega}^{(k)})}, \sum_{u\in \mathcal{U}}\frac{D_{u}}{D \ell^{(k)}}\sum_{\ell=1}^{\ell^{(k)}}  {\nabla  \mathfrak{L}_u(\bm{\omega}^{(k),\ell-1}_{u})}\right\rangle\right]\nonumber\\
    &~~~~+ \eta_{_k}\ell^{(k)} \underbrace{\mathbb{E}_s\left[\left\langle\nabla{\mathfrak{L}(\bm{\omega}^{(k)})},-\mathscr{U}^{(k)} \right\rangle\right]}_{(a)}+ \frac{\beta\eta_{_k}^2\left(\ell^{(k)}\right)^2}{2}  \sum_{u\in \mathcal{U}}\mathbb{E}_s\Bigg[\left(\frac{\widehat{\lambda}_{u}^{(k)}}{ \ell^{(k)}}\right)^2\Bigg\Vert \sum_{\ell=1}^{\ell^{(k)}}\mathscr{N}_u^{\mathsf{G},(k),\ell}\Bigg\Vert^2\Bigg]\nonumber\\
    &~~~~+ \frac{\beta\eta_{_k}^2\left(\ell^{(k)}\right)^2}{2} \mathbb{E}_s\Bigg[\Bigg\Vert \underbrace{ \sum_{u\in \mathcal{U}}\frac{\widehat{\lambda}_{u}^{(k)}}{ \ell^{(k)}}\sum_{\ell=1}^{\ell^{(k)}} {\nabla  \mathfrak{L}_u(\bm{\omega}^{(k),\ell-1}_{u})}}_{(b)}\Bigg\Vert^2\Bigg],
\end{align}
where $(i)$ uses the fact that, term $(x_3)$ is zero, since the noise of gradient estimation of a specific node is independent and zero mean. Using Cauchy-Schwartz and Young's inequalities, for two real valued vectors $\bm{a}$ and $\bm{b}$, we have
\begin{equation}
    \langle\bm{a},\bm{b}\rangle\le \frac{\alpha}{2}\Vert \bm{a}\Vert^2+\frac{1}{2\alpha}\Vert \bm{b}\Vert^2,~ \alpha \in \mathbb{R}^{++},
\end{equation}
Setting $\alpha =2$ in the above inequality implies $\langle\bm{a},\bm{b}\rangle\le \Vert \bm{a}\Vert^2+\frac{1}{4}\Vert \bm{b}\Vert^2$. Using this inequality to expand $(a)$ in \eqref{ineq:main3} and also rewriting $(b)$ in \eqref{ineq:main3} in terms of $\mathscr{U}^{(k)}$, we can further bound \eqref{ineq:main3} as follows:

\begin{align}\label{ineq:main3_1}
    \mathbb{E}_s&\left[\mathfrak{L}(\bm{\omega}^{(k+1)})\right] \leq \mathfrak{L}(\bm{\omega}^{(k)}) - \eta_{_k}\ell^{(k)} \mathbb{E}_s\left[\Bigg\langle\nabla{\mathfrak{L}(\bm{\omega}^{(k)})}, \sum_{u\in \mathcal{U}}\frac{D_{u}}{D \ell^{(k)}}\sum_{\ell=1}^{\ell^{(k)}}  {\nabla  \mathfrak{L}_u(\bm{\omega}^{(k),\ell-1}_{u})}\Bigg\rangle\right]\nonumber\\
    &+ \frac{\beta\eta_{_k}^2\left(\ell^{(k)}\right)^2}{2}  \sum_{u\in \mathcal{U}}\left(\frac{\widehat{\lambda}_{u}^{(k)}}{ \ell^{(k)}}\right)^2\mathbb{E}_s\Bigg[\Bigg\Vert \sum_{\ell=1}^{\ell^{(k)}}\mathscr{N}_u^{\mathsf{G},(k),\ell}\Bigg\Vert^2\Bigg]\nonumber\\
    &+ \frac{\beta\eta_{_k}^2\left(\ell^{(k)}\right)^2}{2} \mathbb{E}_s\Bigg[\Bigg\Vert  \sum_{u\in \mathcal{U}}\frac{D_{u}}{D \ell^{(k)}}\sum_{\ell=1}^{\ell^{(k)}} {\nabla  \mathfrak{L}_u(\bm{\omega}^{(k),\ell-1}_{u})}+\mathscr{U}^{(k)}\Bigg\Vert^2\Bigg]+ \eta_{_k}\ell^{(k)} \mathbb{E}_s\left[\left\langle\nabla{\mathfrak{L}(\bm{\omega}^{(k)})},-\mathscr{U}^{(k)} \right\rangle\right]\nonumber\\
    %%%%%%%
    &\le \mathfrak{L}(\bm{\omega}^{(k)}) - \eta_{_k}\ell^{(k)} \mathbb{E}_s\left[\Bigg\langle\nabla{\mathfrak{L}(\bm{\omega}^{(k)})}, \sum_{u\in \mathcal{U}}\frac{D_{u}}{D \ell^{(k)}}\sum_{\ell=1}^{\ell^{(k)}}  {\nabla  \mathfrak{L}_u(\bm{\omega}^{(k),\ell-1}_{u})}\Bigg\rangle\right]\nonumber\\
    &+ \frac{\beta\eta_{_k}^2\left(\ell^{(k)}\right)^2}{2}  \sum_{u\in \mathcal{U}}\left(\frac{\widehat{\lambda}_{u}^{(k)}}{ \ell^{(k)}}\right)^2\mathbb{E}_s\Bigg[\Bigg\Vert \sum_{\ell=1}^{\ell^{(k)}}\mathscr{N}_u^{\mathsf{G},(k),\ell}\Bigg\Vert^2\Bigg]\nonumber\\
    &+ \frac{\beta\eta_{_k}^2\left(\ell^{(k)}\right)^2}{2} \mathbb{E}_s\Bigg[\underbrace{\Bigg\Vert  \sum_{u\in \mathcal{U}}\frac{D_{u}}{D \ell^{(k)}}\sum_{\ell=1}^{\ell^{(k)}} {\nabla  \mathfrak{L}_u(\bm{\omega}^{(k),\ell-1}_{u})}+\mathscr{U}^{(k)}\Bigg\Vert^2}_{(a)}\Bigg]\nonumber\\
    &~~~~~~~~~~~~~~~~~~~~~~~~~~~~~+ \frac{\eta_{_k}\ell^{(k)}}{4} \mathbb{E}_s\left[\left\Vert\nabla{\mathfrak{L}(\bm{\omega}^{(k)})} \right\Vert^2\right]+ \eta_{_k}\ell^{(k)} \mathbb{E}_s\left[\left\Vert\mathscr{U}^{(k)} \right\Vert^2\right]\nonumber\\
    %%%%%%%
    &\overset{(i)}{\le}\mathfrak{L}(\bm{\omega}^{(k)}) - \eta_{_k} \ell^{(k)}\mathbb{E}_s\left[\underbrace{\left\langle\nabla{\mathfrak{L}(\bm{\omega}^{(k)})}, \sum_{u\in \mathcal{U}}\frac{D_{u}}{D \ell^{(k)}}\sum_{\ell=1}^{\ell^{(k)}}  {\nabla  \mathfrak{L}_u(\bm{\omega}^{(k),\ell-1}_{u})}\right\rangle}_{(b)}\right]\nonumber\\
    &+ \frac{\beta\eta_{_k}^2\left(\ell^{(k)}\right)^2}{2}  \sum_{u\in \mathcal{U}}\left(\frac{\widehat{\lambda}_{u}^{(k)}}{ \ell^{(k)}}\right)^2\mathbb{E}_s\Bigg[\Bigg\Vert \sum_{\ell=1}^{\ell^{(k)}}\mathscr{N}_u^{\mathsf{G},(k),\ell}\Bigg\Vert^2\Bigg]+ \beta\eta_{_k}^2 \left(\ell^{(k)}\right)^2\mathbb{E}_s\Bigg[\Bigg\Vert  \sum_{u\in \mathcal{U}}\frac{D_{u}}{D \ell^{(k)}}\sum_{\ell=1}^{\ell^{(k)}} {\nabla  \mathfrak{L}_u(\bm{\omega}^{(k),\ell-1}_{u})}\Bigg\Vert^2\Bigg]\nonumber\\
    &+ \beta\eta_{_k}^2 \left(\ell^{(k)}\right)^2\mathbb{E}_s\left[\left\Vert\mathscr{U}^{(k)}\right\Vert^2\right]+ \frac{\eta_{_k}\ell^{(k)}}{4} \mathbb{E}_s\left[\left\Vert\nabla{\mathfrak{L}(\bm{\omega}^{(k)})} \right\Vert^2\right]+ \eta_{_k} \ell^{(k)}\mathbb{E}_s\left[\left\Vert\mathscr{U}^{(k)} \right\Vert^2\right],
    %%%%%%%
\end{align}
where in inequality $(i)$ we have used the Cauchy-Schwarz inequality $\Vert \mathbf{a}+\mathbf{b} \Vert^2\leq 2 \Vert \mathbf{a} \Vert^2+2\Vert \mathbf{b} \Vert^2$ to bound $(a)$ in \eqref{ineq:main3_1}. Using $2\langle\bm{a},\bm{b}\rangle=\Vert \bm{a}\Vert^2+\Vert \bm{b}\Vert^2-\Vert \bm{a}-\bm{b}\Vert^2$ to expand $(b)$ in \eqref{ineq:main3_1} and performing some algebraic manipulation, we simplify the above inequality as follows:
\begin{align}
    \mathbb{E}_s&\left[\mathfrak{L}(\bm{\omega}^{(k+1)})\right] \leq \mathfrak{L}(\bm{\omega}^{(k)}) - \frac{\eta_{_k}\ell^{(k)}}{2} \mathbb{E}_s\Bigg[\Bigg\Vert\nabla{\mathfrak{L}(\bm{\omega}^{(k)})}\Bigg\Vert^2+\Bigg\Vert \sum_{u\in \mathcal{U}}\frac{D_{u}}{D \ell^{(k)}}\sum_{\ell=1}^{\ell^{(k)}}  {\nabla  \mathfrak{L}_u(\bm{\omega}^{(k),\ell-1}_{u})}\Bigg\Vert^2\nonumber\\
    &~~~~~~~~~~~~~~-\Bigg\Vert \nabla{\mathfrak{L}(\bm{\omega}^{(k)})}-  \sum_{u\in \mathcal{U}}\frac{D_{u}}{D \ell^{(k)}}\sum_{\ell=1}^{\ell^{(k)}}  {\nabla  \mathfrak{L}_u(\bm{\omega}^{(k),\ell-1}_{u})}\Bigg\Vert^2\Bigg]\nonumber\\
    &+ \frac{\beta\eta_{_k}^2}{2}  \sum_{u\in \mathcal{U}}\left(\frac{\widehat{\lambda}_{u}^{(k)}}{ \ell^{(k)}}\right)^2\mathbb{E}_s\Bigg[\Bigg\Vert \sum_{\ell=1}^{\ell^{(k)}}\mathscr{N}_u^{\mathsf{G},(k),\ell}\Bigg\Vert^2\Bigg]+ \beta\eta_{_k}^2\left(\ell^{(k)}\right)^2 \mathbb{E}_s\Bigg[\Bigg\Vert  \sum_{u\in \mathcal{U}}\frac{D_{u}}{D \ell^{(k)}}\sum_{\ell=1}^{\ell^{(k)}} {\nabla  \mathfrak{L}_u(\bm{\omega}^{(k),\ell-1}_{u})}\Bigg\Vert^2\Bigg]\nonumber\\
    &+ \beta\eta_{_k}^2 \left(\ell^{(k)}\right)^2\mathbb{E}_s\left[\left\Vert\mathscr{U}^{(k)}\right\Vert^2\right]+ \frac{\eta_{_k}\ell^{(k)}}{4} \mathbb{E}_s\left[\left\Vert\nabla{\mathfrak{L}(\bm{\omega}^{(k)})} \right\Vert^2\right]+ \eta_{_k}\ell^{(k)}  \mathbb{E}_s\left[\left\Vert\mathscr{U}^{(k)} \right\Vert^2\right]\nonumber\\
    %%%%%%%
    &=\mathfrak{L}(\bm{\omega}^{(k)}) - \frac{\eta_{_k}\ell^{(k)}}{2} \mathbb{E}_s\Bigg[\Bigg\Vert\nabla{\mathfrak{L}(\bm{\omega}^{(k)})}\Bigg\Vert^2\Bigg]- \frac{\eta_{_k}\ell^{(k)}}{2} \mathbb{E}_s\Bigg[\Bigg\Vert \sum_{u\in \mathcal{U}}\frac{D_{u}}{D \ell^{(k)}}\sum_{\ell=1}^{\ell^{(k)}}  {\nabla  \mathfrak{L}_u(\bm{\omega}^{(k),\ell-1}_{u})}\Bigg\Vert^2\Bigg]\nonumber\\
    &~~~~~~~~~~~~~~+ \frac{\eta_{_k}\ell^{(k)}}{2} \mathbb{E}_s\Bigg[\Bigg\Vert \nabla{\mathfrak{L}(\bm{\omega}^{(k)})}-  \sum_{u\in \mathcal{U}}\frac{D_{u}}{D \ell^{(k)}}\sum_{\ell=1}^{\ell^{(k)}}  {\nabla  \mathfrak{L}_u(\bm{\omega}^{(k),\ell-1}_{u})}\Bigg\Vert^2\Bigg]\nonumber\\
    &+ \frac{\beta\eta_{_k}^2\left(\ell^{(k)}\right)^2}{2}  \sum_{u\in \mathcal{U}}\left(\frac{\widehat{\lambda}_{u}^{(k)}}{ \ell^{(k)}}\right)^2\mathbb{E}_s\Bigg[\Bigg\Vert \sum_{\ell=1}^{\ell^{(k)}}\mathscr{N}_u^{\mathsf{G},(k),\ell}\Bigg\Vert^2\Bigg]+ \beta\eta_{_k}^2\left(\ell^{(k)}\right)^2 \mathbb{E}_s\Bigg[\Bigg\Vert  \sum_{u\in \mathcal{U}}\frac{D_{u}}{D \ell^{(k)}}\sum_{\ell=1}^{\ell^{(k)}} {\nabla  \mathfrak{L}_u(\bm{\omega}^{(k),\ell-1}_{u})}\Bigg\Vert^2\Bigg]\nonumber\\
    &+ \beta\eta_{_k}^2\left(\ell^{(k)}\right)^2 \mathbb{E}_s\left[\left\Vert\mathscr{U}^{(k)}\right\Vert^2\right]+ \frac{\eta_{_k}\ell^{(k)}}{4} \mathbb{E}_s\left[\left\Vert\nabla{\mathfrak{L}(\bm{\omega}^{(k)})} \right\Vert^2\right]+ \eta_{_k} \ell^{(k)} \mathbb{E}_s\left[\left\Vert\mathscr{U}^{(k)} \right\Vert^2\right]\nonumber\\
    %%%%%%%
    &=\mathfrak{L}(\bm{\omega}^{(k)}) - \frac{\eta_{_k}\ell^{(k)}}{4} \mathbb{E}_s\Bigg[\Bigg\Vert\nabla{\mathfrak{L}(\bm{\omega}^{(k)})}\Bigg\Vert^2\Bigg]+\underbrace{\eta_{_k}\left(\beta\eta_{_k}-\frac{1}{2}\right)\ell^{(k)} \mathbb{E}_s\Bigg[\Bigg\Vert \sum_{u\in \mathcal{U}}\frac{D_{u}}{D \ell^{(k)}}\sum_{\ell=1}^{\ell^{(k)}}  {\nabla  \mathfrak{L}_u(\bm{\omega}^{(k),\ell-1}_{u})}\Bigg\Vert^2\Bigg]}_{(a)}\nonumber\\
    &~~~~~~~~~~~~~~+ \frac{\eta_{_k}\ell^{(k)}}{2} \mathbb{E}_s\Bigg[\Bigg\Vert \nabla{\mathfrak{L}(\bm{\omega}^{(k)})}-  \sum_{u\in \mathcal{U}}\frac{D_{u}}{D \ell^{(k)}}\sum_{\ell=1}^{\ell^{(k)}}  {\nabla  \mathfrak{L}_u(\bm{\omega}^{(k),\ell-1}_{u})}\Bigg\Vert^2\Bigg]\nonumber\\
    &+ \frac{\beta\eta_{_k}^2\left(\ell^{(k)}\right)^2}{2}  \sum_{u\in \mathcal{U}}\left(\frac{\widehat{\lambda}_{u}^{(k)}}{ \ell^{(k)}}\right)^2\mathbb{E}_s\Bigg[\Bigg\Vert \sum_{\ell=1}^{\ell^{(k)}}\mathscr{N}_u^{\mathsf{G},(k),\ell}\Bigg\Vert^2\Bigg]+ \eta_{_k}\left(1+\beta\eta_{_k}\right)\ell^{(k)} \mathbb{E}_s\left[\left\Vert\mathscr{U}^{(k)}\right\Vert^2\right].
\end{align}
Assuming 
\begin{equation}\label{ineq:eta_cond1}
 \eta_{_k}\le\frac{1}{2\beta}   
\end{equation}
makes $(a)$ in the above expression negative and thus can be removed from the bound. Moreover, $\eta_{_k}\le\frac{1}{2\beta}$ implies $1+\beta\eta_{_k}\le\frac{3}{2}$. Applying this result to the above bound gives us
\begin{align}\label{ineq:main4_1}
    \mathbb{E}_s&\left[\mathfrak{L}(\bm{\omega}^{(k+1)})\right] \leq \mathfrak{L}(\bm{\omega}^{(k)}) - \frac{\eta_{_k}\ell^{(k)}}{4} \mathbb{E}_s\Bigg[\Bigg\Vert\nabla{\mathfrak{L}(\bm{\omega}^{(k)})}\Bigg\Vert^2\Bigg]\nonumber\\
    &~~~~~~~~~~~~~~+ \frac{\eta_{_k}\ell^{(k)}}{2} \mathbb{E}_s\Bigg[\Bigg\Vert \nabla{\mathfrak{L}(\bm{\omega}^{(k)})}-  \sum_{u\in \mathcal{U}}\frac{D_{u}}{D \ell^{(k)}}\sum_{\ell=1}^{\ell^{(k)}}  {\nabla  \mathfrak{L}_u(\bm{\omega}^{(k),\ell-1}_{u})}\Bigg\Vert^2\Bigg]\nonumber\\
    &+ \frac{\beta\eta_{_k}^2\left(\ell^{(k)}\right)^2}{2}  \sum_{u\in \mathcal{U}}\left(\frac{\widehat{\lambda}_{u}^{(k)}}{ \ell^{(k)}}\right)^2\mathbb{E}_s\Bigg[\underbrace{\Bigg\Vert \sum_{\ell=1}^{\ell^{(k)}}\mathscr{N}_u^{\mathsf{G},(k),\ell}\Bigg\Vert^2}_{(a)}\Bigg]+ \frac{3}{2}\eta_{_k}\ell^{(k)} \mathbb{E}_s\left[\left\Vert\mathscr{U}^{(k)}\right\Vert^2\right].
\end{align}
Using $\Vert \sum_{i=1}^{n}\bm{x}_i\Vert^2=\sum_{i=1}^{n}\Vert\bm{x}_i\Vert^2+\sum_{\substack{j=1\\j\neq i}}^{n}\left\langle\bm{x}_i,\bm{x}_j\right\rangle$ to expand $(a)$ in \eqref{ineq:main4_1} and performing some algebraic manipulation, we rewrite \eqref{ineq:main4_1} as follows: 
\begin{align}\label{ineq:main4}
    \mathbb{E}_s&\left[\mathfrak{L}(\bm{\omega}^{(k+1)})\right] \leq \mathfrak{L}(\bm{\omega}^{(k)}) - \frac{\eta_{_k}\ell^{(k)}}{4} \mathbb{E}_s\Bigg[\Bigg\Vert\nabla{\mathfrak{L}(\bm{\omega}^{(k)})}\Bigg\Vert^2\Bigg]\nonumber\\
    &~~~~~~~~~~~~~~+ \frac{\eta_{_k}\ell^{(k)}}{2} \mathbb{E}_s\Bigg[\Bigg\Vert \nabla{\mathfrak{L}(\bm{\omega}^{(k)})}-  \sum_{u\in \mathcal{U}}\frac{D_{u}}{D \ell^{(k)}}\sum_{\ell=1}^{\ell^{(k)}}  {\nabla  \mathfrak{L}_u(\bm{\omega}^{(k),\ell-1}_{u})}\Bigg\Vert^2\Bigg]+ \frac{3}{2}\eta_{_k}\ell^{(k)} \mathbb{E}_s\left[\left\Vert\mathscr{U}^{(k)}\right\Vert^2\right]\nonumber\\
    &+ \frac{\beta\eta_{_k}^2\left(\ell^{(k)}\right)^2}{2}  \sum_{u\in \mathcal{U}}\left(\frac{\widehat{\lambda}_{u}^{(k)}}{ \ell^{(k)}}\right)^2\left(\mathbb{E}_s\Bigg[\sum_{\ell=1}^{\ell^{(k)}}\Bigg\Vert\mathscr{N}_u^{\mathsf{G},(k),\ell}\Bigg\Vert^2\Bigg]+\underbrace{\mathbb{E}_s\Bigg[\sum_{\ell=1}^{\ell^{(k)}}\sum_{\substack{\ell'=1,\\\ell'\neq \ell}}^{\ell^{(k)}}\left\langle\mathscr{N}_u^{\mathsf{G},(k),\ell},\mathscr{N}_u^{\mathsf{G},(k),\ell'}\right\rangle\Bigg]}_{(x_4)}\right)\nonumber\\
    &\overset{(i)}{=}\mathfrak{L}(\bm{\omega}^{(k)}) - \frac{\eta_{_k}\ell^{(k)}}{4} \mathbb{E}_s\Bigg[\Bigg\Vert\nabla{\mathfrak{L}(\bm{\omega}^{(k)})}\Bigg\Vert^2\Bigg]\nonumber\\
    &~~~~~~~~~~~~~~+ \frac{\eta_{_k}\ell^{(k)}}{2} \mathbb{E}_s\Bigg[\Bigg\Vert \nabla{\mathfrak{L}(\bm{\omega}^{(k)})}-  \sum_{u\in \mathcal{U}}\frac{D_{u}}{D \ell^{(k)}}\sum_{\ell=1}^{\ell^{(k)}}  {\nabla  \mathfrak{L}_u(\bm{\omega}^{(k),\ell-1}_{u})}\Bigg\Vert^2\Bigg]+ \frac{3}{2}\eta_{_k}\ell^{(k)} \mathbb{E}_s\left[\left\Vert\mathscr{U}^{(k)}\right\Vert^2\right]\nonumber\\
    &+ \frac{\beta\eta_{_k}^2\left(\ell^{(k)}\right)^2}{2}  \sum_{u\in \mathcal{U}}\left(\frac{\widehat{\lambda}_{u}^{(k)}}{ \ell^{(k)}}\right)^2\sum_{\ell=1}^{\ell^{(k)}}\mathbb{E}_s\Bigg[\Bigg\Vert\mathscr{N}_u^{\mathsf{G},(k),\ell}\Bigg\Vert^2\Bigg],
\end{align}
where $(i)$ uses the fact that $(x_4)$ is zero since the noise of gradient estimation of a specific node is independent and zero mean. We next aim to simplify $\mathbb{E}_s\left[\left\Vert\mathscr{N}^{\mathsf{G},(k)}_{u}\right\Vert^2\right]$. Considering the result of~\cite{lohr2019sampling} (Chapter 3, Eq. (3.5)), we have 
\begin{align}\label{eq:varGrad}
    \mathbb{E}_s\left[\left\Vert\mathscr{N}^{\mathsf{G},(k)}_{u}\right\Vert^2\right]=\mathbb{E}_s\left[\left\Vert \sum_{d\in \mathcal{B}^{(k),\ell}_{u}} \hspace{-3mm} {\frac{\nabla  f(\bm{\omega}^{(k),\ell-1}_{u},d)}{{B}_{u}}} - \sum_{d\in\mathcal{D}_{u}} \hspace{-3mm} {\frac{\nabla  f(\bm{\omega}^{(k),\ell-1}_{u},d)}{D_{u}}}\right\Vert^2\right]=\left(1-\frac{{B}_{u}}{D_{u}} \right) \frac{\left(\sigma_{u}^{\ell-1}\right)^2}{{B}_{u}},
\end{align}
where $\sigma_{u}^{\ell-1}$ denotes the \textit{variance of the gradients} evaluated at the particular local gradient descent iteration $\ell-1$ for the parameter realization $\bm{\omega}^{(k),\ell-1}_{u}$, and $\left(\sigma_{u}^{\ell{-}1}\right)^2$ is calculated as follows:
\begin{align}\label{eq:dataVar0}
     \left(\sigma_{u}^{\ell{-}1}\right)^2&= \frac{\sum_{d\in\mathcal{D}_{u} } \Big\Vert \nabla  f(\bm{\omega}^{(k),\ell{-}1}_{u},d){-}{\sum_{\tilde{d}\in\mathcal{D}_{u}} }\frac{\nabla  f(\bm{\omega}^{(k),\ell{-}1}_{u},\tilde{d})}{D_{u}}\Big\Vert^2}{D_{u}{-}1}
     \nonumber\\
     &=\frac{\sum_{d\in\mathcal{D}_{u} }\frac{1}{\left(D_{u}\right)^2} \Big\Vert D_{u}\nabla  f(\bm{\omega}^{(D),\ell{-}1}_{u},d){-}{\sum_{\tilde{d}\in\mathcal{D}_{u}} }{\nabla  f(\bm{\omega}^{(k),\ell{-}1}_{u},\tilde{d})}\Big\Vert^2}{D_{u}{-}1}.
\end{align}
Using the Cauchy-Schwarz inequality, we can bound \eqref{eq:dataVar0} as follows:
 \begin{align}\label{eq:dataVar}
     &\left(\sigma_{u}^{\ell{-}1}\right)^2\leq \frac{\sum_{d\in\mathcal{D}_{u} }\frac{D_{u}{-}1}{\left(D_{u}\right)^2} \sum_{\tilde{d}\in\mathcal{D}_{u}} \Big\Vert \nabla  f(\bm{\omega}^{(k),\ell{-}1}_{u},d){-}{\nabla  f(\bm{\omega}^{(k),\ell{-}1}_{u},\tilde{d})}\Big\Vert^2}{D_{u}{-}1}
     \nonumber\\&
     \leq \frac{\sum_{d\in\mathcal{D}_{u} }\frac{(D_{u}{-}1)\Theta^2}{\left(D_{u}\right)^2} \sum_{\tilde{d}\in\mathcal{D}_{u}} \Big\Vert \bm{d}{-}\tilde{\bm{d}}\Big\Vert^2}{D_{u}{-}1}\nonumber\\
     &= \frac{(D_{u}{-}1)\Theta^2}{\left(D_{u}\right)^2}\frac{\sum_{d\in\mathcal{D}_{u} } \sum_{\tilde{d}\in\mathcal{D}_{u}} \Big\Vert \bm{d}{-}\tilde{\bm{d}}{+}\bm{\mu}_{u}{-}\bm{\mu}_{u}\Big\Vert^2}{D_{u}{-}1}\nonumber\\
     &=\frac{(D_{u}{-}1)\Theta^2}{\left(D_{u}\right)^2}\frac{\displaystyle\sum_{d\in\mathcal{D}_{u} } \sum_{\tilde{d}\in\mathcal{D}_{u}} \left[\Big\Vert \bm{d}{-}  \bm{\mu}_{u} \Big\Vert^2 {+} \Big\Vert \tilde{\bm{d}}{-}  \bm{\mu}_{u}\Big\Vert^2 {-} 2\left\langle\bm{d}{-}  \bm{\mu}_{u},\tilde{\bm{d}}{-}  \bm{\mu}_{u}\right\rangle \right]}{D_{u}{-}1}\nonumber\\
     &\overset{(ii)}{=}\frac{(D_{u}{-}1)\Theta^2}{\left(D_{u}\right)^2}\frac{ D_{u} \sum_{d\in\mathcal{D}_{u} } \Big\Vert \bm{d}{-}  \bm{\mu}_{u} \Big\Vert^2 {+}  D_{u} \sum_{\tilde{d}\in\mathcal{D}_{u}} \Big\Vert \tilde{\bm{d}}{-}  \bm{\mu}_{u}\Big\Vert^2}{D_{u}{-}1}\nonumber\\
     &=\frac{2(D_{u}{-}1)\Theta^2}{D_{u}}\left(\sigma_{u}\right)^2, 
\end{align}
where $\bm{\mu}_{u}$ and $\sigma_{u}$ denote the mean and sample variance of data points in dataset $D_{u}$, which are gradient independent. Further, $\Theta {=} \max_{u{\in}\mathcal{U}}\{\Theta_{u} \}$. Also, $\bm{d}$ refers to the feature vector of data point $d$. Furthermore, $(ii)$ used the fact that $\sum_{d\in\mathcal{D}_{u} } (\bm{d}-  \bm{\mu}_{u}) =\bm{0}$. Replacing the above result in~\eqref{eq:varGrad}, inequality \eqref{ineq:main4} can be written as follows:
\begin{align}\label{ineq:main5}
    \mathbb{E}_s&\left[\mathfrak{L}(\bm{\omega}^{(k+1)})\right] \leq \mathfrak{L}(\bm{\omega}^{(k)}) - \frac{\eta_{_k}\ell^{(k)}}{4} \mathbb{E}_s\Bigg[\Bigg\Vert\nabla{\mathfrak{L}(\bm{\omega}^{(k)})}\Bigg\Vert^2\Bigg]\nonumber\\
    &~~~~+ \frac{\eta_{_k}\ell^{(k)}}{2} \underbrace{\mathbb{E}_s\Bigg[\Bigg\Vert \nabla{\mathfrak{L}(\bm{\omega}^{(k)})}-  \sum_{u\in \mathcal{U}}\frac{D_{u}}{D \ell^{(k)}}\sum_{\ell=1}^{\ell^{(k)}}  {\nabla  \mathfrak{L}_u(\bm{\omega}^{(k),\ell-1}_{u})}\Bigg\Vert^2\Bigg]}_{(d)}+ \frac{3}{2}\eta_{_k}\ell^{(k)} \mathbb{E}_s\left[\left\Vert\mathscr{U}^{(k)}\right\Vert^2\right]\nonumber\\
    &~~~~+ \frac{\beta\eta_{_k}^2\left(\ell^{(k)}\right)^2}{2}  \sum_{u\in \mathcal{U}}\frac{ \widehat{\lambda}_{u}^{(k)} }{  \ell^{(k)}}\left(1-\frac{{B}_{u}}{D_{u}} \right) \frac{2(D_{u}-1)\Theta^2\left(\sigma_{u}\right)^2}{D_{u}{B}_{u}}.
\end{align}
In the following, we aim to bound term $(d)$.
\begin{align}\label{eq:res3}
        (d) &=\mathbb{E}_s\left[\left\Vert \nabla{\mathfrak{L}(\bm{\omega}^{(k)})}- \sum_{u\in \mathcal{U}}\frac{D_{u}}{D \ell^{(k)}} \sum_{\ell=1}^{\ell^{(k)}}  {\nabla  \mathfrak{L}_u(\bm{\omega}^{(k),\ell-1}_{u})}\right\Vert^2\right]\nonumber\\
    &\overset{(i)}{\leq}    \sum_{u\in \mathcal{U}}\frac{D_{u}}{D }\mathbb{E}_s\left[\left\Vert \nabla{\mathfrak{L}_u(\bm{\omega}^{(k)})}-  \frac{1}{\ell^{(k)}} \sum_{\ell=1}^{\ell^{(k)}}  {\nabla  \mathfrak{L}_u(\bm{\omega}^{(k),\ell-1}_{u})} \right\Vert^2\right]\nonumber\\
    &\overset{(ii)}{\leq}  \sum_{u\in \mathcal{U}}\frac{D_{u}}{D \ell^{(k)}}\sum_{\ell=1}^{\ell^{(k)}} \mathbb{E}_s\left[\left\Vert
   \nabla{\mathfrak{L}_u(\bm{\omega}^{(k)})}- {\nabla  \mathfrak{L}_u(\bm{\omega}^{(k),\ell-1}_{u})}\right\Vert^2\right] \nonumber\\
   &\overset{(iii)}{\leq} \beta^2 \sum_{u\in \mathcal{U}}\frac{D_{u}}{D \ell^{(k)}}\sum_{\ell=1}^{\ell^{(k)}} \mathbb{E}_s\left[\left\Vert\bm{\omega}^{(k)}-\bm{\omega}^{(k),\ell-1}_{u}\right\Vert^2\right],
\end{align}
where in inequalities $(i)$ and $(ii)$ in~\eqref{eq:res3} we used Jenson's inequality. Inequality $(iii)$ is the result of Assumption \eqref{Assup:lossFun}. To bound $\mathbb{E}_s\left[\left\Vert\bm{\omega}^{(k)}-\bm{\omega}^{(k),\ell-1}_{u}\right\Vert^2\right]$, we take the following steps.
\begin{align}\label{eq:res4}
    &\mathbb{E}_s\left[\left\Vert\bm{\omega}^{(k)}-\bm{\omega}^{(k),\ell-1}_{u}\right\Vert^2\right]\overset{(i)}{=}\eta_k^2  \mathbb{E}_s\left[\left\Vert
    \sum_{\ell'=1}^{\ell-1}\sum_{d\in \mathcal{B}^{(k),\ell'}_{u}} \hspace{-1mm} {\frac{\nabla  f(\bm{\omega}^{(k),\ell'-1}_{u},d)}{{B}_{u}}}\right\Vert^2 \right]
    \nonumber\\
    %%%%%%
    &= \eta_k^2 \mathbb{E}_s\vast[\Bigg\Vert\sum_{\ell'=1}^{\ell-1} \sum_{d\in \mathcal{B}^{(k),\ell'}_{u}} \hspace{-1mm} {\frac{\nabla  f(\bm{\omega}^{(k),\ell'-1}_{u},d)}{{B}_{u}}}\nonumber-\frac{1}{D_{u}} \sum_{\ell'=1}^{\ell-1} \sum_{d\in\mathcal{D}_{u}}\nabla  f(\bm{\omega}^{(k),\ell'-1}_{u},d) + \frac{1}{D_{u}} \sum_{\ell'=1}^{\ell-1}\sum_{d\in\mathcal{D}_{u}}\nabla  f(\bm{\omega}^{(k),\ell'-1}_{u},d)\Bigg\Vert^2 \vast]\nonumber \nonumber\\
    %%%%%%
    & \overset{(ii)}{\leq }2 \eta_k^2  \mathbb{E}_s\left[\Bigg\Vert\sum_{\ell'=1}^{\ell-1} \sum_{d\in \mathcal{B}^{(k),\ell'}_{u}} \hspace{-1mm} {\frac{\nabla  f(\bm{\omega}^{(k),\ell'-1}_{u},d)}{{B}_{u}}} - \frac{1}{D_{u}} \sum_{\ell'=1}^{\ell-1}\sum_{d\in\mathcal{D}_{u}}\nabla f(\bm{\omega}^{(k),\ell'-1}_{u},d)\Bigg\Vert^2\right]  +2 \eta_k^2 \mathbb{E}_s\left[\Bigg\Vert  \frac{1}{D_{u}} \sum_{\ell'=1}^{\ell-1}\sum_{d\in\mathcal{D}_{u}}\nabla  f(\bm{\omega}^{(k),\ell'-1}_{u},d)\Bigg\Vert^2 \right]\nonumber \nonumber\\
    %%%%%%
    &\overset{(iii)}{=} \underbrace{2 \eta_k^2 \sum_{\ell'=1}^{\ell-1}\mathbb{E}_s\left[\Bigg\Vert\sum_{d\in \mathcal{B}^{(k),\ell'}_{u}} \hspace{-1mm} {\frac{\nabla  f(\bm{\omega}^{(k),\ell'-1}_{u},d)}{{B}_{u}}} - \frac{1}{D_{u}} \sum_{d\in\mathcal{D}_{u}}\nabla f(\bm{\omega}^{(k),\ell'-1}_{u},d)\Bigg\Vert^2\right]}_{(e)}  +\underbrace{2 \eta_k^2 \mathbb{E}_s\left[\Bigg\Vert  \frac{1}{D_{u}} \sum_{\ell'=1}^{\ell-1}\sum_{d\in\mathcal{D}_{u}}\nabla  f(\bm{\omega}^{(k),\ell'-1}_{u},d)\Bigg\Vert^2\right]}_{(f)},
\end{align}
where to obtain~\eqref{eq:res4}, in equality $(i)$ we used~\eqref{eq:updaterule}, in inequality $(ii)$ we used  Cauchy–Schwarz inequality, and $(iii)$
uses the fact that each local gradient estimation is unbiased (i.e., zero mean) conditioned on its own local parameter and the law of total expectation (across the mini-batches $\ell'$). Similar to~\eqref{eq:varGrad}, using~\eqref{eq:dataVar} we upper bound term $(e)$ in~\eqref{eq:res4} as follows:
\begin{equation}\label{eq:A2}
    (e) \leq 4 \Theta^2 \eta_k^2 \sum_{\ell'=1}^{\ell-1} \left(1-\frac{{B}_{u}}{D_{u}} \right)  \frac{{(D_{u}-1)}
     \left(\sigma_{u}\right)^2}{D_{u}{B}_{u}}.
\end{equation}
Also, for term $(f)$, we have
\begin{align}\label{eq:B2}
   (f)&
   \overset{(i)}{\leq} 2 \eta_k^2 (\ell-1)\sum_{\ell'=1}^{\ell-1}\mathbb{E}_s\left[\Bigg\Vert \frac{1}{D_{u}} \sum_{d\in\mathcal{D}_{u}}\nabla  f(\bm{\omega}^{(k),\ell'-1}_{u},d) - \nabla \mathfrak{L}_u(\bm{\omega}^{(k)})+ \nabla \mathfrak{L}_u(\bm{\omega}^{(k)})
    \Bigg\Vert^2 \right]
   \nonumber \nonumber\\
    &\overset{(ii)}{\leq} 4 \eta_k^2 (\ell-1)\sum_{\ell'=1}^{\ell-1}\mathbb{E}_s\left[\Bigg\Vert \nabla  \mathfrak{L}_u(\bm{\omega}^{(k),\ell'-1}_{u}) - \nabla \mathfrak{L}_u(\bm{\omega}^{(k)})\Bigg\Vert^2\right] + 4 \eta_k^2 (\ell-1)\sum_{\ell'=1}^{\ell-1} \Bigg\Vert\nabla \mathfrak{L}_u(\bm{\omega}^{(k)})
    \Bigg\Vert^2 
    \nonumber \nonumber\\
    &\leq 4\eta_k^2\beta^2 (\ell-1) \sum_{\ell'=1}^{\ell-1}\mathbb{E}_s\left[\Bigg\Vert \bm{\omega}^{(k),\ell'-1}_{u} - \bm{\omega}^{(k)}\Bigg\Vert^2\right]+ 4 \eta_k^2 (\ell-1)\sum_{\ell'=1}^{\ell-1} \Bigg\Vert\nabla \mathfrak{L}_u(\bm{\omega}^{(k)})
    \Bigg\Vert^2 ,
\end{align}
where inequalities $(i)$ and $(ii)$ are obtained via Cauchy-Schwarz inequality. Replacing the result of~\eqref{eq:A2} and~\eqref{eq:B2} back in~\eqref{eq:res4} we have
\begin{align}
  \mathbb{E}_s\left[  \left\Vert
  \bm{\omega}^{(k)}-
     \bm{\omega}^{(k),\ell-1}_{u}
    \right\Vert^2\right] \leq& 4 \Theta^2 \eta_k^2 \sum_{\ell'=1}^{\ell-1} \left(1-\frac{{B}_{u}}{D_{u}} \right)  \frac{{(D_{u}-1)}
     \left(\sigma_{u}\right)^2}{D_{u}{B}_{u}} \nonumber \nonumber\\
     & + 4\eta_k^2\beta^2 (\ell-1) \sum_{\ell'=1}^{\ell-1}\mathbb{E}_s\left[\Bigg\Vert \bm{\omega}^{(k),\ell'-1}_{u} - \bm{\omega}^{(k)}\Bigg\Vert^2\right]+ 4 \eta_k^2 (\ell-1)\sum_{\ell'=1}^{\ell-1} \Bigg\Vert\nabla \mathfrak{L}_u(\bm{\omega}^{(k)})
    \Bigg\Vert^2, 
\end{align}
which implies
\begin{align}
   &\sum_{\ell=1}^{\ell^{(k)}} \mathbb{E}_s\left[\left\Vert\bm{\omega}^{(k)}-\bm{\omega}^{(k),\ell-1}_{u}\right\Vert^2\right] \leq 4 \Theta^2 \eta_k^2\sum_{\ell=1}^{\ell^{(k)}} \sum_{\ell'=1}^{\ell-1} \left(1-\frac{{B}_{u}}{D_{u}} \right)  \frac{{(D_{u}-1)} \left(\sigma_{u}\right)^2}{D_{u}{B}_{u}} \nonumber \\
     & + 4\eta_k^2\beta^2 \sum_{\ell=1}^{\ell^{(k)}}(\ell-1) \sum_{\ell'=1}^{\ell-1}\mathbb{E}_s\left[\Bigg\Vert \bm{\omega}^{(k),\ell'-1}_{u} - \bm{\omega}^{(k)}\Bigg\Vert^2\right]+ 4 \eta_k^2 \sum_{\ell=1}^{\ell^{(k)}}(\ell-1)\sum_{\ell'=1}^{\ell-1} \Bigg\Vert\nabla \mathfrak{L}_u(\bm{\omega}^{(k)}) \Bigg\Vert^2\nonumber\\
     &\leq 4 \Theta^2 \eta_k^2 \left(\ell^{(k)}\right)\left(\ell^{(k)}-1\right) \left(1-\frac{{B}_{u}}{D_{u}} \right)  \frac{{(D_{u}-1)}\left(\sigma_{u}\right)^2}{D_{u}{B}_{u}} \nonumber\\
     & + 4\eta_k^2\beta^2 \left(\ell^{(k)}\right)\left(\ell^{(k)}-1\right) \sum_{\ell=1}^{\ell^{(k)}}\mathbb{E}_s\left[\Bigg\Vert \bm{\omega}^{(k),\ell-1}_{u} - \bm{\omega}^{(k)}\Bigg\Vert^2\right]+ 4 \eta_k^2 \left(\ell^{(k)}\right)\left(\ell^{(k)}-1\right) \sum_{\ell=1}^{\ell^{(k)}} \Bigg\Vert\nabla \mathfrak{L}_u(\bm{\omega}^{(k)})\Bigg\Vert^2.
\end{align}
Assuming $\eta_k \leq \left(2\beta\sqrt{\ell^{(k)}(\ell^{(k)}-1)}\right)^{-1},\forall u$, the above inequality implies
\begin{align}\label{eq:f_bound}
    \sum_{\ell=1}^{\ell^{(k)}} \mathbb{E}_s\left[\left\Vert\bm{\omega}^{(k)}- \bm{\omega}^{(k),\ell-1}_{u} \right\Vert^2\right] \leq&\frac{4 \Theta^2 \eta_k^2 \ell^{(k)}\left(\ell^{(k)}-1\right)}{1- 4\eta_k^2\beta^2 \ell^{(k)}\left(\ell^{(k)}-1\right)} \left(1-\frac{{B}_{u}}{D_{u}} \right)  \frac{{(D_{u}-1)}\left(\sigma_{u}\right)^2}{D_{u}{B}_{u}} + \frac{4 \eta_k^2 \left(\ell^{(k)}\right)^2\left(\ell^{(k)}-1\right)}{1- 4\eta_k^2\beta^2 \ell^{(k)}\left(\ell^{(k)}-1\right)}  \Bigg\Vert\nabla \mathfrak{L}_u(\bm{\omega}^{(k)})\Bigg\Vert^2.
\end{align}
Replacing this result back in~\eqref{eq:res3} we have
\begin{align}\label{eq:res5}
    (d) &\leq {4 \beta^2\Theta^2 \eta_k^2} \sum_{u\in \mathcal{U}}\frac{D_{u}}{D \ell^{(k)}}\frac{\left(\ell^{(k)}\right)\left(\ell^{(k)}-1\right)}{1- 4\eta_k^2\beta^2 \ell^{(k)}\left(\ell^{(k)}-1\right)}\left(1-\frac{{B}_{u}}{D_{u}} \right)  \frac{{(D_{u}-1)}
    \left(\sigma_{u}\right)^2}{D_{u}{B}_{u}}\nonumber  \nonumber\\
    %%%%%%%%%
    &+ {4 \eta_k^2\beta^2 } \sum_{u\in \mathcal{U}}\frac{D_{u}}{D \ell^{(k)}}\frac{\left(\ell^{(k)}\right)^2\left(\ell^{(k)}-1\right)}{{1- 4\eta_k^2\beta^2 \ell^{(k)}\left(\ell^{(k)}-1\right)}}  \Bigg\Vert\nabla \mathfrak{L}_u(\bm{\omega}^{(k)}) \Bigg\Vert^2\nonumber\\
    &= {4 \beta^2\Theta^2 \eta_k^2} \sum_{u\in \mathcal{U}}\frac{D_{u}}{D \ell^{(k)}}\frac{\left(\ell^{(k)}\right)\left(\ell^{(k)}-1\right)}{1- 4\eta_k^2\beta^2 \ell^{(k)}\left(\ell^{(k)}-1\right)}\left(1-\frac{{B}_{u}}{D_{u}} \right)  \frac{{(D_{u}-1)}
    \left(\sigma_{u}\right)^2}{D_{u}{B}_{u}}\nonumber  \nonumber\\
    %%%%%%%%%
    &+ {4 \eta_k^2\beta^2 } \sum_{u\in \mathcal{U}}\frac{D_{u}}{D \ell^{(k)}}\frac{\left(\ell^{(k)}\right)^2\left(\ell^{(k)}-1\right)}{{1- 4\eta_k^2\beta^2 \ell^{(k)}\left(\ell^{(k)}-1\right)}}  \Bigg\Vert\nabla \mathfrak{L}_u(\bm{\omega}^{(k)}) - \nabla \mathfrak{L}(\bm{\omega}^{(k)}) + \nabla \mathfrak{L}(\bm{\omega}^{(k)}) \Bigg\Vert^2\nonumber\\
    &\le {4 \beta^2\Theta^2 \eta_k^2} \sum_{u\in \mathcal{U}}\frac{D_{u}}{D \ell^{(k)}}\frac{\left(\ell^{(k)}\right)\left(\ell^{(k)}-1\right)}{1- 4\eta_k^2\beta^2 \ell^{(k)}\left(\ell^{(k)}-1\right)}\left(1-\frac{{B}_{u}}{D_{u}} \right)  \frac{{(D_{u}-1)}
    \left(\sigma_{u}\right)^2}{D_{u}{B}_{u}}  \nonumber\\
    %%%%%%%%%
    &+ {8 \eta_k^2\beta^2 } \sum_{u\in \mathcal{U}}\frac{D_{u}}{D}\frac{\left(\ell^{(k)}\right) \left(\ell^{(k)}-1\right)}{{1- 4\eta_k^2\beta^2 \ell^{(k)}\left(\ell^{(k)}-1\right)}}  \Bigg\Vert\nabla \mathfrak{L}_u(\bm{\omega}^{(k)}) - \nabla \mathfrak{L}(\bm{\omega}^{(k)}) \Bigg\Vert^2\nonumber\\
    &+ {8 \eta_k^2\beta^2 } \sum_{u\in \mathcal{U}}\frac{D_{u}}{D}\frac{\left(\ell^{(k)}\right) \left(\ell^{(k)}-1\right)}{{1- 4\eta_k^2\beta^2 \ell^{(k)}\left(\ell^{(k)}-1\right)}}  \Bigg\Vert  \nabla \mathfrak{L}(\bm{\omega}^{(k)}) \Bigg\Vert^2 
\end{align}
Performing some algebraic manipulations give us
 \begin{align}\label{eq:res09444}
    (d) \leq &{4 \beta^2\Theta^2 \eta_k^2} \sum_{u\in \mathcal{U}}\frac{D_{u}}{D \ell^{(k)}}\frac{\left(\ell^{(k)}\right)\left(\ell^{(k)}-1\right)}{1- 4\eta_k^2\beta^2 \ell^{(k)}\left(\ell^{(k)}-1\right)}\left(1-\frac{{B}_{u}}{D_{u}} \right)  \frac{{(D_{u}-1)}
    \left(\sigma_{u}\right)^2}{D_{u}{B}_{u}} \nonumber\\
    %%%%%%%%%
    &+ {8 \eta_k^2\beta^2 } \sum_{u\in \mathcal{U}}\frac{D_{u}}{D}\frac{\left(\ell^{(k)}\right) \left(\ell^{(k)}-1\right)}{{1- 4\eta_k^2\beta^2 \ell^{(k)}\left(\ell^{(k)}-1\right)}}  \mathfrak{X}^{(k)}_u +  \frac{{8 \eta_k^2\beta^2 }\left(\ell^{(k)}\right) \left(\ell^{(k)}-1\right)}{{1- 4\eta_k^2\beta^2 \ell^{(k)}\left(\ell^{(k)}-1\right)}}  \Bigg\Vert  \nabla \mathfrak{L}(\bm{\omega}^{(k)}) \Bigg\Vert^2 
\end{align}

 Replacing the above result back in~\eqref{ineq:main5} and gathering the terms leads to
 \begin{align}\label{ineq:main6}
    \mathbb{E}_s&\left[\mathfrak{L}(\bm{\omega}^{(k+1)})\right] \leq \mathfrak{L}(\bm{\omega}^{(k)}) - \frac{\eta_{_k}\ell^{(k)}}{4} \mathbb{E}_s\Bigg[\Bigg\Vert\nabla{\mathfrak{L}(\bm{\omega}^{(k)})}\Bigg\Vert^2\Bigg] +     \sum_{u\in \mathcal{U}}\frac{D_{u}}{D \ell^{(k)}}\frac{{\ell^{(k)}  2 \beta^2\Theta^2 \eta_k^3}\left(\ell^{(k)}\right)\left(\ell^{(k)}-1\right)}{1- 4\eta_k^2\beta^2 \ell^{(k)}\left(\ell^{(k)}-1\right)}\left(1-\frac{{B}_{u}}{D_{u}} \right)  \frac{{(D_{u}-1)}
    \left(\sigma_{u}\right)^2}{D_{u}{B}_{u}}\nonumber  \nonumber\\
    %%%%%%%%%
    &+    \sum_{u\in \mathcal{U}}\frac{D_{u}}{D}\frac{ {4 \eta_k^3 \ell^{(k)} \beta^2 }\left(\ell^{(k)}\right) \left(\ell^{(k)}-1\right)}{{1- 4\eta_k^2\beta^2 \ell^{(k)}\left(\ell^{(k)}-1\right)}}  \mathfrak{X}^{(k)}_u +    \frac{{4 \eta_k^3\ell^{(k)}\beta^2 }\left(\ell^{(k)}\right) \left(\ell^{(k)}-1\right)}{{1- 4\eta_k^2\beta^2 \ell^{(k)}\left(\ell^{(k)}-1\right)}}  \Bigg\Vert  \nabla \mathfrak{L}(\bm{\omega}^{(k)}) \Bigg\Vert^2 + \frac{3}{2}\eta_{_k}\ell^{(k)} \mathbb{E}_s\left[\left\Vert\mathscr{U}^{(k)}\right\Vert^2\right]\nonumber\\
    &~~~~+ \frac{\beta\eta_{_k}^2\left(\ell^{(k)}\right)^2}{2}  \sum_{u\in \mathcal{U}}\frac{ \widehat{\lambda}_{u}^{(k)} }{  \ell^{(k)}}\left(1-\frac{{B}_{u}}{D_{u}} \right) \frac{2(D_{u}-1)\Theta^2\left(\sigma_{u}\right)^2}{D_{u}{B}_{u}}.
\end{align}

Considering the definition of the noise of node selection presented in \eqref{eq:node_selection_noise}, we next aim to bound $\mathbb{E}_s\left[\left\Vert\mathscr{U}^{(k)}\right\Vert^2\right]$ in the above inequality.
\begin{align}
    \mathbb{E}_s\left[\left\Vert\mathscr{U}^{(k)}\right\Vert^2\right]&= \mathbb{E}_s\left[\left\Vert \sum_{u\in \mathcal{U}}\frac{\widehat{\lambda}_{u}^{(k)}}{ \ell^{(k)}}\sum_{\ell=1}^{\ell^{(k)}} {\nabla  \mathfrak{L}_u(\bm{\omega}^{(k),\ell-1}_{u})}- \sum_{u\in \mathcal{U}}\frac{D_{u}}{D \ell^{(k)}}\sum_{\ell=1}^{\ell^{(k)}} {\nabla  \mathfrak{L}_u(\bm{\omega}^{(k),\ell-1}_{u})} \right\Vert^2\right]\nonumber\\
    &= \mathbb{E}_s\left[\left\Vert \sum_{u\in \mathcal{U}} \left(\frac{\widehat{\lambda}_{u}^{(k)}D-D_{u}}{\ell^{(k)}D}\right)\sum_{\ell=1}^{\ell^{(k)}} {\nabla  \mathfrak{L}_u(\bm{\omega}^{(k),\ell-1}_{u})} \right\Vert^2\right]\nonumber\\
    &\overset{(i)}{\le}  |\mathcal{U}|  \sum_{u\in \mathcal{U}}\frac{\left(\frac{\widehat{\lambda}_{u}^{(k)}D-D_{u}}{D}\right)^2}{\left(\ell^{(k)}\right)^2}\underbrace{\mathbb{E}_s\left[\left\Vert \sum_{\ell=1}^{\ell^{(k)}} {\nabla  \mathfrak{L}_u(\bm{\omega}^{(k),\ell-1}_{u})} \right\Vert^2\right]}_{(a)}
\end{align}
where (i) uses Cauchy-Schwarz inequality. Using the same technique utilized in \eqref{eq:B2} and performing some algebraic manipulation give us an upper bound for $(a)$ in the above inequality. Let $\Delta_{D_u}^{(k)}= \left(\frac{\widehat{\lambda}_{u}^{(k)}D-D_{u}}{D}\right)^2$. Using the above result, we can further bound $\mathbb{E}_s\left[\left\Vert\mathscr{U}^{(k)}\right\Vert^2\right]$ as follows:  
\begin{align}
    \hspace{-2.5mm}\mathbb{E}_s\left[\left\Vert\mathscr{U}^{(k)}\right\Vert^2\right]&\le \mathbb{E}_s\Bigg[  |\mathcal{U}|\sum_{u\in \mathcal{U}} \left(\frac{2\beta^2 \Delta_{D_u}^{(k)}}{(\ell^{(k)})} \sum_{\ell=1}^{\ell^{(k)}}\Bigg\Vert \bm{\omega}^{(k)}-\bm{\omega}^{(k),\ell-1}_{u}\Bigg\Vert^2+ \frac{2 \Delta_{D_u}^{(k)}}{(\ell^{(k)})}\sum_{\ell=1}^{\ell^{(k)}} \Bigg\Vert\nabla \mathfrak{L}_u(\bm{\omega}^{(k)})\Bigg\Vert^2\right)\Bigg]\nonumber\\
    & =  2 |\mathcal{U}|\sum_{u\in \mathcal{U}} \left(\frac{\beta^2 \Delta_{D_u}^{(k)}}{(\ell^{(k)})} \sum_{\ell=1}^{\ell^{(k)}}\mathbb{E}_s\left[\Bigg\Vert \bm{\omega}^{(k)}-\bm{\omega}^{(k),\ell-1}_{u}\Bigg\Vert^2\right]+ \Delta_{D_u}^{(k)}\Bigg\Vert\nabla \mathfrak{L}_u(\bm{\omega}^{(k)})\Bigg\Vert^2\right)
\end{align}

Substituting $\sum_{\ell=1}^{\ell^{(k)}}\mathbb{E}_s\left[\left\Vert \bm{\omega}^{(k)}-\bm{\omega}^{(k),\ell-1}_{u}\right\Vert^2 \right]$ from \eqref{eq:f_bound} into the above inequality and performing some algebraic operations give us

\begin{align}
    \mathbb{E}_s\left[\left\Vert\mathscr{U}^{(k)}\right\Vert^2\right] &\le 2 |\mathcal{U}| \sum_{u\in \mathcal{U}}  \Bigg( \frac{\beta^2 \Delta_{D_u}^{(k)}}{(\ell^{(k)})}\frac{4 \Theta^2 \eta_k^2 \ell^{(k)}\left(\ell^{(k)}-1\right)}{1- 4\eta_k^2\beta^2 \ell^{(k)}\left(\ell^{(k)}-1\right)} \left(1-\frac{{B}_{u}}{D_{u}} \right)  \frac{{(D_{u}-1)}\left(\sigma_{u}\right)^2}{D_{u}{B}_{u}}\nonumber\\
     & ~~~~~~~~~~~~+ \frac{\beta^2 \Delta_{D_u}^{(k)}}{(\ell^{(k)})}\frac{4 \eta_k^2 \left(\ell^{(k)}\right)^2\left(\ell^{(k)}-1\right)}{1- 4\eta_k^2\beta^2 \ell^{(k)}\left(\ell^{(k)}-1\right)}  \Bigg\Vert\nabla \mathfrak{L}_u(\bm{\omega}^{(k)})\Bigg\Vert^2+ \Delta_{D_u}^{(k)}\Bigg\Vert\nabla \mathfrak{L}_u(\bm{\omega}^{(k)})\Bigg\Vert^2\Bigg)\nonumber\\
     &\le 2 |\mathcal{U}| \sum_{u\in \mathcal{U}}\Bigg( \frac{\beta^2 \Delta_{D_u}^{(k)}}{(\ell^{(k)})}\frac{4 \Theta^2 \eta_k^2 \ell^{(k)}\left(\ell^{(k)}-1\right)}{1- 4\eta_k^2\beta^2 \ell^{(k)}\left(\ell^{(k)}-1\right)} \left(1-\frac{{B}_{u}}{D_{u}} \right)  \frac{{(D_{u}-1)}\left(\sigma_{u}\right)^2}{D_{u}{B}_{u}}\nonumber\\
     & ~~~~~~~~~~~~+ \Delta_{D_u}^{(k)}\left(\frac{4 \eta_k^2 \beta^2  \ell^{(k)}\left(\ell^{(k)}-1\right)}{1- 4\eta_k^2\beta^2 \ell^{(k)}\left(\ell^{(k)}-1\right)}+1\right)  \Bigg\Vert\nabla \mathfrak{L}_u(\bm{\omega}^{(k)})\Bigg\Vert^2\Bigg)\nonumber\\
     &\le 2 |\mathcal{U}| \sum_{u\in \mathcal{U}}\Bigg( \frac{\beta^2 \Delta_{D_u}^{(k)}}{(\ell^{(k)})}\frac{4 \Theta^2 \eta_k^2 \ell^{(k)}\left(\ell^{(k)}-1\right)}{1- 4\eta_k^2\beta^2 \ell^{(k)}\left(\ell^{(k)}-1\right)} \left(1-\frac{{B}_{u}}{D_{u}} \right)  \frac{{(D_{u}-1)}\left(\sigma_{u}\right)^2}{D_{u}{B}_{u}}\nonumber\\
     & ~~~~~~~~~~~~+ \Delta_{D_u}^{(k)}\left(\frac{4 \eta_k^2 \beta^2 \ell^{(k)}\left(\ell^{(k)}-1\right)+1- 4\eta_k^2\beta^2 \ell^{(k)}\left(\ell^{(k)}-1\right)}{1- 4\eta_k^2\beta^2 \ell^{(k)}\left(\ell^{(k)}-1\right)}\right)  \Bigg\Vert\nabla \mathfrak{L}_u(\bm{\omega}^{(k)})\Bigg\Vert^2\Bigg)\nonumber\\
     &\le 2 |\mathcal{U}| \sum_{u\in \mathcal{U}}\Bigg( \frac{\beta^2 \Delta_{D_u}^{(k)}}{(\ell^{(k)})}\frac{4 \Theta^2 \eta_k^2 \ell^{(k)}\left(\ell^{(k)}-1\right)}{1- 4\eta_k^2\beta^2 \ell^{(k)}\left(\ell^{(k)}-1\right)} \left(1-\frac{{B}_{u}}{D_{u}} \right)  \frac{{(D_{u}-1)}\left(\sigma_{u}\right)^2}{D_{u}{B}_{u}}\nonumber\\
     & ~~~~~~~~~~~~+ \left(\frac{\Delta_{D_u}^{(k)}}{1- 4\eta_k^2\beta^2 \ell^{(k)}\left(\ell^{(k)}-1\right)}\right)  \Bigg\Vert\nabla \mathfrak{L}_u(\bm{\omega}^{(k)})\Bigg\Vert^2\Bigg)\nonumber\\
     &\le 2 |\mathcal{U}| \sum_{u\in \mathcal{U}}\Bigg(\frac{4 \Theta^2 \beta^2 \Delta_{D_u}^{(k)} \eta_k^2 \left(\ell^{(k)}-1\right)}{1- 4\eta_k^2\beta^2 \ell^{(k)}\left(\ell^{(k)}-1\right)} \left(1-\frac{{B}_{u}}{D_{u}} \right)  \frac{{(D_{u}-1)}\left(\sigma_{u}\right)^2}{D_{u}{B}_{u}}\Bigg)\nonumber\\
     &+2 |\mathcal{U}| \sum_{u\in \mathcal{U}}\left(\frac{\Delta_{D_u}^{(k)}}{1- 4\eta_k^2\beta^2 \ell^{(k)}\left(\ell^{(k)}-1\right)}\right) \Bigg\Vert\nabla \mathfrak{L}_u(\bm{\omega}^{(k)})\Bigg\Vert^2.
\end{align}
Doing simplifications and gathering terms give us
\begin{align}
    \mathbb{E}_s\left[\left\Vert\mathscr{U}^{(k)}\right\Vert^2\right]&\le 2 |\mathcal{U}| \sum_{u\in \mathcal{U}}\Bigg(\frac{4 \Theta^2 \beta^2 \Delta_{D_u}^{(k)} \eta_k^2 \left(\ell^{(k)}-1\right)}{1- 4\eta_k^2\beta^2 \ell^{(k)}\left(\ell^{(k)}-1\right)} \left(1-\frac{{B}_{u}}{D_{u}} \right)  \frac{{(D_{u}-1)}\left(\sigma_{u}\right)^2}{D_{u}{B}_{u}}\Bigg)\nonumber\\
     &+2 |\mathcal{U}| \sum_{u\in \mathcal{U}}\left(\frac{\Delta_{D_u}^{(k)}}{1- 4\eta_k^2\beta^2 \ell^{(k)}\left(\ell^{(k)}-1\right)}\right)\Bigg\Vert\nabla \mathfrak{L}_u(\bm{\omega}^{(k)})-\nabla \mathfrak{L}(\bm{\omega}^{(k)})+\nabla \mathfrak{L}(\bm{\omega}^{(k)})\Bigg\Vert^2\nonumber\\
     &\overset{(i)}{\le} 2 |\mathcal{U}| \sum_{u\in \mathcal{U}}\Bigg(\frac{4 \Theta^2 \beta^2 \Delta_{D_u}^{(k)} \eta_k^2 \left(\ell^{(k)}-1\right)}{1- 4\eta_k^2\beta^2 \ell^{(k)}\left(\ell^{(k)}-1\right)} \left(1-\frac{{B}_{u}}{D_{u}} \right)  \frac{{(D_{u}-1)}\left(\sigma_{u}\right)^2}{D_{u}{B}_{u}}\Bigg)\nonumber\\
     &+4 |\mathcal{U}| \sum_{u\in \mathcal{U}}\left(\frac{\Delta_{D_u}^{(k)}}{1- 4\eta_k^2\beta^2 \ell^{(k)}\left(\ell^{(k)}-1\right)}\right)\Bigg\Vert\nabla \mathfrak{L}_u(\bm{\omega}^{(k)})-\nabla \mathfrak{L}(\bm{\omega}^{(k)})\Bigg\Vert^2\nonumber\\
     &+4 |\mathcal{U}| \sum_{u\in \mathcal{U}}\left(\frac{\Delta_{D_u}^{(k)}}{1- 4\eta_k^2\beta^2 \ell^{(k)}\left(\ell^{(k)}-1\right)}\right)\Bigg\Vert\nabla \mathfrak{L}(\bm{\omega}^{(k)})\Bigg\Vert^2\nonumber\\
     &\le 2 |\mathcal{U}| \sum_{u\in \mathcal{U}}\Bigg(\frac{4 \Theta^2 \beta^2 \Delta_{D_u}^{(k)} \eta_k^2 \left(\ell^{(k)}-1\right)}{1- 4\eta_k^2\beta^2 \ell^{(k)}\left(\ell^{(k)}-1\right)} \left(1-\frac{{B}_{u}}{D_{u}} \right)  \frac{{(D_{u}-1)}\left(\sigma_{u}\right)^2}{D_{u}{B}_{u}}\Bigg)\nonumber\\
     &+4 |\mathcal{U}| \sum_{u\in \mathcal{U}}\left(\frac{\Delta_{D_u}^{(k)}}{1- 4\eta_k^2\beta^2 \ell^{(k)}\left(\ell^{(k)}-1\right)}\right)\mathfrak{X}^{(k)}_{u} +4 |\mathcal{U}| \sum_{u\in \mathcal{U}}\left(\frac{\Delta_{D_u}^{(k)}}{1- 4\eta_k^2\beta^2 \ell^{(k)}\left(\ell^{(k)}-1\right)}\right)G^2,
\end{align}
where in inequality $(i)$ we have used the Cauchy-Schwarz inequality $\Vert \mathbf{a}+\mathbf{b} \Vert^2\leq 2 \Vert \mathbf{a} \Vert^2+2\Vert \mathbf{b} \Vert^2$. Replacing the above result back in \eqref{ineq:main6} yields 

 \begin{align}\label{ineq:main7}
    \mathbb{E}_s&\left[\mathfrak{L}(\bm{\omega}^{(k+1)})\right] \leq \mathfrak{L}(\bm{\omega}^{(k)}) - \frac{\eta_{_k}\ell^{(k)}}{4} \mathbb{E}_s\Bigg[\Bigg\Vert\nabla{\mathfrak{L}(\bm{\omega}^{(k)})}\Bigg\Vert^2\Bigg]\nonumber\\
    &~~~~+     \sum_{u\in \mathcal{U}}\frac{D_{u}}{D \ell^{(k)}}\frac{{\ell^{(k)}  2 \beta^2\Theta^2 \eta_k^3}\left(\ell^{(k)}\right)\left(\ell^{(k)}-1\right)}{1- 4\eta_k^2\beta^2 \ell^{(k)}\left(\ell^{(k)}-1\right)}\left(1-\frac{{B}_{u}}{D_{u}} \right)  \frac{{(D_{u}-1)}
    \left(\sigma_{u}\right)^2}{D_{u}{B}_{u}}\nonumber  \nonumber\\
    %%%%%%%%%
    &+    \sum_{u\in \mathcal{U}}\frac{D_{u}}{D}\frac{ {4 \eta_k^3 \ell^{(k)} \beta^2 }\left(\ell^{(k)}\right) \left(\ell^{(k)}-1\right)}{{1- 4\eta_k^2\beta^2 \ell^{(k)}\left(\ell^{(k)}-1\right)}}  \mathfrak{X}^{(k)}_u +    \frac{{4 \eta_k^3\ell^{(k)}\beta^2 }\left(\ell^{(k)}\right) \left(\ell^{(k)}-1\right)}{{1- 4\eta_k^2\beta^2 \ell^{(k)}\left(\ell^{(k)}-1\right)}}  \Bigg\Vert  \nabla \mathfrak{L}(\bm{\omega}^{(k)}) \Bigg\Vert^2\nonumber\\
    &+   |\mathcal{U}| \sum_{u\in \mathcal{U}}\Bigg(\frac{12 \Theta^2 \beta^2 \Delta_{D_u}^{(k)} \eta_k^3 \ell^{(k)}  \left(\ell^{(k)}-1\right)}{1- 4\eta_k^2\beta^2 \ell^{(k)}\left(\ell^{(k)}-1\right)} \left(1-\frac{{B}_{u}}{D_{u}} \right)  \frac{{(D_{u}-1)}\left(\sigma_{u}\right)^2}{D_{u}{B}_{u}}\Bigg)\nonumber\\
     &+ |\mathcal{U}| \sum_{u\in \mathcal{U}}\left(\frac{6\eta_{_k}\ell^{(k)} \Delta_{D_u}^{(k)}}{1- 4\eta_k^2\beta^2 \ell^{(k)}\left(\ell^{(k)}-1\right)}\right)\mathfrak{X}^{(k)}_{u} +  |\mathcal{U}| \sum_{u\in \mathcal{U}}\left(\frac{6\eta_{_k}\ell^{(k)}  \Delta_{D_u}^{(k)}}{1- 4\eta_k^2\beta^2 \ell^{(k)}\left(\ell^{(k)}-1\right)}\right)G^2 \nonumber\\
    &~~~~+ \frac{\beta\eta_{_k}^2\left(\ell^{(k)}\right)^2}{2}  \sum_{u\in \mathcal{U}}\frac{ \widehat{\lambda}_{u}^{(k)} }{  \ell^{(k)}}\left(1-\frac{{B}_{u}}{D_{u}} \right) \frac{2(D_{u}-1)\Theta^2\left(\sigma_{u}\right)^2}{D_{u}{B}_{u}}\nonumber\\
    &=\mathfrak{L}(\bm{\omega}^{(k)}) +  \eta_{_k}\ell^{(k)}  \left(\frac{{4 \eta_k^2 \beta^2 }\left(\ell^{(k)}\right) \left(\ell^{(k)}-1\right)}{{1- 4\eta_k^2\beta^2 \ell^{(k)}\left(\ell^{(k)}-1\right)}} - \frac{1}{4}\right) \Bigg\Vert  \nabla \mathfrak{L}(\bm{\omega}^{(k)}) \Bigg\Vert^2\nonumber\\
    &~~~~+     \sum_{u\in \mathcal{U}}\frac{D_{u}}{D \ell^{(k)}}\frac{{\ell^{(k)}  2 \beta^2\Theta^2 \eta_k^3}\left(\ell^{(k)}\right)\left(\ell^{(k)}-1\right)}{1- 4\eta_k^2\beta^2 \ell^{(k)}\left(\ell^{(k)}-1\right)}\left(1-\frac{{B}_{u}}{D_{u}} \right)  \frac{{(D_{u}-1)}
    \left(\sigma_{u}\right)^2}{D_{u}{B}_{u}}\nonumber  \nonumber\\
    %%%%%%%%%
    &+    \sum_{u\in \mathcal{U}}\left(\frac{\frac{D_{u}}{D} {4 \eta_k^3 \ell^{(k)} \beta^2 }\left(\ell^{(k)}\right) \left(\ell^{(k)}-1\right)+|\mathcal{U}|6\eta_{_k}\ell^{(k)} \Delta_{D_u}^{(k)}}{{1- 4\eta_k^2\beta^2 \ell^{(k)}\left(\ell^{(k)}-1\right)}}\right)\mathfrak{X}_{u}\nonumber\\
    &+   |\mathcal{U}| \sum_{u\in \mathcal{U}}\Bigg(\frac{12 \Theta^2 \beta^2 \Delta_{D_u}^{(k)} \eta_k^3 \ell^{(k)}  \left(\ell^{(k)}-1\right)}{1- 4\eta_k^2\beta^2 \ell^{(k)}\left(\ell^{(k)}-1\right)} \left(1-\frac{{B}_{u}}{D_{u}} \right)  \frac{{(D_{u}-1)}\left(\sigma_{u}\right)^2}{D_{u}{B}_{u}}\Bigg)\nonumber\\
     &+  |\mathcal{U}| \sum_{u\in \mathcal{U}}\left(\frac{6\eta_{_k}\ell^{(k)}  \Delta_{D_u}^{(k)}}{1- 4\eta_k^2\beta^2 \ell^{(k)}\left(\ell^{(k)}-1\right)}\right)G^2  + \frac{\beta\eta_{_k}^2\left(\ell^{(k)}\right)^2}{2}  \sum_{u\in \mathcal{U}}\frac{ \widehat{\lambda}_{u}^{(k)} }{  \ell^{(k)}}\left(1-\frac{{B}_{u}}{D_{u}} \right) \frac{2(D_{u}-1)\Theta^2\left(\sigma_{u}\right)^2}{D_{u}{B}_{u}}.
\end{align}
Assuming
\begin{align}\label{eta_cond_main}
    \frac{{4 \eta_k^2 \beta^2 }\left(\ell^{(k)}\right) \left(\ell^{(k)}-1\right)}{{1- 4\eta_k^2\beta^2 \ell^{(k)}\left(\ell^{(k)}-1\right)}} <\zeta^{(k)}< \frac{1}{4}
\end{align}
yields
\begin{align}\label{eta_cond}
      \eta_k   < \frac{1}{2 \beta}\sqrt{\frac{\zeta^{(k)}}{\left(1+ \zeta^{(k)}\right)  \left(\ell^{(k)}\right) \left(\ell^{(k)}-1\right)}}.
\end{align}
Applying the above condition on \eqref{ineq:main7} and performing some algebraic manipulations give us
\begin{equation}
\footnotesize
\begin{aligned}\label{ineq:main7}
     \mathbb{E}_s \left[\Bigg\Vert  \nabla \mathfrak{L}(\bm{\omega}^{(k)}) \Bigg\Vert^2\right]  &\leq \frac{\mathbb{E}_s \left[\mathfrak{L}(\bm{\omega}^{(k)})\right]- \mathbb{E}_s \left[\mathfrak{L}(\bm{\omega}^{(k+1)})\right]}{\eta_{_k}\ell^{(k)}  \left(\frac{1}{4} - \zeta^{(k)}\right)} +     \frac{1}{\eta_{_k}\ell^{(k)}  \left(\frac{1}{4} - \zeta^{(k)}\right)}\sum_{u\in \mathcal{U}}\frac{D_{u}}{D \ell^{(k)}}\frac{{\ell^{(k)}  2 \beta^2\Theta^2 \eta_k^3}\left(\ell^{(k)}\right)\left(\ell^{(k)}-1\right)}{1- 4\eta_k^2\beta^2 \ell^{(k)}\left(\ell^{(k)}-1\right)}\left(1-\frac{{B}_{u}}{D_{u}} \right)  \frac{{(D_{u}-1)}
    \left(\sigma_{u}\right)^2}{D_{u}{B}_{u}}  \nonumber\\
    %%%%%%%%%
    &+    \frac{1}{\eta_{_k}\ell^{(k)}  \left(\frac{1}{4} - \zeta^{(k)}\right)}\sum_{u\in \mathcal{U}}\left(\frac{\frac{D_{u}}{D} {4 \eta_k^3 \ell^{(k)} \beta^2 }\left(\ell^{(k)}\right) \left(\ell^{(k)}-1\right)+|\mathcal{U}|6\eta_{_k}\ell^{(k)} \Delta_{D_u}^{(k)}}{{1- 4\eta_k^2\beta^2 \ell^{(k)}\left(\ell^{(k)}-1\right)}}\right)\mathfrak{X}^{(k)}_{u} + \frac{\beta\eta_{_k}^2\left(\ell^{(k)}\right)^2}{2}  \sum_{u\in \mathcal{U}}\frac{ \widehat{\lambda}_{u}^{(k)} }{  \ell^{(k)}}\left(1-\frac{{B}_{u}}{D_{u}} \right) \frac{2(D_{u}-1)\Theta^2\left(\sigma_{u}\right)^2}{D_{u}{B}_{u}}\nonumber\\
    &+    \frac{1}{\eta_{_k}\ell^{(k)}  \left(\frac{1}{4} - \zeta^{(k)}\right)}|\mathcal{U}| \sum_{u\in \mathcal{U}}\Bigg(\frac{12 \Theta^2 \beta^2 \Delta_{D_u}^{(k)} \eta_k^3 \ell^{(k)}  \left(\ell^{(k)}-1\right)}{1- 4\eta_k^2\beta^2 \ell^{(k)}\left(\ell^{(k)}-1\right)} \left(1-\frac{{B}_{u}}{D_{u}} \right)  \frac{{(D_{u}-1)}\left(\sigma_{u}\right)^2}{D_{u}{B}_{u}}\Bigg)\nonumber\\
     &+   \frac{1}{\eta_{_k}\ell^{(k)}  \left(\frac{1}{4} - \zeta^{(k)}\right)}|\mathcal{U}| \sum_{u\in \mathcal{U}}\left(\frac{6\eta_{_k}\ell^{(k)}  \Delta_{D_u}^{(k)}}{1- 4\eta_k^2\beta^2 \ell^{(k)}\left(\ell^{(k)}-1\right)}\right)G^2 \nonumber\\
     &= \frac{\mathbb{E}_s \left[\mathfrak{L}(\bm{\omega}^{(k)})\right]- \mathbb{E}_s \left[\mathfrak{L}(\bm{\omega}^{(k+1)})\right]}{\eta_{_k}\ell^{(k)}  \left(\frac{1}{4} - \zeta^{(k)}\right)} +     \frac{1}{ \left(\frac{1}{4} - \zeta^{(k)}\right)}\sum_{u\in \mathcal{U}}\frac{D_{u}}{D \ell^{(k)}}\frac{{  2 \beta^2\Theta^2 \eta_k^2}\left(\ell^{(k)}\right)\left(\ell^{(k)}-1\right)}{1- 4\eta_k^2\beta^2 \ell^{(k)}\left(\ell^{(k)}-1\right)}\left(1-\frac{{B}_{u}}{D_{u}} \right)  \frac{{(D_{u}-1)}
    \left(\sigma_{u}\right)^2}{D_{u}{B}_{u}}  \nonumber\\
    %%%%%%%%%
    &+    \frac{1}{\left(\frac{1}{4} - \zeta^{(k)}\right)}\sum_{u\in \mathcal{U}}\left(\frac{\frac{D_{u}}{D} {4 \eta_k^2  \beta^2 }\left(\ell^{(k)}\right) \left(\ell^{(k)}-1\right)+|\mathcal{U}|6  \Delta_{D_u}^{(k)}}{{1- 4\eta_k^2\beta^2 \ell^{(k)}\left(\ell^{(k)}-1\right)}}\right)\mathfrak{X}^{(k)}_{u} + \frac{\beta\eta_{_k}\ell^{(k)}}{\left(\frac{1}{4} - \zeta^{(k)}\right)}   \sum_{u\in \mathcal{U}}\frac{ \widehat{\lambda}_{u}^{(k)} }{  \ell^{(k)}}\left(1-\frac{{B}_{u}}{D_{u}} \right) \frac{(D_{u}-1)\Theta^2\left(\sigma_{u}\right)^2}{D_{u}{B}_{u}}\nonumber\\
    &+    \frac{|\mathcal{U}|}{ \left(\frac{1}{4} - \zeta^{(k)}\right)} \sum_{u\in \mathcal{U}}\Bigg(\frac{12 \Theta^2 \beta^2 \Delta_{D_u}^{(k)} \eta_k^2  \left(\ell^{(k)}-1\right)}{1- 4\eta_k^2\beta^2 \ell^{(k)}\left(\ell^{(k)}-1\right)} \left(1-\frac{{B}_{u}}{D_{u}} \right)  \frac{{(D_{u}-1)}\left(\sigma_{u}\right)^2}{D_{u}{B}_{u}}\Bigg) +   \frac{|\mathcal{U}|}{ \left(\frac{1}{4} - \zeta^{(k)}\right)} \sum_{u\in \mathcal{U}}\left(\frac{6  \Delta_{D_u}^{(k)}}{1- 4\eta_k^2\beta^2 \ell^{(k)}\left(\ell^{(k)}-1\right)}\right)G^2
\end{aligned}
\end{equation}
Further, from \eqref{eta_cond_main}, we get
\begin{align}
    \frac{1}{{1- 4\eta_k^2\beta^2 \ell^{(k)}\left(\ell^{(k)}-1\right)}}<\left(\zeta^{(k)}+1\right)
\end{align}
Applying this condition to the above inequality and taking total expectation and averaging across global aggregations lead to
\begin{equation}
\footnotesize
\begin{aligned}\label{ineq:main7}
     \frac{1}{K} \sum_{k=0}^{K-1}\mathbb{E}_s \left[\Bigg\Vert  \nabla \mathfrak{L}(\bm{\omega}^{(k)}) \Bigg\Vert^2\right]  &\leq \frac{1}{K} \sum_{k=0}^{K-1}\frac{\mathbb{E}_s \left[\mathfrak{L}(\bm{\omega}^{(k)})\right]- \mathbb{E}_s \left[\mathfrak{L}(\bm{\omega}^{(k+1)})\right]}{\eta_{_k}\ell^{(k)}  \left(\frac{1}{4} - \zeta^{(k)}\right)} \nonumber\\
     &+   \frac{1}{K} \sum_{k=0}^{K-1}\vast[  \frac{\left(\zeta^{(k)}+1\right)}{ \left(\frac{1}{4} - \zeta^{(k)}\right)}\sum_{u\in \mathcal{U}}\frac{D_{u}}{D }{  2 \beta^2\Theta^2 \eta_k^2} \left(\ell^{(k)}-1\right)  \left(1-\frac{{B}_{u}}{D_{u}} \right)  \frac{{(D_{u}-1)}
    \left(\sigma_{u}\right)^2}{D_{u}{B}_{u}}  \nonumber\\
    %%%%%%%%%
    &+    \frac{\left(\zeta^{(k)}+1\right)}{\left(\frac{1}{4} - \zeta^{(k)}\right)}\sum_{u\in \mathcal{U}}  \left( \frac{D_{u}}{D} {4 \eta_k^2  \beta^2 }\left(\ell^{(k)}\right) \left(\ell^{(k)}-1\right)+|\mathcal{U}|6  \Delta_{D_u}^{(k)}\right)\mathfrak{X}^{(k)}_{u} + \frac{\beta\eta_{_k}}{\left(\frac{1}{4} - \zeta^{(k)}\right)}   \sum_{u\in \mathcal{U}} \widehat{\lambda}_{u}^{(k)} \left(1-\frac{{B}_{u}}{D_{u}} \right) \frac{(D_{u}-1)\Theta^2\left(\sigma_{u}\right)^2}{D_{u}{B}_{u}}\nonumber\\
    &+    \frac{|\mathcal{U}| \left(\zeta^{(k)}+1\right)}{ \left(\frac{1}{4} - \zeta^{(k)}\right)} \sum_{u\in \mathcal{U}}\Bigg( 12 \Theta^2 \beta^2 \Delta_{D_u}^{(k)} \eta_k^2  \left(\ell^{(k)}-1\right) \left(1-\frac{{B}_{u}}{D_{u}} \right)  \frac{{(D_{u}-1)}\left(\sigma_{u}\right)^2}{D_{u}{B}_{u}}\Bigg) +   \frac{|\mathcal{U}|\left(\zeta^{(k)}+1\right)}{ \left(\frac{1}{4} - \zeta^{(k)}\right)} \sum_{u\in \mathcal{U}}\left(6  \Delta_{D_u}^{(k)}\right)G^2\vast]
\end{aligned}
\end{equation}
which concludes the proof.
\newpage

\newpage
\section{Proof of Corollary \ref{cor:main}}\label{app:cor:main}

\noindent Using the $\beta$-smoothness of the global loss function (Assumption~\ref{Assup:lossFun}) and considering Lemma~\ref{lemma:smooth}, we have 
\begin{equation}
 \mathfrak{L}(\bm{\omega}^{(k+1)}) \leq \mathfrak{L}(\bm{\omega}^{(k)}) +  \left\langle\nabla{\mathfrak{L}(\bm{\omega}^{(k)})},\left( \bm{\omega}^{(k+1)} - \bm{\omega}^{(k)}\right)\right\rangle+ \frac{\beta}{2} \left\Vert \bm{\omega}^{(k+1)} - \bm{\omega}^{(k)}\right\Vert^2.
\end{equation}
Replacing the updating rule for $\bm{\omega}^{(k+1)}$ and taking the conditional expectation (with respect to randomized data sampling  (referred to by $\mathbb{E}_{s}$) and stochastic node selection (referred to by $\mathbb{E}_{\hat{\lambda}}$)) from both hand sides yields
\begin{align}\label{ineqq:main1}
 \mathbb{E}_{s,\hat{\lambda}}\left[\mathfrak{L}(\bm{\omega}^{(k+1)})\right] &\leq \mathbb{E}_{s,\hat{\lambda}}\left[\mathfrak{L}(\bm{\omega}^{(k)}) -  \left\langle\nabla{\mathfrak{L}(\bm{\omega}^{(k)})},\eta_{_k}\widetilde{\nabla \mathfrak{L}}^{(k)}\right\rangle+ \frac{\beta}{2} \left\Vert \eta_{_k}\widetilde{\nabla \mathfrak{L}}^{(k)}\right\Vert^2\right]\nonumber\\
 %%%%%%%
 &= \mathfrak{L}(\bm{\omega}^{(k)}) - \eta_{_k}\ell^{(k)}\mathbb{E}_{s,\hat{\lambda}}\left[\left\langle\nabla{\mathfrak{L}(\bm{\omega}^{(k)})}, \sum_{u\in \mathcal{U}}\frac{\widehat{\lambda}_{u}^{(k)}}{ \ell^{(k)}}\widetilde{\nabla \mathfrak{L}}_{u}^{(k)}\right\rangle\right] + \frac{\beta\eta_{_k}^2\left(\ell^{(k)}\right)^2}{2} \mathbb{E}_{s,\hat{\lambda}}\left[\left\Vert \sum_{u\in \mathcal{U}}\frac{\widehat{\lambda}_{u}^{(k)}}{ \ell^{(k)}}\widetilde{\nabla \mathfrak{L}}_{u}^{(k)}\right\Vert^2\right].
\end{align}
Here, the notation $\mathbb{E}_{s,\hat{\lambda}}$ denotes the expectation with respect to both random data sampling and random node selection.
Since $\widetilde{\nabla \mathfrak{L}}_{u}^{(k)} = \frac{1}{\eta_{_k}}\left(\bm{\omega}^{(k)}-\bm{\omega}_{u}^{(k),\ell^{(k)}}\right)$, via recursive expansion of the update rule in~\eqref{eq:updaterule}, we get
\begin{equation}\label{eqq:nablabarF}
\widetilde{\nabla \mathfrak{L}}_{u}^{(k)} = \sum_{\ell=1}^{\ell^{(k)}} \sum_{d\in \mathcal{B}^{(k),\ell}_{u}} \hspace{-3mm} {\frac{\nabla  f(\bm{\omega}^{(k),\ell-1}_{u},d)}{{B}_{u}}}.
\end{equation}
Replacing the above result back in \eqref{ineqq:main1} leads to
\begin{equation}
\begin{aligned}
    \mathbb{E}_{s,\hat{\lambda}}\left[\mathfrak{L}(\bm{\omega}^{(k+1)})\right] &\leq  \mathfrak{L}(\bm{\omega}^{(k)}) - \eta_{_k} \ell^{(k)}\mathbb{E}_{s,\hat{\lambda}}\left[\left\langle\nabla{\mathfrak{L}(\bm{\omega}^{(k)})}, \sum_{u\in \mathcal{U}}\frac{\widehat{\lambda}_{u}^{(k)}}{ \ell^{(k)}}\sum_{\ell=1}^{\ell^{(k)}} \sum_{d\in \mathcal{B}^{(k),\ell}_{u}} \hspace{-3mm} {\frac{\nabla  f(\bm{\omega}^{(k),\ell-1}_{u},d)}{{B}_{u}}}\right\rangle\right] + \frac{\beta\eta_{_k}^2\left(\ell^{(k)}\right)^2}{2} \mathbb{E}_{s,\hat{\lambda}}\left[\left\Vert  \sum_{u\in \mathcal{U}}\frac{\widehat{\lambda}_{u}^{(k)}}{ \ell^{(k)}}\widetilde{\nabla \mathfrak{L}}_{u}^{(k)}\right\Vert^2\right].
\end{aligned}
\end{equation}
Let $\mathscr{N}_u^{\mathsf{G},(k),\ell}=\sum_{d\in \mathcal{B}^{(k),\ell}_{u}}{\frac{\nabla  f(\bm{\omega}^{(k),\ell-1}_{u},d)}{{B}_{u}}}-{\nabla  \mathfrak{L}_u(\bm{\omega}^{(k),\ell-1}_{u})}$ denote the noise of SGD of node $u$, where ${\nabla  \mathfrak{L}_u(\bm{\omega}^{(k),\ell-1}_{u})}=\sum_{d\in\mathcal{D}_{u}}\frac{\nabla  f(\bm{\omega}^{(k),\ell-1}_{u},d)}{D_{u}}$. The above inequality can be written as follows:
\begin{align}\label{ineqq:main2}
    \mathbb{E}_{s,\hat{\lambda}}&\left[\mathfrak{L}(\bm{\omega}^{(k+1)})\right] \leq  \mathfrak{L}(\bm{\omega}^{(k)})- \eta_{_k}\ell^{(k)} \mathbb{E}_{s,\hat{\lambda}}\left[\left\langle\nabla{\mathfrak{L}(\bm{\omega}^{(k)})}, \sum_{u\in \mathcal{U}}\frac{\widehat{\lambda}_{u}^{(k)}}{ \ell^{(k)}}\sum_{\ell=1}^{\ell^{(k)}}  \left({\nabla  \mathfrak{L}_u(\bm{\omega}^{(k),\ell-1}_{u})}+\mathscr{N}_u^{\mathsf{G},(k),\ell}\right)\right\rangle\right]\nonumber\\
    &~~~~~~~~+ \frac{\beta\eta_{_k}^2\left(\ell^{(k)}\right)^2}{2} \mathbb{E}_{s,\hat{\lambda}}\left[\left\Vert  \sum_{u\in \mathcal{U}}\frac{\widehat{\lambda}_{u}^{(k)}}{ \ell^{(k)}}\widetilde{\nabla \mathfrak{L}}_{u}^{(k)}\right\Vert^2\right]\nonumber\\
    &=\mathfrak{L}(\bm{\omega}^{(k)}) - \eta_{_k}\ell^{(k)} \mathbb{E}_{s,\hat{\lambda}}\left[\left\langle\nabla{\mathfrak{L}(\bm{\omega}^{(k)})}, \sum_{u\in \mathcal{U}}\frac{\widehat{\lambda}_{u}^{(k)}}{ \ell^{(k)}}\sum_{\ell=1}^{\ell^{(k)}}  {\nabla  \mathfrak{L}_u(\bm{\omega}^{(k),\ell-1}_{u})}\right\rangle\right]\nonumber\\
    &~~~~~~~-\underbrace{\eta_{_k}\ell^{(k)} \mathbb{E}_{s,\hat{\lambda}}\left[\left\langle\nabla{\mathfrak{L}(\bm{\omega}^{(k)})}, \sum_{u\in \mathcal{U}}\frac{\widehat{\lambda}_{u}^{(k)}}{ \ell^{(k)}}\sum_{\ell=1}^{\ell^{(k)}}\mathscr{N}_u^{\mathsf{G},(k),\ell} \right\rangle\right]}_{(x_1)}+ \frac{\beta\eta_{_k}^2\left(\ell^{(k)}\right)^2}{2} \mathbb{E}_{s,\hat{\lambda}}\left[\left\Vert  \sum_{u\in \mathcal{U}}\frac{\widehat{\lambda}_{u}^{(k)}}{ \ell^{(k)}}\widetilde{\nabla \mathfrak{L}}_{u}^{(k)}\right\Vert^2\right]\nonumber\\
    &\overset{(i)}{=}\mathfrak{L}(\bm{\omega}^{(k)}) - \eta_{_k}\ell^{(k)} \mathbb{E}_{s,\hat{\lambda}}\left[\left\langle\nabla{\mathfrak{L}(\bm{\omega}^{(k)})}, \sum_{u\in \mathcal{U}}\frac{\widehat{\lambda}_{u}^{(k)}}{ \ell^{(k)}}\sum_{\ell=1}^{\ell^{(k)}}  {\nabla  \mathfrak{L}_u(\bm{\omega}^{(k),\ell-1}_{u})}\right\rangle\right] + \frac{\beta \eta_{_k}^2\left(\ell^{(k)}\right)^2}{2} \mathbb{E}_{s,\hat{\lambda}}\left[\left\Vert  \sum_{u\in \mathcal{U}}\frac{\widehat{\lambda}_{u}^{(k)}}{ \ell^{(k)}}\widetilde{\nabla \mathfrak{L}}_{u}^{(k)}\right\Vert^2\right],
\end{align}
% {\color{blue} \begin{equation}
%    \frac{D_u}{D}- E[\widehat{\lambda}_{u}^{(k)}]\leq \eta_k\Rightarrow E[-\mathscr{U}^{(k)}]= \eta_k \sum_{u\in \mathcal{U}}\frac{D_{u}}{D \ell^{(k)}}\sum_{\ell=1}^{\ell^{(k)}} {\nabla  \mathfrak{L}_u(\bm{\omega}^{(k),\ell-1}_{u})} 
% \end{equation}
% }
where $(i)$ uses the fact that $(x_1)$ is zero since the noise of gradient estimation across the mini-batches are independent and zero mean. Further, let 
\begin{align}\label{eqq:user_selection_noise}
    \mathscr{U}^{(k)}&= \sum_{u\in \mathcal{U}}\frac{\widehat{\lambda}_{u}^{(k)}}{ \ell^{(k)}}\sum_{\ell=1}^{\ell^{(k)}} {\nabla  \mathfrak{L}_u(\bm{\omega}^{(k),\ell-1}_{u})}- \sum_{u\in \mathcal{U}}\frac{D_{u}}{D \ell^{(k)}}\sum_{\ell=1}^{\ell^{(k)}} {\nabla  \mathfrak{L}_u(\bm{\omega}^{(k),\ell-1}_{u})} \nonumber\\
    &= \sum_{u\in \mathcal{U}}\left(\widehat{\lambda}_{u}^{(k)}- \frac{D_{u}}{D}\right)\frac{1}{\ell^{(k)}}\sum_{\ell=1}^{\ell^{(k)}} {\nabla  \mathfrak{L}_u(\bm{\omega}^{(k),\ell-1}_{u})}
\end{align}
denote the noise of node selection of the random walker. Inequality \eqref{ineqq:main2} can be rewritten as follows:
\begin{align}\label{ineqq:main2_user}
    \hspace{-8mm}\mathbb{E}_{s,\hat{\lambda}}\left[\mathfrak{L}(\bm{\omega}^{(k+1)})\right] &\leq \mathfrak{L}(\bm{\omega}^{(k)}) - \eta_{_k} \ell^{(k)}\mathbb{E}_{s,\hat{\lambda}}\left[\left\langle\nabla{\mathfrak{L}(\bm{\omega}^{(k)})}, \sum_{u\in \mathcal{U}}\frac{D_{u}}{D \ell^{(k)}}\sum_{\ell=1}^{\ell^{(k)}}  {\nabla  \mathfrak{L}_u(\bm{\omega}^{(k),\ell-1}_{u})}+ \mathscr{U}^{(k)} \right\rangle\right] \nonumber\\
    &+ \frac{\beta\eta_{_k}^2\left(\ell^{(k)}\right)^2}{2} \mathbb{E}_{s,\hat{\lambda}}\left[\left\Vert  \sum_{u\in \mathcal{U}}\frac{\widehat{\lambda}_{u}^{(k)}}{ \ell^{(k)}}\widetilde{\nabla \mathfrak{L}}_{u}^{(k)}\right\Vert^2\right]
\end{align}
If random walk is ergodic and walker starts at stationary distribution, taking expected value of \eqref{ineqq:main2_user} with respect to the node selection gives us
\begin{align}\label{ineqq:main_user_upper}
    \mathbb{E}_{s,\hat{\lambda}}\left[\mathfrak{L}(\bm{\omega}^{(k+1)})\right] &\leq \mathfrak{L}(\bm{\omega}^{(k)}) - \eta_{_k} \ell^{(k)}\mathbb{E}_{s}\left[\left\langle\nabla{\mathfrak{L}(\bm{\omega}^{(k)})}, \sum_{u\in \mathcal{U}}\frac{D_{u}}{D \ell^{(k)}}\sum_{\ell=1}^{\ell^{(k)}}  {\nabla  \mathfrak{L}_u(\bm{\omega}^{(k),\ell-1}_{u})}+ \mathbb{E}_{\hat{\lambda}}\left[\mathscr{U}^{(k)}\right] \right\rangle\right]\nonumber\\
    &~~~~~~~~~~~+ \frac{\beta\eta_{_k}^2\left(\ell^{(k)}\right)^2}{2} \mathbb{E}_{s,\hat{\lambda}}\left[\left\Vert  \sum_{u\in \mathcal{U}}\frac{\widehat{\lambda}_{u}^{(k)}}{ \ell^{(k)}}\widetilde{\nabla \mathfrak{L}}_{u}^{(k)}\right\Vert^2\right]
\end{align}
where $\mathbb{E}_{\hat{\lambda}}\left[\mathscr{U}^{(k)}\right]$ is the expected value of the noise of the node selection, calculated as follows:
\begin{align}\label{eqq:user_selection_noise}
    \mathbb{E}_{\hat{\lambda}}\left[\mathscr{U}^{(k)}\right]&=  \sum_{u\in \mathcal{U}}\left(\mathbb{E}_{\hat{\lambda}}\left[\widehat{\lambda}_{u}^{(k)}\right]- \frac{D_{u}}{D}\right)\frac{1}{\ell^{(k)}}\sum_{\ell=1}^{\ell^{(k)}} {\nabla  \mathfrak{L}_u(\bm{\omega}^{(k),\ell-1}_{u})}
\end{align}
Similarly, writing the last term of the above inequality in terms of $\mathscr{N}_u^{\mathsf{G},(k),\ell}$ gives us
\begin{align}
    %%%%%%%%%
    \mathbb{E}_{s,\hat{\lambda}}&\left[\mathfrak{L}(\bm{\omega}^{(k+1)})\right] \leq \mathfrak{L}(\bm{\omega}^{(k)}) - \eta_{_k} \ell^{(k)}\mathbb{E}_s\left[\left\langle\nabla{\mathfrak{L}(\bm{\omega}^{(k)})}, \sum_{u\in \mathcal{U}}\frac{D_{u}}{D \ell^{(k)}}\sum_{\ell=1}^{\ell^{(k)}}  {\nabla  \mathfrak{L}_u(\bm{\omega}^{(k),\ell-1}_{u})}\right\rangle\right]\nonumber\\
    &+ \eta_{_k}\ell^{(k)} \mathbb{E}_s\left[\left\langle\nabla{\mathfrak{L}(\bm{\omega}^{(k)})},-\mathbb{E}_{\hat{\lambda}}\left[\mathscr{U}^{(k)}\right] \right\rangle\right]+ \frac{\beta\eta_{_k}^2\left(\ell^{(k)}\right)^2}{2} \mathbb{E}_{s,\hat{\lambda}}\Bigg[\Bigg\Vert  \sum_{u\in \mathcal{U}}\frac{\widehat{\lambda}_{u}^{(k)}}{ \ell^{(k)}}\sum_{\ell=1}^{\ell^{(k)}} \sum_{d\in \mathcal{B}^{(k),\ell}_{u}} \hspace{-3mm} {\frac{\nabla  f(\bm{\omega}^{(k),\ell-1}_{u},d)}{{B}_{u}}} \Bigg\Vert^2\Bigg]\nonumber\\
    &=\mathfrak{L}(\bm{\omega}^{(k)}) - \eta_{_k} \ell^{(k)}\mathbb{E}_s\left[\left\langle\nabla{\mathfrak{L}(\bm{\omega}^{(k)})}, \sum_{u\in \mathcal{U}}\frac{D_{u}}{D \ell^{(k)}}\sum_{\ell=1}^{\ell^{(k)}}  {\nabla  \mathfrak{L}_u(\bm{\omega}^{(k),\ell-1}_{u})}\right\rangle\right]\nonumber\\
    &+ \eta_{_k}\ell^{(k)} \mathbb{E}_s\left[\left\langle\nabla{\mathfrak{L}(\bm{\omega}^{(k)})},-\mathbb{E}_{\hat{\lambda}}\left[\mathscr{U}^{(k)}\right] \right\rangle\right]+ \frac{\beta\eta_{_k}^2\left(\ell^{(k)}\right)^2}{2} \mathbb{E}_{s,\hat{\lambda}}\Bigg[\Bigg\Vert  \sum_{u\in \mathcal{U}}\frac{\widehat{\lambda}_{u}^{(k)}}{ \ell^{(k)}}\sum_{\ell=1}^{\ell^{(k)}} \left({\nabla  \mathfrak{L}_u(\bm{\omega}^{(k),\ell-1}_{u})}+\mathscr{N}_u^{\mathsf{G},(k),\ell}\right) \Bigg\Vert^2\Bigg]\nonumber\\
    %%%%%%%
    &=\mathfrak{L}(\bm{\omega}^{(k)}) - \eta_{_k}\ell^{(k)} \mathbb{E}_s\left[\left\langle\nabla{\mathfrak{L}(\bm{\omega}^{(k)})}, \sum_{u\in \mathcal{U}}\frac{D_{u}}{D \ell^{(k)}}\sum_{\ell=1}^{\ell^{(k)}}  {\nabla  \mathfrak{L}_u(\bm{\omega}^{(k),\ell-1}_{u})}\right\rangle\right] + \eta_{_k}\ell^{(k)} \mathbb{E}_s\left[\left\langle\nabla{\mathfrak{L}(\bm{\omega}^{(k)})},-\mathbb{E}_{\hat{\lambda}}\left[\mathscr{U}^{(k)}\right] \right\rangle\right]\nonumber\\
    &~~+ \frac{\beta\eta_{_k}^2\left(\ell^{(k)}\right)^2}{2} \mathbb{E}_{s,\hat{\lambda}}\left[\left\Vert  \sum_{u\in \mathcal{U}}\frac{\widehat{\lambda}_{u}^{(k)}}{ \ell^{(k)}}\sum_{\ell=1}^{\ell^{(k)}} {\nabla  \mathfrak{L}_u(\bm{\omega}^{(k),\ell-1}_{u})}+\left( \sum_{u\in \mathcal{U}}\frac{\widehat{\lambda}_{u}^{(k)}}{ \ell^{(k)}}\sum_{\ell=1}^{\ell^{(k)}}\mathscr{N}_u^{\mathsf{G},(k),\ell}\right) \right\Vert^2\right]\nonumber\\
    %%%%%%%
    &\overset{(i)}{=}\mathfrak{L}(\bm{\omega}^{(k)}) - \eta_{_k}\ell^{(k)} \mathbb{E}_s\left[\left\langle\nabla{\mathfrak{L}(\bm{\omega}^{(k)})}, \sum_{u\in \mathcal{U}}\frac{D_{u}}{D \ell^{(k)}}\sum_{\ell=1}^{\ell^{(k)}}  {\nabla  \mathfrak{L}_u(\bm{\omega}^{(k),\ell-1}_{u})}\right\rangle\right] + \eta_{_k}\ell^{(k)} \mathbb{E}_s\left[\left\langle\nabla{\mathfrak{L}(\bm{\omega}^{(k)})},-\mathbb{E}_{\hat{\lambda}}\left[\mathscr{U}^{(k)}\right] \right\rangle\right]\nonumber\\
    &~~+ \frac{\beta\eta_{_k}^2\left(\ell^{(k)}\right)^2}{2} \mathbb{E}_{s,\hat{\lambda}}\Bigg[\Bigg\Vert  \sum_{u\in \mathcal{U}}\frac{\widehat{\lambda}_{u}^{(k)}}{ \ell^{(k)}}\sum_{\ell=1}^{\ell^{(k)}} {\nabla  \mathfrak{L}_u(\bm{\omega}^{(k),\ell-1}_{u})}\Bigg\Vert^2\Bigg]+\frac{\beta\eta_{_k}^2\left(\ell^{(k)}\right)^2}{2}\mathbb{E}_{s,\hat{\lambda}}\Bigg[\Bigg\Vert \sum_{u\in \mathcal{U}}\frac{\widehat{\lambda}_{u}^{(k)}}{ \ell^{(k)}}\sum_{\ell=1}^{\ell^{(k)}}\mathscr{N}_u^{\mathsf{G},(k),\ell}\Bigg\Vert^2\Bigg]\nonumber\\
    &~~+\underbrace{\beta\eta_{_k}^2\left(\ell^{(k)}\right)^2 \mathbb{E}_{s,\hat{\lambda}}\Bigg[\left\langle  \sum_{u\in \mathcal{U}}\frac{\widehat{\lambda}_{u}^{(k)}}{ \ell^{(k)}}\sum_{\ell=1}^{\ell^{(k)}} {\nabla  \mathfrak{L}_u(\bm{\omega}^{(k),\ell-1}_{u})} , \sum_{u\in \mathcal{U}}\frac{\widehat{\lambda}_{u}^{(k)}}{ \ell^{(k)}}\sum_{\ell=1}^{\ell^{(k)}}\mathscr{N}_u^{\mathsf{G},(k),\ell}\right\rangle\Bigg]}_{(x_2)}\nonumber\\
    %%%%%%%
    &\overset{(ii)}{=}\mathfrak{L}(\bm{\omega}^{(k)}) - \eta_{_k}\ell^{(k)} \mathbb{E}_s\left[\left\langle\nabla{\mathfrak{L}(\bm{\omega}^{(k)})}, \sum_{u\in \mathcal{U}}\frac{D_{u}}{D \ell^{(k)}}\sum_{\ell=1}^{\ell^{(k)}}  {\nabla  \mathfrak{L}_u(\bm{\omega}^{(k),\ell-1}_{u})}\right\rangle\right] + \eta_{_k}\ell^{(k)} \mathbb{E}_s\left[\left\langle\nabla{\mathfrak{L}(\bm{\omega}^{(k)})},-\mathbb{E}_{\hat{\lambda}}\left[\mathscr{U}^{(k)}\right] \right\rangle\right]\nonumber\\
    &~+ \frac{\beta\eta_{_k}^2\left(\ell^{(k)}\right)^2}{2} \mathbb{E}_{s,\hat{\lambda}}\Bigg[\Bigg\Vert  \sum_{u\in \mathcal{U}}\frac{\widehat{\lambda}_{u}^{(k)}}{ \ell^{(k)}}\sum_{\ell=1}^{\ell^{(k)}} {\nabla  \mathfrak{L}_u(\bm{\omega}^{(k),\ell-1}_{u})}\Bigg\Vert^2\Bigg]+\frac{\beta\eta_{_k}^2\left(\ell^{(k)}\right)^2}{2}\mathbb{E}_{s,\hat{\lambda}}\Bigg[\Bigg\Vert \sum_{u\in \mathcal{U}}\frac{\widehat{\lambda}_{u}^{(k)}}{ \ell^{(k)}}\sum_{\ell=1}^{\ell^{(k)}}\mathscr{N}_u^{\mathsf{G},(k),\ell}\Bigg\Vert^2\Bigg],
\end{align}
where $(i)$ uses the fact that for any two real valued vectors $\bm{a}$ and $\bm{b}$ with the same length, we have $\Vert \bm{a}+\bm{b}\Vert^2=\Vert \bm{a}\Vert^2+\Vert \bm{b}\Vert^2+2\langle\bm{a},\bm{b}\rangle$. Further, $(ii)$ in the above inequality is due to the fact that $(x_2)$ is zero since the noise of gradient estimation of users are independent and zero mean. Using $\Vert \sum_{i=1}^{n}\bm{x}_i\Vert^2=\sum_{i=1}^{n}\Vert\bm{x}_i\Vert^2+\sum_{\substack{j=1\\j\neq i}}^{n}\left\langle\bm{x}_i,\bm{x}_j\right\rangle$, where $\bm{x}\in \mathbb{R}^m$, and performing some algebraic manipulation, we expand the last term of the above inequality as follows: 
\begin{align}\label{ineqq:main3}
    %%%%%%%%%
    \mathbb{E}_{s,\hat{\lambda}}&\left[\mathfrak{L}(\bm{\omega}^{(k+1)})\right] \leq \mathfrak{L}(\bm{\omega}^{(k)}) - \eta_{_k}\ell^{(k)} \mathbb{E}_s\left[\left\langle\nabla{\mathfrak{L}(\bm{\omega}^{(k)})}, \sum_{u\in \mathcal{U}}\frac{D_{u}}{D \ell^{(k)}}\sum_{\ell=1}^{\ell^{(k)}}  {\nabla  \mathfrak{L}_u(\bm{\omega}^{(k),\ell-1}_{u})}\right\rangle\right]\nonumber\\
    &~~~~+ \eta_{_k}\ell^{(k)} \mathbb{E}_s\left[\left\langle\nabla{\mathfrak{L}(\bm{\omega}^{(k)})},-\mathbb{E}_{\hat{\lambda}}\left[\mathscr{U}^{(k)}\right] \right\rangle\right]+ \frac{\beta\eta_{_k}^2\left(\ell^{(k)}\right)^2}{2}  \sum_{u\in \mathcal{U}}\mathbb{E}_{s,\hat{\lambda}}\Bigg[\left(\frac{\widehat{\lambda}_{u}^{(k)}}{ \ell^{(k)}}\right)^2\Bigg\Vert \sum_{\ell=1}^{\ell^{(k)}}\mathscr{N}_u^{\mathsf{G},(k),\ell}\Bigg\Vert^2\Bigg]\nonumber\\
    &+\underbrace{ \sum_{u\in \mathcal{U}} \sum_{u'\in \mathcal{U}\setminus u}\frac{\beta\eta_{_k}^2\left(\ell^{(k)}\right)^2}{2}\mathbb{E}_{s,\hat{\lambda}}\Bigg[\left\langle \frac{\widehat{\lambda}_{u}^{(k)}}{ \ell^{(k)}}\sum_{\ell=1}^{\ell^{(k)}}\mathscr{N}_u^{\mathsf{G},(k),\ell}, \frac{\widehat{\lambda}_{u'}^{(k)}}{ \ell^{(k)}}\sum_{\ell=1}^{\ell^{(k)}}\mathscr{N}_{u'}^{\mathsf{G},(k),\ell}\right\rangle\Bigg]}_{(x_3)}\nonumber\\
    &~~~~+ \frac{\beta\eta_{_k}^2\left(\ell^{(k)}\right)^2}{2} \mathbb{E}_{s,\hat{\lambda}}\Bigg[\Bigg\Vert  \sum_{u\in \mathcal{U}}\frac{\widehat{\lambda}_{u}^{(k)}}{ \ell^{(k)}}\sum_{\ell=1}^{\ell^{(k)}} {\nabla  \mathfrak{L}_u(\bm{\omega}^{(k),\ell-1}_{u})}\Bigg\Vert^2\Bigg]\nonumber\\
    %%%%%%%
    &\overset{(i)}{=}\mathfrak{L}(\bm{\omega}^{(k)}) - \eta_{_k}\ell^{(k)} \mathbb{E}_s\left[\left\langle\nabla{\mathfrak{L}(\bm{\omega}^{(k)})}, \sum_{u\in \mathcal{U}}\frac{D_{u}}{D \ell^{(k)}}\sum_{\ell=1}^{\ell^{(k)}}  {\nabla  \mathfrak{L}_u(\bm{\omega}^{(k),\ell-1}_{u})}\right\rangle\right]\nonumber\\
    &~~~~+ \eta_{_k}\ell^{(k)} \underbrace{\mathbb{E}_s\left[\left\langle\nabla{\mathfrak{L}(\bm{\omega}^{(k)})},-\mathbb{E}_{\hat{\lambda}}\left[\mathscr{U}^{(k)}\right] \right\rangle\right]}_{(a)}+ \frac{\beta\eta_{_k}^2\left(\ell^{(k)}\right)^2}{2}  \sum_{u\in \mathcal{U}}\mathbb{E}_{s,\hat{\lambda}}\Bigg[\left(\frac{\widehat{\lambda}_{u}^{(k)}}{ \ell^{(k)}}\right)^2\Bigg\Vert \sum_{\ell=1}^{\ell^{(k)}}\mathscr{N}_u^{\mathsf{G},(k),\ell}\Bigg\Vert^2\Bigg]\nonumber\\
    &~~~~+ \frac{\beta\eta_{_k}^2\left(\ell^{(k)}\right)^2}{2} \mathbb{E}_{s,\hat{\lambda}}\Bigg[\Bigg\Vert \underbrace{ \sum_{u\in \mathcal{U}}\frac{\widehat{\lambda}_{u}^{(k)}}{ \ell^{(k)}}\sum_{\ell=1}^{\ell^{(k)}} {\nabla  \mathfrak{L}_u(\bm{\omega}^{(k),\ell-1}_{u})}}_{(b)}\Bigg\Vert^2\Bigg],
\end{align}
where $(i)$ uses the fact that, term $(x_3)$ is zero, since the noise of gradient estimation of a specific node is independent and zero mean. Using Cauchy-Schwartz and Young's inequalities, for two real valued vectors $\bm{a}$ and $\bm{b}$, we have
\begin{equation}
    \langle\bm{a},\bm{b}\rangle\le \frac{\alpha}{2}\Vert \bm{a}\Vert^2+\frac{1}{2\alpha}\Vert \bm{b}\Vert^2,~ \alpha \in \mathbb{R}^{++},
\end{equation}
Setting $\alpha =2$ in the above inequality implies $\langle\bm{a},\bm{b}\rangle\le \Vert \bm{a}\Vert^2+\frac{1}{4}\Vert \bm{b}\Vert^2$. Using this inequality to expand $(a)$ in \eqref{ineqq:main3} and also rewriting $(b)$ in \eqref{ineqq:main3} in terms of $\mathscr{U}^{(k)}$, we can further bound \eqref{ineqq:main3} as follows:

\begin{align}\label{ineqq:main3_1}
    \mathbb{E}_{s,\hat{\lambda}}&\left[\mathfrak{L}(\bm{\omega}^{(k+1)})\right] \leq \mathfrak{L}(\bm{\omega}^{(k)}) - \eta_{_k}\ell^{(k)} \mathbb{E}_s\left[\Bigg\langle\nabla{\mathfrak{L}(\bm{\omega}^{(k)})}, \sum_{u\in \mathcal{U}}\frac{D_{u}}{D \ell^{(k)}}\sum_{\ell=1}^{\ell^{(k)}}  {\nabla  \mathfrak{L}_u(\bm{\omega}^{(k),\ell-1}_{u})}\Bigg\rangle\right]\nonumber\\
    &+ \frac{\beta\eta_{_k}^2\left(\ell^{(k)}\right)^2}{2}  \sum_{u\in \mathcal{U}}\mathbb{E}_{\hat{\lambda}}\left[\left(\frac{\widehat{\lambda}_{u}^{(k)}}{ \ell^{(k)}}\right)^2\right]\mathbb{E}_s\Bigg[\Bigg\Vert \sum_{\ell=1}^{\ell^{(k)}}\mathscr{N}_u^{\mathsf{G},(k),\ell}\Bigg\Vert^2\Bigg]\nonumber\\
    &+ \frac{\beta\eta_{_k}^2\left(\ell^{(k)}\right)^2}{2} \mathbb{E}_{s,\hat{\lambda}}\Bigg[\Bigg\Vert  \sum_{u\in \mathcal{U}}\frac{D_{u}}{D \ell^{(k)}}\sum_{\ell=1}^{\ell^{(k)}} {\nabla  \mathfrak{L}_u(\bm{\omega}^{(k),\ell-1}_{u})}+\mathscr{U}^{(k)}\Bigg\Vert^2\Bigg]+ \eta_{_k}\ell^{(k)} \mathbb{E}_s\left[\left\langle\nabla{\mathfrak{L}(\bm{\omega}^{(k)})},-\mathbb{E}_{\hat{\lambda}}\left[\mathscr{U}^{(k)}\right] \right\rangle\right]\nonumber\\
    %%%%%%%
    &\le \mathfrak{L}(\bm{\omega}^{(k)}) - \eta_{_k}\ell^{(k)} \mathbb{E}_s\left[\Bigg\langle\nabla{\mathfrak{L}(\bm{\omega}^{(k)})}, \sum_{u\in \mathcal{U}}\frac{D_{u}}{D \ell^{(k)}}\sum_{\ell=1}^{\ell^{(k)}}  {\nabla  \mathfrak{L}_u(\bm{\omega}^{(k),\ell-1}_{u})}\Bigg\rangle\right]\nonumber\\
    &+ \frac{\beta\eta_{_k}^2\left(\ell^{(k)}\right)^2}{2}  \sum_{u\in \mathcal{U}}\mathbb{E}_{\hat{\lambda}}\left[\left(\frac{\widehat{\lambda}_{u}^{(k)}}{ \ell^{(k)}}\right)^2\right]\mathbb{E}_s\Bigg[\Bigg\Vert \sum_{\ell=1}^{\ell^{(k)}}\mathscr{N}_u^{\mathsf{G},(k),\ell}\Bigg\Vert^2\Bigg]\nonumber\\
    &+ \frac{\beta\eta_{_k}^2\left(\ell^{(k)}\right)^2}{2} \mathbb{E}_s\Bigg[\underbrace{\Bigg\Vert  \sum_{u\in \mathcal{U}}\frac{D_{u}}{D \ell^{(k)}}\sum_{\ell=1}^{\ell^{(k)}} {\nabla  \mathfrak{L}_u(\bm{\omega}^{(k),\ell-1}_{u})}+\mathscr{U}^{(k)}\Bigg\Vert^2}_{(a)}\Bigg] + \frac{\eta_{_k}\ell^{(k)}}{4} \mathbb{E}_s\left[\left\Vert\nabla{\mathfrak{L}(\bm{\omega}^{(k)})} \right\Vert^2\right]+ \eta_{_k}\ell^{(k)} \mathbb{E}_s\left[\left\Vert\mathbb{E}_{\hat{\lambda}}\left[\mathscr{U}^{(k)}\right] \right\Vert^2\right]\nonumber\\
    %%%%%%%
    &\overset{(i)}{\le}\mathfrak{L}(\bm{\omega}^{(k)}) - \eta_{_k} \ell^{(k)}\mathbb{E}_s\left[ \left\langle\nabla{\mathfrak{L}(\bm{\omega}^{(k)})}, \sum_{u\in \mathcal{U}}\frac{D_{u}}{D \ell^{(k)}}\sum_{\ell=1}^{\ell^{(k)}}  {\nabla  \mathfrak{L}_u(\bm{\omega}^{(k),\ell-1}_{u})}\right\rangle \right]\nonumber\\
    &+ \frac{\beta\eta_{_k}^2\left(\ell^{(k)}\right)^2}{2}  \sum_{u\in \mathcal{U}}\mathbb{E}_{\hat{\lambda}}\left[\left(\frac{\widehat{\lambda}_{u}^{(k)}}{ \ell^{(k)}}\right)^2\right]\mathbb{E}_s\Bigg[\Bigg\Vert \sum_{\ell=1}^{\ell^{(k)}}\mathscr{N}_u^{\mathsf{G},(k),\ell}\Bigg\Vert^2\Bigg]+ \beta\eta_{_k}^2 \left(\ell^{(k)}\right)^2\mathbb{E}_s\Bigg[\Bigg\Vert  \sum_{u\in \mathcal{U}}\frac{D_{u}}{D \ell^{(k)}}\sum_{\ell=1}^{\ell^{(k)}} {\nabla  \mathfrak{L}_u(\bm{\omega}^{(k),\ell-1}_{u})}\Bigg\Vert^2\Bigg]\nonumber\\
    &+ \beta\eta_{_k}^2 \left(\ell^{(k)}\right)^2\mathbb{E}_{s,\hat{\lambda}}\left[\left\Vert\mathscr{U}^{(k)}\right\Vert^2\right]+ \frac{\eta_{_k}\ell^{(k)}}{4} \mathbb{E}_s\left[\left\Vert\nabla{\mathfrak{L}(\bm{\omega}^{(k)})} \right\Vert^2\right]+ \eta_{_k} \ell^{(k)}\mathbb{E}_s\left[\left\Vert\mathbb{E}_{\hat{\lambda}}\left[\mathscr{U}^{(k)}\right] \right\Vert^2\right]\nonumber\\
    &\overset{(ii)}{\le}\mathfrak{L}(\bm{\omega}^{(k)}) - \eta_{_k} \ell^{(k)}\mathbb{E}_s\left[ \left\langle\nabla{\mathfrak{L}(\bm{\omega}^{(k)})}, \sum_{u\in \mathcal{U}}\frac{D_{u}}{D \ell^{(k)}}\sum_{\ell=1}^{\ell^{(k)}}  {\nabla  \mathfrak{L}_u(\bm{\omega}^{(k),\ell-1}_{u})}\right\rangle \right]\nonumber\\
    &+ \frac{\beta\eta_{_k}^2\left(\ell^{(k)}\right)^2}{2}  \sum_{u\in \mathcal{U}}\mathbb{E}_{\hat{\lambda}}\left[\left(\frac{\widehat{\lambda}_{u}^{(k)}}{ \ell^{(k)}}\right)^2\right]\mathbb{E}_s\Bigg[\Bigg\Vert \sum_{\ell=1}^{\ell^{(k)}}\mathscr{N}_u^{\mathsf{G},(k),\ell}\Bigg\Vert^2\Bigg]+ \beta\eta_{_k}^2 \left(\ell^{(k)}\right)^2\mathbb{E}_s\Bigg[\Bigg\Vert  \sum_{u\in \mathcal{U}}\frac{D_{u}}{D \ell^{(k)}}\sum_{\ell=1}^{\ell^{(k)}} {\nabla  \mathfrak{L}_u(\bm{\omega}^{(k),\ell-1}_{u})}\Bigg\Vert^2\Bigg]\nonumber\\
    &+ \beta\eta_{_k}^2 \left(\ell^{(k)}\right)^2\mathbb{E}_{s,\hat{\lambda}}\left[\left\Vert\mathscr{U}^{(k)}\right\Vert^2\right]+ \frac{\eta_{_k}\ell^{(k)}}{4} \mathbb{E}_s\left[\left\Vert\nabla{\mathfrak{L}(\bm{\omega}^{(k)})} \right\Vert^2\right]\nonumber\\
    &+ \eta_{_k} \ell^{(k)} \sum_{u\in \mathcal{U}}\left(\mathbb{E}_{\hat{\lambda}}\left[\widehat{\lambda}_{u}^{(k)}\right]- \frac{D_{u}}{D}\right)^2\sum_{u\in \mathcal{U}}\mathbb{E}_s\left[\left\Vert\frac{1}{\ell^{(k)}}\sum_{\ell=1}^{\ell^{(k)}} {\nabla  \mathfrak{L}_u(\bm{\omega}^{(k),\ell-1}_{u})} \right\Vert^2\right],
    %%%%%%%
\end{align}
where in inequality $(i)$ we have used the Cauchy-Schwarz inequality $\Vert \mathbf{a}+\mathbf{b} \Vert^2\leq 2 \Vert \mathbf{a} \Vert^2+2\Vert \mathbf{b} \Vert^2$ to bound $(a)$ in \eqref{ineqq:main3_1}. Moreover, in $(ii)$, we are using \eqref{eqq:user_selection_noise} and Cauchy-Schwarz inequality to bound $\mathbb{E}_s\left[\left\Vert\mathbb{E}_{\hat{\lambda}}\left[\mathscr{U}^{(k)}\right] \right\Vert^2\right]$. Assuming $\mathbb{E}_{\hat{\lambda}}\left[\widehat{\lambda}_{u}^{(k)}\right]- \frac{D_{u}}{D} \le \eta_k$, we can further bound the above inequality as follows:
\begin{align}\label{ineqq:main3_12}
    \mathbb{E}_{s,\hat{\lambda}}&\left[\mathfrak{L}(\bm{\omega}^{(k+1)})\right] \leq  \mathfrak{L}(\bm{\omega}^{(k)}) - \eta_{_k} \ell^{(k)}\mathbb{E}_s\left[\underbrace{\left\langle\nabla{\mathfrak{L}(\bm{\omega}^{(k)})}, \sum_{u\in \mathcal{U}}\frac{D_{u}}{D \ell^{(k)}}\sum_{\ell=1}^{\ell^{(k)}}  {\nabla  \mathfrak{L}_u(\bm{\omega}^{(k),\ell-1}_{u})}\right\rangle}_{(b)}\right]\nonumber\\
    &+ \frac{\beta\eta_{_k}^2\left(\ell^{(k)}\right)^2}{2}  \sum_{u\in \mathcal{U}}\frac{\mathbb{E}_{\hat{\lambda}}\left[\widehat{\lambda}_{u}^{(k)}\right]}{ \left(\ell^{(k)}\right)^2}\mathbb{E}_s\Bigg[\Bigg\Vert \sum_{\ell=1}^{\ell^{(k)}}\mathscr{N}_u^{\mathsf{G},(k),\ell}\Bigg\Vert^2\Bigg]+ \beta\eta_{_k}^2 \left(\ell^{(k)}\right)^2\mathbb{E}_s\Bigg[\Bigg\Vert  \sum_{u\in \mathcal{U}}\frac{D_{u}}{D \ell^{(k)}}\sum_{\ell=1}^{\ell^{(k)}} {\nabla  \mathfrak{L}_u(\bm{\omega}^{(k),\ell-1}_{u})}\Bigg\Vert^2\Bigg]\nonumber\\
    &+ \beta\eta_{_k}^2 \left(\ell^{(k)}\right)^2\mathbb{E}_{s,\hat{\lambda}}\left[\left\Vert\mathscr{U}^{(k)}\right\Vert^2\right]+ \frac{\eta_{_k}\ell^{(k)}}{4} \mathbb{E}_s\left[\left\Vert\nabla{\mathfrak{L}(\bm{\omega}^{(k)})} \right\Vert^2\right]+ \ell^{(k)} |\mathcal{U}|\eta_k^3\sum_{u\in \mathcal{U}}\mathbb{E}_s\left[\left\Vert\frac{1}{\ell^{(k)}}\sum_{\ell=1}^{\ell^{(k)}} {\nabla  \mathfrak{L}_u(\bm{\omega}^{(k),\ell-1}_{u})} \right\Vert^2\right],
    %%%%%%%
\end{align}
 Using $2\langle\bm{a},\bm{b}\rangle=\Vert \bm{a}\Vert^2+\Vert \bm{b}\Vert^2-\Vert \bm{a}-\bm{b}\Vert^2$ to expand $(b)$ in \eqref{ineqq:main3_12} and performing some algebraic manipulation, we simplify the above inequality as follows:
\begin{align}
    \hspace{-10mm}\mathbb{E}_{s,\hat{\lambda}}&\left[\mathfrak{L}(\bm{\omega}^{(k+1)})\right] \leq \mathfrak{L}(\bm{\omega}^{(k)}) - \frac{\eta_{_k}\ell^{(k)}}{2} \mathbb{E}_s\Bigg[\Bigg\Vert\nabla{\mathfrak{L}(\bm{\omega}^{(k)})}\Bigg\Vert^2+\Bigg\Vert \sum_{u\in \mathcal{U}}\frac{D_{u}}{D \ell^{(k)}}\sum_{\ell=1}^{\ell^{(k)}}  {\nabla  \mathfrak{L}_u(\bm{\omega}^{(k),\ell-1}_{u})}\Bigg\Vert^2\nonumber\\
    &-\Bigg\Vert \nabla{\mathfrak{L}(\bm{\omega}^{(k)})}-  \sum_{u\in \mathcal{U}}\frac{D_{u}}{D \ell^{(k)}}\sum_{\ell=1}^{\ell^{(k)}}  {\nabla  \mathfrak{L}_u(\bm{\omega}^{(k),\ell-1}_{u})}\Bigg\Vert^2\Bigg]\nonumber\\
    &+ \frac{\beta\eta_{_k}^2}{2}  \sum_{u\in \mathcal{U}}\frac{\mathbb{E}_{\hat{\lambda}}\left[\widehat{\lambda}_{u}^{(k)}\right]}{ \left(\ell^{(k)}\right)^2}\mathbb{E}_s\Bigg[\Bigg\Vert \sum_{\ell=1}^{\ell^{(k)}}\mathscr{N}_u^{\mathsf{G},(k),\ell}\Bigg\Vert^2\Bigg]+ \beta\eta_{_k}^2\left(\ell^{(k)}\right)^2 \mathbb{E}_s\Bigg[\Bigg\Vert  \sum_{u\in \mathcal{U}}\frac{D_{u}}{D \ell^{(k)}}\sum_{\ell=1}^{\ell^{(k)}} {\nabla  \mathfrak{L}_u(\bm{\omega}^{(k),\ell-1}_{u})}\Bigg\Vert^2\Bigg]\nonumber\\
    &+ \beta\eta_{_k}^2 \left(\ell^{(k)}\right)^2\mathbb{E}_{s,\hat{\lambda}}\left[\left\Vert\mathscr{U}^{(k)}\right\Vert^2\right]+ \frac{\eta_{_k}\ell^{(k)}}{4} \mathbb{E}_s\left[\left\Vert\nabla{\mathfrak{L}(\bm{\omega}^{(k)})} \right\Vert^2\right]+ \ell^{(k)} |\mathcal{U}|\eta_k^3\sum_{u\in \mathcal{U}}\mathbb{E}_s\left[\left\Vert\frac{1}{\ell^{(k)}}\sum_{\ell=1}^{\ell^{(k)}} {\nabla  \mathfrak{L}_u(\bm{\omega}^{(k),\ell-1}_{u})} \right\Vert^2\right]\nonumber\\
    %%%%%%%
    &=\mathfrak{L}(\bm{\omega}^{(k)}) - \frac{\eta_{_k}\ell^{(k)}}{2} \mathbb{E}_s\Bigg[\Bigg\Vert\nabla{\mathfrak{L}(\bm{\omega}^{(k)})}\Bigg\Vert^2\Bigg]- \frac{\eta_{_k}\ell^{(k)}}{2} \mathbb{E}_s\Bigg[\Bigg\Vert \sum_{u\in \mathcal{U}}\frac{D_{u}}{D \ell^{(k)}}\sum_{\ell=1}^{\ell^{(k)}}  {\nabla  \mathfrak{L}_u(\bm{\omega}^{(k),\ell-1}_{u})}\Bigg\Vert^2\Bigg]\nonumber\\
    &+ \frac{\eta_{_k}\ell^{(k)}}{2} \mathbb{E}_s\Bigg[\Bigg\Vert \nabla{\mathfrak{L}(\bm{\omega}^{(k)})}-  \sum_{u\in \mathcal{U}}\frac{D_{u}}{D \ell^{(k)}}\sum_{\ell=1}^{\ell^{(k)}}  {\nabla  \mathfrak{L}_u(\bm{\omega}^{(k),\ell-1}_{u})}\Bigg\Vert^2\Bigg]\nonumber\\
    &+ \frac{\beta\eta_{_k}^2\left(\ell^{(k)}\right)^2}{2}  \sum_{u\in \mathcal{U}}\frac{\mathbb{E}_{\hat{\lambda}}\left[\widehat{\lambda}_{u}^{(k)}\right]}{ \left(\ell^{(k)}\right)^2}\mathbb{E}_s\Bigg[\Bigg\Vert \sum_{\ell=1}^{\ell^{(k)}}\mathscr{N}_u^{\mathsf{G},(k),\ell}\Bigg\Vert^2\Bigg]+ \beta\eta_{_k}^2\left(\ell^{(k)}\right)^2 \mathbb{E}_s\Bigg[\Bigg\Vert  \sum_{u\in \mathcal{U}}\frac{D_{u}}{D \ell^{(k)}}\sum_{\ell=1}^{\ell^{(k)}} {\nabla  \mathfrak{L}_u(\bm{\omega}^{(k),\ell-1}_{u})}\Bigg\Vert^2\Bigg]\nonumber\\
    &+ \beta\eta_{_k}^2\left(\ell^{(k)}\right)^2 \mathbb{E}_{s,\hat{\lambda}}\left[\left\Vert\mathscr{U}^{(k)}\right\Vert^2\right]+ \frac{\eta_{_k}\ell^{(k)}}{4} \mathbb{E}_s\left[\left\Vert\nabla{\mathfrak{L}(\bm{\omega}^{(k)})} \right\Vert^2\right]+ \ell^{(k)} |\mathcal{U}|\eta_k^3\sum_{u\in \mathcal{U}}\mathbb{E}_s\left[\left\Vert\frac{1}{\ell^{(k)}}\sum_{\ell=1}^{\ell^{(k)}} {\nabla  \mathfrak{L}_u(\bm{\omega}^{(k),\ell-1}_{u})} \right\Vert^2\right]\nonumber\\
    %%%%%%%
    &=\mathfrak{L}(\bm{\omega}^{(k)}) - \frac{\eta_{_k}\ell^{(k)}}{4} \mathbb{E}_s\Bigg[\Bigg\Vert\nabla{\mathfrak{L}(\bm{\omega}^{(k)})}\Bigg\Vert^2\Bigg]+\underbrace{\eta_{_k}\left(\beta\eta_{_k}-\frac{1}{2}\right)\ell^{(k)} \mathbb{E}_s\Bigg[\Bigg\Vert \sum_{u\in \mathcal{U}}\frac{D_{u}}{D \ell^{(k)}}\sum_{\ell=1}^{\ell^{(k)}}  {\nabla  \mathfrak{L}_u(\bm{\omega}^{(k),\ell-1}_{u})}\Bigg\Vert^2\Bigg]}_{(a)}\nonumber\\
    &+ \frac{\eta_{_k}\ell^{(k)}}{2} \mathbb{E}_s\Bigg[\Bigg\Vert \nabla{\mathfrak{L}(\bm{\omega}^{(k)})}-  \sum_{u\in \mathcal{U}}\frac{D_{u}}{D \ell^{(k)}}\sum_{\ell=1}^{\ell^{(k)}}  {\nabla  \mathfrak{L}_u(\bm{\omega}^{(k),\ell-1}_{u})}\Bigg\Vert^2\Bigg]\nonumber\\
    &+ \frac{\beta\eta_{_k}^2\left(\ell^{(k)}\right)^2}{2}  \sum_{u\in \mathcal{U}}\frac{\mathbb{E}_{\hat{\lambda}}\left[\widehat{\lambda}_{u}^{(k)}\right]}{ \left(\ell^{(k)}\right)^2}\mathbb{E}_s\Bigg[\Bigg\Vert \sum_{\ell=1}^{\ell^{(k)}}\mathscr{N}_u^{\mathsf{G},(k),\ell}\Bigg\Vert^2\Bigg]+ \beta\eta^2_{_k}\left(\ell^{(k)}\right)^2 \mathbb{E}_{s,\hat{\lambda}}\left[\left\Vert\mathscr{U}^{(k)}\right\Vert^2\right]\nonumber\\
    &+ \ell^{(k)} |\mathcal{U}|\eta_k^3\sum_{u\in \mathcal{U}}\mathbb{E}_s\left[\left\Vert\frac{1}{\ell^{(k)}}\sum_{\ell=1}^{\ell^{(k)}} {\nabla  \mathfrak{L}_u(\bm{\omega}^{(k),\ell-1}_{u})} \right\Vert^2\right].
\end{align}
Assuming 
\begin{equation}\label{ineqq:qqeta_cond1}
 \eta_{_k}\le\frac{1}{2\beta}   
\end{equation}
makes $(a)$ in the above expression negative and thus can be removed from the bound. Moreover, $\eta_{_k}\le\frac{1}{2\beta}$ implies $1+\beta\eta_{_k}\le\frac{3}{2}$. Applying this result to the above bound gives us
\begin{align}\label{ineqq:main4_1}
    \mathbb{E}_{s,\hat{\lambda}}&\left[\mathfrak{L}(\bm{\omega}^{(k+1)})\right] \leq \mathfrak{L}(\bm{\omega}^{(k)}) - \frac{\eta_{_k}\ell^{(k)}}{4} \mathbb{E}_s\Bigg[\Bigg\Vert\nabla{\mathfrak{L}(\bm{\omega}^{(k)})}\Bigg\Vert^2\Bigg] + \frac{\eta_{_k}\ell^{(k)}}{2} \mathbb{E}_s\Bigg[\Bigg\Vert \nabla{\mathfrak{L}(\bm{\omega}^{(k)})}-  \sum_{u\in \mathcal{U}}\frac{D_{u}}{D \ell^{(k)}}\sum_{\ell=1}^{\ell^{(k)}}  {\nabla  \mathfrak{L}_u(\bm{\omega}^{(k),\ell-1}_{u})}\Bigg\Vert^2\Bigg]\nonumber\\
    &+ \frac{\beta\eta_{_k}^2\left(\ell^{(k)}\right)^2}{2}  \sum_{u\in \mathcal{U}}\frac{\mathbb{E}_{\hat{\lambda}}\left[\widehat{\lambda}_{u}^{(k)}\right]}{ \left(\ell^{(k)}\right)^2}\mathbb{E}_s\Bigg[\underbrace{\Bigg\Vert \sum_{\ell=1}^{\ell^{(k)}}\mathscr{N}_u^{\mathsf{G},(k),\ell}\Bigg\Vert^2}_{(a)}\Bigg]+ \beta\eta^2_{_k}\left(\ell^{(k)}\right)^2 \mathbb{E}_{s,\hat{\lambda}}\left[\left\Vert\mathscr{U}^{(k)}\right\Vert^2\right]\nonumber\\
    &+ \ell^{(k)} |\mathcal{U}|\eta_k^3\sum_{u\in \mathcal{U}}\mathbb{E}_s\left[\left\Vert\frac{1}{\ell^{(k)}}\sum_{\ell=1}^{\ell^{(k)}} {\nabla  \mathfrak{L}_u(\bm{\omega}^{(k),\ell-1}_{u})} \right\Vert^2\right].
\end{align}
Using $\Vert \sum_{i=1}^{n}\bm{x}_i\Vert^2=\sum_{i=1}^{n}\Vert\bm{x}_i\Vert^2+\sum_{\substack{j=1\\j\neq i}}^{n}\left\langle\bm{x}_i,\bm{x}_j\right\rangle$ to expand $(a)$ in \eqref{ineqq:main4_1} and performing some algebraic manipulation, we rewrite \eqref{ineqq:main4_1} as follows: 
\begin{align}\label{ineqq:main4}
    \mathbb{E}_{s,\hat{\lambda}}\left[\mathfrak{L}(\bm{\omega}^{(k+1)})\right] &\leq \mathfrak{L}(\bm{\omega}^{(k)}) - \frac{\eta_{_k}\ell^{(k)}}{4} \mathbb{E}_s\Bigg[\Bigg\Vert\nabla{\mathfrak{L}(\bm{\omega}^{(k)})}\Bigg\Vert^2\Bigg] + \frac{\eta_{_k}\ell^{(k)}}{2} \mathbb{E}_s\Bigg[\Bigg\Vert \nabla{\mathfrak{L}(\bm{\omega}^{(k)})}-  \sum_{u\in \mathcal{U}}\frac{D_{u}}{D \ell^{(k)}}\sum_{\ell=1}^{\ell^{(k)}}  {\nabla  \mathfrak{L}_u(\bm{\omega}^{(k),\ell-1}_{u})}\Bigg\Vert^2\Bigg]\nonumber\\
    &+ \beta\eta^2_{_k}\left(\ell^{(k)}\right)^2 \mathbb{E}_{s,\hat{\lambda}}\left[\left\Vert\mathscr{U}^{(k)}\right\Vert^2\right]+ \ell^{(k)} |\mathcal{U}|\eta_k^3\sum_{u\in \mathcal{U}}\mathbb{E}_s\left[\left\Vert\frac{1}{\ell^{(k)}}\sum_{\ell=1}^{\ell^{(k)}} {\nabla  \mathfrak{L}_u(\bm{\omega}^{(k),\ell-1}_{u})} \right\Vert^2\right]\nonumber\\
    &+ \frac{\beta\eta_{_k}^2\left(\ell^{(k)}\right)^2}{2}  \sum_{u\in \mathcal{U}}\frac{\mathbb{E}_{\hat{\lambda}}\left[\widehat{\lambda}_{u}^{(k)}\right]}{ \left(\ell^{(k)}\right)^2}\left(\mathbb{E}_s\Bigg[\sum_{\ell=1}^{\ell^{(k)}}\Bigg\Vert\mathscr{N}_u^{\mathsf{G},(k),\ell}\Bigg\Vert^2\Bigg]+\underbrace{\mathbb{E}_s\Bigg[\sum_{\ell=1}^{\ell^{(k)}}\sum_{\substack{\ell'=1,\\\ell'\neq \ell}}^{\ell^{(k)}}\left\langle\mathscr{N}_u^{\mathsf{G},(k),\ell},\mathscr{N}_u^{\mathsf{G},(k),\ell'}\right\rangle\Bigg]}_{(x_4)}\right)\nonumber\\
    &\overset{(i)}{=}\mathfrak{L}(\bm{\omega}^{(k)}) - \frac{\eta_{_k}\ell^{(k)}}{4} \mathbb{E}_s\Bigg[\Bigg\Vert\nabla{\mathfrak{L}(\bm{\omega}^{(k)})}\Bigg\Vert^2\Bigg] + \frac{\eta_{_k}\ell^{(k)}}{2} \mathbb{E}_s\Bigg[\Bigg\Vert \nabla{\mathfrak{L}(\bm{\omega}^{(k)})}-  \sum_{u\in \mathcal{U}}\frac{D_{u}}{D \ell^{(k)}}\sum_{\ell=1}^{\ell^{(k)}}  {\nabla  \mathfrak{L}_u(\bm{\omega}^{(k),\ell-1}_{u})}\Bigg\Vert^2\Bigg]\nonumber\\
    &+ \beta\eta^2_{_k}\left(\ell^{(k)}\right)^2 \mathbb{E}_{s,\hat{\lambda}}\left[\left\Vert\mathscr{U}^{(k)}\right\Vert^2\right]+ \ell^{(k)} |\mathcal{U}|\eta_k^3\sum_{u\in \mathcal{U}}\mathbb{E}_s\left[\left\Vert\frac{1}{\ell^{(k)}}\sum_{\ell=1}^{\ell^{(k)}} {\nabla  \mathfrak{L}_u(\bm{\omega}^{(k),\ell-1}_{u})} \right\Vert^2\right]\nonumber\\
    &+ \frac{\beta\eta_{_k}^2\left(\ell^{(k)}\right)^2}{2}  \sum_{u\in \mathcal{U}}\frac{\mathbb{E}_{\hat{\lambda}}\left[\widehat{\lambda}_{u}^{(k)}\right]}{ \left(\ell^{(k)}\right)^2}\sum_{\ell=1}^{\ell^{(k)}}\mathbb{E}_s\Bigg[\Bigg\Vert\mathscr{N}_u^{\mathsf{G},(k),\ell}\Bigg\Vert^2\Bigg],
\end{align}
where $(i)$ uses the fact that $(x_4)$ is zero since the noise of gradient estimation of a specific node is independent and zero mean. We next aim to simplify $\mathbb{E}_s\left[\left\Vert\mathscr{N}^{\mathsf{G},(k)}_{u}\right\Vert^2\right]$. Considering the result of~\cite{lohr2019sampling} (Chapter 3, Eq. (3.5)), we have 
\begin{align}\label{eqq:varGrad}
    \mathbb{E}_s\left[\left\Vert\mathscr{N}^{\mathsf{G},(k)}_{u}\right\Vert^2\right]=\mathbb{E}_s\left[\left\Vert \sum_{d\in \mathcal{B}^{(k),\ell}_{u}} \hspace{-3mm} {\frac{\nabla  f(\bm{\omega}^{(k),\ell-1}_{u},d)}{{B}_{u}}} - \sum_{d\in\mathcal{D}_{u}} \hspace{-3mm} {\frac{\nabla  f(\bm{\omega}^{(k),\ell-1}_{u},d)}{D_{u}}}\right\Vert^2\right]=\left(1-\frac{{B}_{u}}{D_{u}} \right) \frac{\left(\sigma_{u}^{\ell-1}\right)^2}{{B}_{u}},
\end{align}
where $\sigma_{u}^{\ell-1}$ denotes the \textit{variance of the gradients} evaluated at the particular local gradient descent iteration $\ell-1$ for the parameter realization $\bm{\omega}^{(k),\ell-1}_{u}$, and $\left(\sigma_{u}^{\ell{-}1}\right)^2$ is calculated as follows:
\begin{align}\label{eqq:dataVar0}
     \left(\sigma_{u}^{\ell{-}1}\right)^2&= \frac{\sum_{d\in\mathcal{D}_{u} } \Big\Vert \nabla  f(\bm{\omega}^{(k),\ell{-}1}_{u},d){-}{\sum_{\tilde{d}\in\mathcal{D}_{u}} }\frac{\nabla  f(\bm{\omega}^{(k),\ell{-}1}_{u},\tilde{d})}{D_{u}}\Big\Vert^2}{D_{u}{-}1}
     \nonumber\\
     &=\frac{\sum_{d\in\mathcal{D}_{u} }\frac{1}{\left(D_{u}\right)^2} \Big\Vert D_{u}\nabla  f(\bm{\omega}^{(D),\ell{-}1}_{u},d){-}{\sum_{\tilde{d}\in\mathcal{D}_{u}} }{\nabla  f(\bm{\omega}^{(k),\ell{-}1}_{u},\tilde{d})}\Big\Vert^2}{D_{u}{-}1}.
\end{align}
Using the Cauchy-Schwarz inequality, we can bound \eqref{eqq:dataVar0} as follows:
 \begin{align}\label{eqq:dataVar}
     &\left(\sigma_{u}^{\ell{-}1}\right)^2\leq \frac{\sum_{d\in\mathcal{D}_{u} }\frac{D_{u}{-}1}{\left(D_{u}\right)^2} \sum_{\tilde{d}\in\mathcal{D}_{u}} \Big\Vert \nabla  f(\bm{\omega}^{(k),\ell{-}1}_{u},d){-}{\nabla  f(\bm{\omega}^{(k),\ell{-}1}_{u},\tilde{d})}\Big\Vert^2}{D_{u}{-}1}
     \nonumber\\&
     \leq \frac{\sum_{d\in\mathcal{D}_{u} }\frac{(D_{u}{-}1)\Theta^2}{\left(D_{u}\right)^2} \sum_{\tilde{d}\in\mathcal{D}_{u}} \Big\Vert \bm{d}{-}\tilde{\bm{d}}\Big\Vert^2}{D_{u}{-}1}\nonumber\\
     &= \frac{(D_{u}{-}1)\Theta^2}{\left(D_{u}\right)^2}\frac{\sum_{d\in\mathcal{D}_{u} } \sum_{\tilde{d}\in\mathcal{D}_{u}} \Big\Vert \bm{d}{-}\tilde{\bm{d}}{+}\bm{\mu}_{u}{-}\bm{\mu}_{u}\Big\Vert^2}{D_{u}{-}1}\nonumber\\
     &=\frac{(D_{u}{-}1)\Theta^2}{\left(D_{u}\right)^2}\frac{\displaystyle\sum_{d\in\mathcal{D}_{u} } \sum_{\tilde{d}\in\mathcal{D}_{u}} \left[\Big\Vert \bm{d}{-}  \bm{\mu}_{u} \Big\Vert^2 {+} \Big\Vert \tilde{\bm{d}}{-}  \bm{\mu}_{u}\Big\Vert^2 {-} 2\left\langle\bm{d}{-}  \bm{\mu}_{u},\tilde{\bm{d}}{-}  \bm{\mu}_{u}\right\rangle \right]}{D_{u}{-}1}\nonumber\\
     &\overset{(ii)}{=}\frac{(D_{u}{-}1)\Theta^2}{\left(D_{u}\right)^2}\frac{ D_{u} \sum_{d\in\mathcal{D}_{u} } \Big\Vert \bm{d}{-}  \bm{\mu}_{u} \Big\Vert^2 {+}  D_{u} \sum_{\tilde{d}\in\mathcal{D}_{u}} \Big\Vert \tilde{\bm{d}}{-}  \bm{\mu}_{u}\Big\Vert^2}{D_{u}{-}1}\nonumber\\
     &=\frac{2(D_{u}{-}1)\Theta^2}{D_{u}}\left(\sigma_{u}\right)^2, 
\end{align}
where $\bm{\mu}_{u}$ and $\sigma_{u}$ denote the mean and sample variance of data points in dataset $D_{u}$, which are gradient independent. Further, $\Theta {=} \max_{u{\in}\mathcal{U}}\{\Theta_{u} \}$. Also, $\bm{d}$ refers to the feature vector of data point $d$. Furthermore, $(ii)$ used the fact that $\sum_{d\in\mathcal{D}_{u} } (\bm{d}-  \bm{\mu}_{u}) =\bm{0}$. Replacing the above result in~\eqref{eqq:varGrad}, inequality \eqref{ineqq:main4} can be written as follows:
\begin{align}\label{ineqq:main5}
    \mathbb{E}_{s,\hat{\lambda}}&\left[\mathfrak{L}(\bm{\omega}^{(k+1)})\right] \leq \mathfrak{L}(\bm{\omega}^{(k)}) - \frac{\eta_{_k}\ell^{(k)}}{4} \mathbb{E}_s\Bigg[\Bigg\Vert\nabla{\mathfrak{L}(\bm{\omega}^{(k)})}\Bigg\Vert^2\Bigg]+ \frac{\eta_{_k}\ell^{(k)}}{2} \underbrace{\mathbb{E}_s\Bigg[\Bigg\Vert \nabla{\mathfrak{L}(\bm{\omega}^{(k)})}-  \sum_{u\in \mathcal{U}}\frac{D_{u}}{D \ell^{(k)}}\sum_{\ell=1}^{\ell^{(k)}}  {\nabla  \mathfrak{L}_u(\bm{\omega}^{(k),\ell-1}_{u})}\Bigg\Vert^2\Bigg]}_{(d)}\nonumber\\
    &+ \beta\eta^2_{_k}\left(\ell^{(k)}\right)^2 \mathbb{E}_{s,\hat{\lambda}}\left[\left\Vert\mathscr{U}^{(k)}\right\Vert^2\right]+ \ell^{(k)} |\mathcal{U}|\eta_k^3\sum_{u\in \mathcal{U}}\mathbb{E}_s\left[\left\Vert\frac{1}{\ell^{(k)}}\sum_{\ell=1}^{\ell^{(k)}} {\nabla  \mathfrak{L}_u(\bm{\omega}^{(k),\ell-1}_{u})} \right\Vert^2\right]\nonumber\\
    &~~~~+ \frac{\beta\eta_{_k}^2\left(\ell^{(k)}\right)^2}{2}  \sum_{u\in \mathcal{U}}\frac{\mathbb{E}_{\hat{\lambda}}\left[\widehat{\lambda}_{u}^{(k)}\right] }{\ell^{(k)}}\left(1-\frac{{B}_{u}}{D_{u}} \right) \frac{2(D_{u}-1)\Theta^2\left(\sigma_{u}\right)^2}{D_{u}{B}_{u}}.
\end{align}
In the following, we aim to bound term $(d)$.
\begin{align}\label{eqq:res3}
        (d) &=\mathbb{E}_s\left[\left\Vert \nabla{\mathfrak{L}(\bm{\omega}^{(k)})}- \sum_{u\in \mathcal{U}}\frac{D_{u}}{D \ell^{(k)}} \sum_{\ell=1}^{\ell^{(k)}}  {\nabla  \mathfrak{L}_u(\bm{\omega}^{(k),\ell-1}_{u})}\right\Vert^2\right]\nonumber\\
    &\overset{(i)}{\leq}    \sum_{u\in \mathcal{U}}\frac{D_{u}}{D }\mathbb{E}_s\left[\left\Vert \nabla{\mathfrak{L}_u(\bm{\omega}^{(k)})}-  \frac{1}{\ell^{(k)}} \sum_{\ell=1}^{\ell^{(k)}}  {\nabla  \mathfrak{L}_u(\bm{\omega}^{(k),\ell-1}_{u})} \right\Vert^2\right]\nonumber\\
    &\overset{(ii)}{\leq}  \sum_{u\in \mathcal{U}}\frac{D_{u}}{D \ell^{(k)}}\sum_{\ell=1}^{\ell^{(k)}} \mathbb{E}_s\left[\left\Vert
   \nabla{\mathfrak{L}_u(\bm{\omega}^{(k)})}- {\nabla  \mathfrak{L}_u(\bm{\omega}^{(k),\ell-1}_{u})}\right\Vert^2\right] \nonumber\\
   &\overset{(iii)}{\leq} \beta^2 \sum_{u\in \mathcal{U}}\frac{D_{u}}{D \ell^{(k)}}\sum_{\ell=1}^{\ell^{(k)}} \mathbb{E}_s\left[\left\Vert\bm{\omega}^{(k)}-\bm{\omega}^{(k),\ell-1}_{u}\right\Vert^2\right],
\end{align}
where in inequalities $(i)$ and $(ii)$ in~\eqref{eqq:res3} we used Jenson's inequality. Inequality $(iii)$ is the result of Assumption \eqref{Assup:lossFun}. To bound $\mathbb{E}_s\left[\left\Vert\bm{\omega}^{(k)}-\bm{\omega}^{(k),\ell-1}_{u}\right\Vert^2\right]$, we take the following steps.
\begin{align}\label{eqq:res4}
    &\mathbb{E}_s\left[\left\Vert\bm{\omega}^{(k)}-\bm{\omega}^{(k),\ell-1}_{u}\right\Vert^2\right]\overset{(i)}{=}\eta_k^2  \mathbb{E}_s\left[\left\Vert
    \sum_{\ell'=1}^{\ell-1}\sum_{d\in \mathcal{B}^{(k),\ell'}_{u}} \hspace{-1mm} {\frac{\nabla  f(\bm{\omega}^{(k),\ell'-1}_{u},d)}{{B}_{u}}}\right\Vert^2 \right]
    \nonumber\\
    %%%%%%
    &= \eta_k^2 \mathbb{E}_s\vast[\Bigg\Vert\sum_{\ell'=1}^{\ell-1} \sum_{d\in \mathcal{B}^{(k),\ell'}_{u}} \hspace{-1mm} {\frac{\nabla  f(\bm{\omega}^{(k),\ell'-1}_{u},d)}{{B}_{u}}}\nonumber-\frac{1}{D_{u}} \sum_{\ell'=1}^{\ell-1} \sum_{d\in\mathcal{D}_{u}}\nabla  f(\bm{\omega}^{(k),\ell'-1}_{u},d) + \frac{1}{D_{u}} \sum_{\ell'=1}^{\ell-1}\sum_{d\in\mathcal{D}_{u}}\nabla  f(\bm{\omega}^{(k),\ell'-1}_{u},d)\Bigg\Vert^2 \vast]\nonumber \nonumber\\
    %%%%%%
    & \overset{(ii)}{\leq }2 \eta_k^2  \mathbb{E}_s\left[\Bigg\Vert\sum_{\ell'=1}^{\ell-1} \sum_{d\in \mathcal{B}^{(k),\ell'}_{u}} \hspace{-1mm} {\frac{\nabla  f(\bm{\omega}^{(k),\ell'-1}_{u},d)}{{B}_{u}}} - \frac{1}{D_{u}} \sum_{\ell'=1}^{\ell-1}\sum_{d\in\mathcal{D}_{u}}\nabla f(\bm{\omega}^{(k),\ell'-1}_{u},d)\Bigg\Vert^2\right]  +2 \eta_k^2 \mathbb{E}_s\left[\Bigg\Vert  \frac{1}{D_{u}} \sum_{\ell'=1}^{\ell-1}\sum_{d\in\mathcal{D}_{u}}\nabla  f(\bm{\omega}^{(k),\ell'-1}_{u},d)\Bigg\Vert^2 \right]\nonumber \nonumber\\
    %%%%%%
    &\overset{(iii)}{=} \underbrace{2 \eta_k^2 \sum_{\ell'=1}^{\ell-1}\mathbb{E}_s\left[\Bigg\Vert\sum_{d\in \mathcal{B}^{(k),\ell'}_{u}} \hspace{-1mm} {\frac{\nabla  f(\bm{\omega}^{(k),\ell'-1}_{u},d)}{{B}_{u}}} - \frac{1}{D_{u}} \sum_{d\in\mathcal{D}_{u}}\nabla f(\bm{\omega}^{(k),\ell'-1}_{u},d)\Bigg\Vert^2\right]}_{(e)}  +\underbrace{2 \eta_k^2 \mathbb{E}_s\left[\Bigg\Vert  \frac{1}{D_{u}} \sum_{\ell'=1}^{\ell-1}\sum_{d\in\mathcal{D}_{u}}\nabla  f(\bm{\omega}^{(k),\ell'-1}_{u},d)\Bigg\Vert^2\right]}_{(f)},
\end{align}
where to obtain~\eqref{eqq:res4}, in equality $(i)$ we used~\eqref{eq:updaterule}, in inequality $(ii)$ we used  Cauchy–Schwarz inequality, and $(iii)$
uses the fact that each local gradient estimation is unbiased (i.e., zero mean) conditioned on its own local parameter and the law of total expectation (across the mini-batches $\ell'$). Similar to~\eqref{eqq:varGrad}, using~\eqref{eqq:dataVar} we upper bound term $(e)$ in~\eqref{eqq:res4} as follows:
\begin{equation}\label{eqq:A2}
    (e) \leq 4 \Theta^2 \eta_k^2 \sum_{\ell'=1}^{\ell-1} \left(1-\frac{{B}_{u}}{D_{u}} \right)  \frac{{(D_{u}-1)}
     \left(\sigma_{u}\right)^2}{D_{u}{B}_{u}}.
\end{equation}
Also, for term $(f)$, we have
\begin{align}\label{eqq:B2}
   (f)&
   \overset{(i)}{\leq} 2 \eta_k^2 (\ell-1)\sum_{\ell'=1}^{\ell-1}\mathbb{E}_s\left[\Bigg\Vert \frac{1}{D_{u}} \sum_{d\in\mathcal{D}_{u}}\nabla  f(\bm{\omega}^{(k),\ell'-1}_{u},d) - \nabla \mathfrak{L}_u(\bm{\omega}^{(k)})+ \nabla \mathfrak{L}_u(\bm{\omega}^{(k)})
    \Bigg\Vert^2 \right]
   \nonumber \nonumber\\
    &\overset{(ii)}{\leq} 4 \eta_k^2 (\ell-1)\sum_{\ell'=1}^{\ell-1}\mathbb{E}_s\left[\Bigg\Vert \nabla  \mathfrak{L}_u(\bm{\omega}^{(k),\ell'-1}_{u}) - \nabla \mathfrak{L}_u(\bm{\omega}^{(k)})\Bigg\Vert^2\right] + 4 \eta_k^2 (\ell-1)\sum_{\ell'=1}^{\ell-1} \Bigg\Vert\nabla \mathfrak{L}_u(\bm{\omega}^{(k)})
    \Bigg\Vert^2 
    \nonumber \nonumber\\
    &\leq 4\eta_k^2\beta^2 (\ell-1) \sum_{\ell'=1}^{\ell-1}\mathbb{E}_s\left[\Bigg\Vert \bm{\omega}^{(k),\ell'-1}_{u} - \bm{\omega}^{(k)}\Bigg\Vert^2\right]+ 4 \eta_k^2 (\ell-1)\sum_{\ell'=1}^{\ell-1} \Bigg\Vert\nabla \mathfrak{L}_u(\bm{\omega}^{(k)})
    \Bigg\Vert^2 ,
\end{align}
where inequalities $(i)$ and $(ii)$ are obtained via Cauchy-Schwarz inequality. Replacing the result of~\eqref{eqq:A2} and~\eqref{eqq:B2} back in~\eqref{eqq:res4} we have
\begin{align}
  \mathbb{E}_s\left[  \left\Vert
  \bm{\omega}^{(k)}-
     \bm{\omega}^{(k),\ell-1}_{u}
    \right\Vert^2\right] \leq& 4 \Theta^2 \eta_k^2 \sum_{\ell'=1}^{\ell-1} \left(1-\frac{{B}_{u}}{D_{u}} \right)  \frac{{(D_{u}-1)}
     \left(\sigma_{u}\right)^2}{D_{u}{B}_{u}} \nonumber \nonumber\\
     & + 4\eta_k^2\beta^2 (\ell-1) \sum_{\ell'=1}^{\ell-1}\mathbb{E}_s\left[\Bigg\Vert \bm{\omega}^{(k),\ell'-1}_{u} - \bm{\omega}^{(k)}\Bigg\Vert^2\right]+ 4 \eta_k^2 (\ell-1)\sum_{\ell'=1}^{\ell-1} \Bigg\Vert\nabla \mathfrak{L}_u(\bm{\omega}^{(k)})
    \Bigg\Vert^2, 
\end{align}
which implies
\begin{align}
   &\sum_{\ell=1}^{\ell^{(k)}} \mathbb{E}_s\left[\left\Vert\bm{\omega}^{(k)}-\bm{\omega}^{(k),\ell-1}_{u}\right\Vert^2\right] \leq 4 \Theta^2 \eta_k^2\sum_{\ell=1}^{\ell^{(k)}} \sum_{\ell'=1}^{\ell-1} \left(1-\frac{{B}_{u}}{D_{u}} \right)  \frac{{(D_{u}-1)} \left(\sigma_{u}\right)^2}{D_{u}{B}_{u}} \nonumber \\
     & + 4\eta_k^2\beta^2 \sum_{\ell=1}^{\ell^{(k)}}(\ell-1) \sum_{\ell'=1}^{\ell-1}\mathbb{E}_s\left[\Bigg\Vert \bm{\omega}^{(k),\ell'-1}_{u} - \bm{\omega}^{(k)}\Bigg\Vert^2\right]+ 4 \eta_k^2 \sum_{\ell=1}^{\ell^{(k)}}(\ell-1)\sum_{\ell'=1}^{\ell-1} \Bigg\Vert\nabla \mathfrak{L}_u(\bm{\omega}^{(k)}) \Bigg\Vert^2\nonumber\\
     &\leq 4 \Theta^2 \eta_k^2 \left(\ell^{(k)}\right)\left(\ell^{(k)}-1\right) \left(1-\frac{{B}_{u}}{D_{u}} \right)  \frac{{(D_{u}-1)}\left(\sigma_{u}\right)^2}{D_{u}{B}_{u}} \nonumber\\
     & + 4\eta_k^2\beta^2 \left(\ell^{(k)}\right)\left(\ell^{(k)}-1\right) \sum_{\ell=1}^{\ell^{(k)}}\mathbb{E}_s\left[\Bigg\Vert \bm{\omega}^{(k),\ell-1}_{u} - \bm{\omega}^{(k)}\Bigg\Vert^2\right]+ 4 \eta_k^2 \left(\ell^{(k)}\right)\left(\ell^{(k)}-1\right) \sum_{\ell=1}^{\ell^{(k)}} \Bigg\Vert\nabla \mathfrak{L}_u(\bm{\omega}^{(k)})\Bigg\Vert^2.
\end{align}
Assuming $\eta_k \leq \left(2\beta\sqrt{\ell^{(k)}(\ell^{(k)}-1)}\right)^{-1},\forall u$, the above inequality implies
\begin{align}\label{eqq:f_bound}
    \hspace{-8mm}\sum_{\ell=1}^{\ell^{(k)}} \mathbb{E}_s\left[\left\Vert\bm{\omega}^{(k)}- \bm{\omega}^{(k),\ell-1}_{u} \right\Vert^2\right] \leq&\frac{4 \Theta^2 \eta_k^2 \ell^{(k)}\left(\ell^{(k)}-1\right)}{1- 4\eta_k^2\beta^2 \ell^{(k)}\left(\ell^{(k)}-1\right)} \left(1-\frac{{B}_{u}}{D_{u}} \right)  \frac{{(D_{u}-1)}\left(\sigma_{u}\right)^2}{D_{u}{B}_{u}}  + \frac{4 \eta_k^2 \left(\ell^{(k)}\right)^2\left(\ell^{(k)}-1\right)}{1- 4\eta_k^2\beta^2 \ell^{(k)}\left(\ell^{(k)}-1\right)}  \Bigg\Vert\nabla \mathfrak{L}_u(\bm{\omega}^{(k)})\Bigg\Vert^2.
\end{align}
Replacing this result back in~\eqref{eqq:res3} we have
\begin{align}\label{eqq:res5}
    (d) &\leq {4 \beta^2\Theta^2 \eta_k^2} \sum_{u\in \mathcal{U}}\frac{D_{u}}{D \ell^{(k)}}\frac{\left(\ell^{(k)}\right)\left(\ell^{(k)}-1\right)}{1- 4\eta_k^2\beta^2 \ell^{(k)}\left(\ell^{(k)}-1\right)}\left(1-\frac{{B}_{u}}{D_{u}} \right)  \frac{{(D_{u}-1)}
    \left(\sigma_{u}\right)^2}{D_{u}{B}_{u}}\nonumber  \nonumber\\
    %%%%%%%%%
    &+ {4 \eta_k^2\beta^2 } \sum_{u\in \mathcal{U}}\frac{D_{u}}{D \ell^{(k)}}\frac{\left(\ell^{(k)}\right)^2\left(\ell^{(k)}-1\right)}{{1- 4\eta_k^2\beta^2 \ell^{(k)}\left(\ell^{(k)}-1\right)}}  \Bigg\Vert\nabla \mathfrak{L}_u(\bm{\omega}^{(k)}) \Bigg\Vert^2\nonumber\\
    &= {4 \beta^2\Theta^2 \eta_k^2} \sum_{u\in \mathcal{U}}\frac{D_{u}}{D \ell^{(k)}}\frac{\left(\ell^{(k)}\right)\left(\ell^{(k)}-1\right)}{1- 4\eta_k^2\beta^2 \ell^{(k)}\left(\ell^{(k)}-1\right)}\left(1-\frac{{B}_{u}}{D_{u}} \right)  \frac{{(D_{u}-1)}
    \left(\sigma_{u}\right)^2}{D_{u}{B}_{u}}\nonumber  \nonumber\\
    %%%%%%%%%
    &+ {4 \eta_k^2\beta^2 } \sum_{u\in \mathcal{U}}\frac{D_{u}}{D \ell^{(k)}}\frac{\left(\ell^{(k)}\right)^2\left(\ell^{(k)}-1\right)}{{1- 4\eta_k^2\beta^2 \ell^{(k)}\left(\ell^{(k)}-1\right)}}  \Bigg\Vert\nabla \mathfrak{L}_u(\bm{\omega}^{(k)}) - \nabla \mathfrak{L}(\bm{\omega}^{(k)}) + \nabla \mathfrak{L}(\bm{\omega}^{(k)}) \Bigg\Vert^2\nonumber\\
    &\le {4 \beta^2\Theta^2 \eta_k^2} \sum_{u\in \mathcal{U}}\frac{D_{u}}{D \ell^{(k)}}\frac{\left(\ell^{(k)}\right)\left(\ell^{(k)}-1\right)}{1- 4\eta_k^2\beta^2 \ell^{(k)}\left(\ell^{(k)}-1\right)}\left(1-\frac{{B}_{u}}{D_{u}} \right)  \frac{{(D_{u}-1)}
    \left(\sigma_{u}\right)^2}{D_{u}{B}_{u}}  \nonumber\\
    %%%%%%%%%
    &+ {8 \eta_k^2\beta^2 } \sum_{u\in \mathcal{U}}\frac{D_{u}}{D}\frac{\left(\ell^{(k)}\right) \left(\ell^{(k)}-1\right)}{{1- 4\eta_k^2\beta^2 \ell^{(k)}\left(\ell^{(k)}-1\right)}}  \Bigg\Vert\nabla \mathfrak{L}_u(\bm{\omega}^{(k)}) - \nabla \mathfrak{L}(\bm{\omega}^{(k)}) \Bigg\Vert^2\nonumber\\
    &+ {8 \eta_k^2\beta^2 } \sum_{u\in \mathcal{U}}\frac{D_{u}}{D}\frac{\left(\ell^{(k)}\right) \left(\ell^{(k)}-1\right)}{{1- 4\eta_k^2\beta^2 \ell^{(k)}\left(\ell^{(k)}-1\right)}}  \Bigg\Vert  \nabla \mathfrak{L}(\bm{\omega}^{(k)}) \Bigg\Vert^2 
\end{align}
Performing some algebraic manipulations give us
 \begin{align}\label{eqq:res09444}
    (d) \leq &{4 \beta^2\Theta^2 \eta_k^2} \sum_{u\in \mathcal{U}}\frac{D_{u}}{D \ell^{(k)}}\frac{\left(\ell^{(k)}\right)\left(\ell^{(k)}-1\right)}{1- 4\eta_k^2\beta^2 \ell^{(k)}\left(\ell^{(k)}-1\right)}\left(1-\frac{{B}_{u}}{D_{u}} \right)  \frac{{(D_{u}-1)}
    \left(\sigma_{u}\right)^2}{D_{u}{B}_{u}} \nonumber\\
    %%%%%%%%%
    &+ {8 \eta_k^2\beta^2 } \sum_{u\in \mathcal{U}}\frac{D_{u}}{D}\frac{\left(\ell^{(k)}\right) \left(\ell^{(k)}-1\right)}{{1- 4\eta_k^2\beta^2 \ell^{(k)}\left(\ell^{(k)}-1\right)}}  \mathfrak{X}^{(k)}_u +  \frac{{8 \eta_k^2\beta^2 }\left(\ell^{(k)}\right) \left(\ell^{(k)}-1\right)}{{1- 4\eta_k^2\beta^2 \ell^{(k)}\left(\ell^{(k)}-1\right)}}  \Bigg\Vert  \nabla \mathfrak{L}(\bm{\omega}^{(k)}) \Bigg\Vert^2 
\end{align}
 Replacing the above result back in~\eqref{ineqq:main5} and gathering the term leads to
 \begin{align}\label{ineqq:main6}
    \mathbb{E}_{s,\hat{\lambda}}&\left[\mathfrak{L}(\bm{\omega}^{(k+1)})\right] \leq \mathfrak{L}(\bm{\omega}^{(k)}) - \frac{\eta_{_k}\ell^{(k)}}{4} \mathbb{E}_s\Bigg[\Bigg\Vert\nabla{\mathfrak{L}(\bm{\omega}^{(k)})}\Bigg\Vert^2\Bigg] +     \sum_{u\in \mathcal{U}}\frac{D_{u}}{D \ell^{(k)}}\frac{{\ell^{(k)}  2 \beta^2\Theta^2 \eta_k^3}\left(\ell^{(k)}\right)\left(\ell^{(k)}-1\right)}{1- 4\eta_k^2\beta^2 \ell^{(k)}\left(\ell^{(k)}-1\right)}\left(1-\frac{{B}_{u}}{D_{u}} \right)  \frac{{(D_{u}-1)}
    \left(\sigma_{u}\right)^2}{D_{u}{B}_{u}}\nonumber  \nonumber\\
    %%%%%%%%%
    &+    \sum_{u\in \mathcal{U}}\frac{D_{u}}{D}\frac{ {4 \eta_k^3 \ell^{(k)} \beta^2 }\left(\ell^{(k)}\right) \left(\ell^{(k)}-1\right)}{{1- 4\eta_k^2\beta^2 \ell^{(k)}\left(\ell^{(k)}-1\right)}}  \mathfrak{X}^{(k)}_u +    \frac{{4 \eta_k^3\ell^{(k)}\beta^2 }\left(\ell^{(k)}\right) \left(\ell^{(k)}-1\right)}{{1- 4\eta_k^2\beta^2 \ell^{(k)}\left(\ell^{(k)}-1\right)}}  \Bigg\Vert  \nabla \mathfrak{L}(\bm{\omega}^{(k)}) \Bigg\Vert^2 \nonumber\\
    & + \beta\eta^2_{_k}\left(\ell^{(k)}\right)^2 \mathbb{E}_{s,\hat{\lambda}}\left[\left\Vert\mathscr{U}^{(k)}\right\Vert^2\right]+ \ell^{(k)} |\mathcal{U}|\eta_k^3\sum_{u\in \mathcal{U}}\mathbb{E}_s\left[\left\Vert\frac{1}{\ell^{(k)}}\sum_{\ell=1}^{\ell^{(k)}} {\nabla  \mathfrak{L}_u(\bm{\omega}^{(k),\ell-1}_{u})} \right\Vert^2\right]\nonumber\\
    &~~~~+ \frac{\beta\eta_{_k}^2\left(\ell^{(k)}\right)^2}{2}  \sum_{u\in \mathcal{U}}\frac{ \mathbb{E}_{\hat{\lambda}}\left[\widehat{\lambda}_{u}^{(k)}\right] }{  \ell^{(k)}}\left(1-\frac{{B}_{u}}{D_{u}} \right) \frac{2(D_{u}-1)\Theta^2\left(\sigma_{u}\right)^2}{D_{u}{B}_{u}}.
\end{align}
Considering the definition of the noise of node selection presented in \eqref{eqq:user_selection_noise}, we next aim to bound $\mathbb{E}_{s,\hat{\lambda}}\left[\left\Vert\mathscr{U}^{(k)}\right\Vert^2\right]$ bellow.
\begin{align}
    \mathbb{E}_{s,\hat{\lambda}}\left[\left\Vert\mathscr{U}^{(k)}\right\Vert^2\right]&= \mathbb{E}_{s,\hat{\lambda}}\left[\left\Vert \sum_{u\in \mathcal{U}}\frac{\widehat{\lambda}_{u}^{(k)}}{ \ell^{(k)}}\sum_{\ell=1}^{\ell^{(k)}} {\nabla  \mathfrak{L}_u(\bm{\omega}^{(k),\ell-1}_{u})}- \sum_{u\in \mathcal{U}}\frac{D_{u}}{D \ell^{(k)}}\sum_{\ell=1}^{\ell^{(k)}} {\nabla  \mathfrak{L}_u(\bm{\omega}^{(k),\ell-1}_{u})} \right\Vert^2\right]\nonumber\\
    &= \mathbb{E}_{s,\hat{\lambda}}\left[\left\Vert \sum_{u\in \mathcal{U}} \left(\frac{\widehat{\lambda}_{u}^{(k)}D-D_{u}}{\ell^{(k)}D}\right)\sum_{\ell=1}^{\ell^{(k)}} {\nabla  \mathfrak{L}_u(\bm{\omega}^{(k),\ell-1}_{u})} \right\Vert^2\right]\nonumber\\
    &\overset{(i)}{\le}  |\mathcal{U}|  \sum_{u\in \mathcal{U}}\frac{\mathbb{E}_{\hat{\lambda}}\left[\left(\frac{\widehat{\lambda}_{u}^{(k)}D-D_{u}}{D}\right)^2\right]}{\left(\ell^{(k)}\right)^2}\mathbb{E}_s\left[\left\Vert \sum_{\ell=1}^{\ell^{(k)}} {\nabla  \mathfrak{L}_u(\bm{\omega}^{(k),\ell-1}_{u})} \right\Vert^2\right]
\end{align}
where (i) uses Cauchy-Schwarz inequality. Replacing the above result back in \eqref{ineqq:main6} and gathering terms give us
 \begin{align}\label{ineqq:main66}
    \mathbb{E}_{s,\hat{\lambda}}&\left[\mathfrak{L}(\bm{\omega}^{(k+1)})\right] \leq \mathfrak{L}(\bm{\omega}^{(k)}) - \frac{\eta_{_k}\ell^{(k)}}{4} \mathbb{E}_s\Bigg[\Bigg\Vert\nabla{\mathfrak{L}(\bm{\omega}^{(k)})}\Bigg\Vert^2\Bigg]\nonumber\\
    &+     \sum_{u\in \mathcal{U}}\frac{D_{u}}{D \ell^{(k)}}\frac{{\ell^{(k)}  2 \beta^2\Theta^2 \eta_k^3}\left(\ell^{(k)}\right)\left(\ell^{(k)}-1\right)}{1- 4\eta_k^2\beta^2 \ell^{(k)}\left(\ell^{(k)}-1\right)}\left(1-\frac{{B}_{u}}{D_{u}} \right)  \frac{{(D_{u}-1)}
    \left(\sigma_{u}\right)^2}{D_{u}{B}_{u}}\nonumber  \nonumber\\
    %%%%%%%%%
    &+    \sum_{u\in \mathcal{U}}\frac{D_{u}}{D}\frac{ {4 \eta_k^3 \ell^{(k)} \beta^2 }\left(\ell^{(k)}\right) \left(\ell^{(k)}-1\right)}{{1- 4\eta_k^2\beta^2 \ell^{(k)}\left(\ell^{(k)}-1\right)}}  \mathfrak{X}^{(k)}_u+    \frac{{4 \eta_k^3\ell^{(k)}\beta^2 }\left(\ell^{(k)}\right) \left(\ell^{(k)}-1\right)}{{1- 4\eta_k^2\beta^2 \ell^{(k)}\left(\ell^{(k)}-1\right)}}  \Bigg\Vert  \nabla \mathfrak{L}(\bm{\omega}^{(k)}) \Bigg\Vert^2 \nonumber\\
    & + \beta\eta^2_{_k}\left(\ell^{(k)}\right)^2  |\mathcal{U}|  \sum_{u\in \mathcal{U}}\frac{\mathbb{E}_{\hat{\lambda}}\left[\left(\frac{\widehat{\lambda}_{u}^{(k)}D-D_{u}}{D}\right)^2\right]}{\left(\ell^{(k)}\right)^2}\mathbb{E}_s\left[\left\Vert \sum_{\ell=1}^{\ell^{(k)}} {\nabla  \mathfrak{L}_u(\bm{\omega}^{(k),\ell-1}_{u})} \right\Vert^2\right] + \ell^{(k)} |\mathcal{U}|\eta_k^3\sum_{u\in \mathcal{U}}\mathbb{E}_s\left[\left\Vert\frac{1}{\ell^{(k)}}\sum_{\ell=1}^{\ell^{(k)}} {\nabla  \mathfrak{L}_u(\bm{\omega}^{(k),\ell-1}_{u})} \right\Vert^2\right]\nonumber\\
    &+ \frac{\beta\eta_{_k}^2\left(\ell^{(k)}\right)^2}{2}  \sum_{u\in \mathcal{U}}\frac{ \mathbb{E}_{\hat{\lambda}}\left[\widehat{\lambda}_{u}^{(k)}\right] }{  \ell^{(k)}}\left(1-\frac{{B}_{u}}{D_{u}} \right) \frac{2(D_{u}-1)\Theta^2\left(\sigma_{u}\right)^2}{D_{u}{B}_{u}}\nonumber\\
    &= \mathfrak{L}(\bm{\omega}^{(k)}) - \frac{\eta_{_k}\ell^{(k)}}{4} \mathbb{E}_s\Bigg[\Bigg\Vert\nabla{\mathfrak{L}(\bm{\omega}^{(k)})}\Bigg\Vert^2\Bigg] +     \sum_{u\in \mathcal{U}}\frac{D_{u}}{D \ell^{(k)}}\frac{{\ell^{(k)}  2 \beta^2\Theta^2 \eta_k^3}\left(\ell^{(k)}\right)\left(\ell^{(k)}-1\right)}{1- 4\eta_k^2\beta^2 \ell^{(k)}\left(\ell^{(k)}-1\right)}\left(1-\frac{{B}_{u}}{D_{u}} \right)  \frac{{(D_{u}-1)}
    \left(\sigma_{u}\right)^2}{D_{u}{B}_{u}}\nonumber  \nonumber\\
    %%%%%%%%%
    &+    \sum_{u\in \mathcal{U}}\frac{D_{u}}{D}\frac{ {4 \eta_k^3 \ell^{(k)} \beta^2 }\left(\ell^{(k)}\right) \left(\ell^{(k)}-1\right)}{{1- 4\eta_k^2\beta^2 \ell^{(k)}\left(\ell^{(k)}-1\right)}}  \mathfrak{X}^{(k)}_u +    \frac{{4 \eta_k^3\ell^{(k)}\beta^2 }\left(\ell^{(k)}\right) \left(\ell^{(k)}-1\right)}{{1- 4\eta_k^2\beta^2 \ell^{(k)}\left(\ell^{(k)}-1\right)}}  \Bigg\Vert  \nabla \mathfrak{L}(\bm{\omega}^{(k)}) \Bigg\Vert^2 \nonumber\\
    & + \frac{\ell^{(k)} |\mathcal{U}|\eta_k^2}{\left(\ell^{(k)}\right)^2} \sum_{u\in \mathcal{U}}\left(\frac{\beta \ell^{(k)}\left(\mathbb{E}_{\hat{\lambda}}\left[\widehat{\lambda}_{u}^{(k)}\right]D^2-2\mathbb{E}_{\hat{\lambda}}\left[\widehat{\lambda}_{u}^{(k)}\right]D D_{u}+ D^2_{u}\right)+\eta_k D^2}{D^2}\right)\underbrace{\mathbb{E}_s\left[\left\Vert \sum_{\ell=1}^{\ell^{(k)}} {\nabla  \mathfrak{L}_u(\bm{\omega}^{(k),\ell-1}_{u})} \right\Vert^2\right]}_{(a)} \nonumber\\
    &+ \frac{\beta\eta_{_k}^2\left(\ell^{(k)}\right)^2}{2}  \sum_{u\in \mathcal{U}}\frac{ \mathbb{E}_{\hat{\lambda}}\left[\widehat{\lambda}_{u}^{(k)} \right]}{  \ell^{(k)}}\left(1-\frac{{B}_{u}}{D_{u}} \right) \frac{2(D_{u}-1)\Theta^2\left(\sigma_{u}\right)^2}{D_{u}{B}_{u}}.
\end{align}
Using the same technique utilized in \eqref{eqq:B2} and performing some algebraic manipulation give us an upper bound for $(a)$ in the above inequality. 
\begin{align}
    (a)&\le \mathbb{E}_s\Bigg[ 2\beta^2 (\ell^{(k)}) \sum_{\ell=1}^{\ell^{(k)}}\Bigg\Vert \bm{\omega}^{(k)}-\bm{\omega}^{(k),\ell-1}_{u}\Bigg\Vert^2+ 2(\ell^{(k)})\sum_{\ell=1}^{\ell^{(k)}} \Bigg\Vert\nabla \mathfrak{L}_u(\bm{\omega}^{(k)})\Bigg\Vert^2\Bigg]\nonumber\\
    & =  2\beta^2 (\ell^{(k)}) \sum_{\ell=1}^{\ell^{(k)}} \mathbb{E}_s\Bigg[\Bigg\Vert \bm{\omega}^{(k)}-\bm{\omega}^{(k),\ell-1}_{u}\Bigg\Vert^2\Bigg]+ 2(\ell^{(k)})^2 \Bigg\Vert\nabla \mathfrak{L}_u(\bm{\omega}^{(k)})\Bigg\Vert^2\Bigg]
\end{align}
Substituting $\sum_{\ell=1}^{\ell^{(k)}}\mathbb{E}_s\left[\left\Vert \bm{\omega}^{(k)}-\bm{\omega}^{(k),\ell-1}_{u}\right\Vert^2 \right]$ from \eqref{eqq:f_bound} into the above inequality and performing some algebraic operations give us 
 \begin{align}
    \hspace{-10mm}(a)&\le  \frac{8 \Theta^2 \beta^2 \eta_k^2 \left(\ell^{(k)}\right)^2\left(\ell^{(k)}-1\right)}{1- 4\eta_k^2\beta^2 \ell^{(k)}\left(\ell^{(k)}-1\right)} \left(1-\frac{{B}_{u}}{D_{u}} \right)  \frac{{(D_{u}-1)}\left(\sigma_{u}\right)^2}{D_{u}{B}_{u}}  +  \frac{8\beta^2 \eta_k^2 \left(\ell^{(k)}\right)^3\left(\ell^{(k)}-1\right)}{1- 4\eta_k^2\beta^2 \ell^{(k)}\left(\ell^{(k)}-1\right)}  \Bigg\Vert\nabla \mathfrak{L}_u(\bm{\omega}^{(k)})\Bigg\Vert^2+ 2(\ell^{(k)})^2 \Bigg\Vert\nabla \mathfrak{L}_u(\bm{\omega}^{(k)})\Bigg\Vert^2\nonumber\\
     &= \frac{8 \Theta^2 \beta^2 \eta_k^2 \left(\ell^{(k)}\right)^2\left(\ell^{(k)}-1\right)}{1- 4\eta_k^2\beta^2 \ell^{(k)}\left(\ell^{(k)}-1\right)} \left(1-\frac{{B}_{u}}{D_{u}} \right)  \frac{{(D_{u}-1)}\left(\sigma_{u}\right)^2}{D_{u}{B}_{u}}  +  \left(\frac{8\beta^2 \eta_k^2 \left(\ell^{(k)}\right)^3\left(\ell^{(k)}-1\right)}{1- 4\eta_k^2\beta^2 \ell^{(k)}\left(\ell^{(k)}-1\right)}  + 2(\ell^{(k)})^2 \right) \Bigg\Vert\nabla \mathfrak{L}_u(\bm{\omega}^{(k)})\Bigg\Vert^2\nonumber\\
     &= \frac{8 \Theta^2 \beta^2 \eta_k^2 \left(\ell^{(k)}\right)^2\left(\ell^{(k)}-1\right)}{1- 4\eta_k^2\beta^2 \ell^{(k)}\left(\ell^{(k)}-1\right)} \left(1-\frac{{B}_{u}}{D_{u}} \right)  \frac{{(D_{u}-1)}\left(\sigma_{u}\right)^2}{D_{u}{B}_{u}}  +  \left(\frac{2(\ell^{(k)})^2}{1- 4\eta_k^2\beta^2 \ell^{(k)}\left(\ell^{(k)}-1\right)}  \right) \Bigg\Vert\nabla \mathfrak{L}_u(\bm{\omega}^{(k)})\Bigg\Vert^2
\end{align}
Letting $q_{u}^{(k)}= \left(\frac{\beta \ell^{(k)}\left(\mathbb{E}_{\hat{\lambda}}\left[\widehat{\lambda}_{u}^{(k)}\right]D^2-2\mathbb{E}_{\hat{\lambda}}\left[\widehat{\lambda}_{u}^{(k)}\right]D D_{u}+ D^2_{u}\right)+\eta_k D^2}{D^2}\right)$ and replacing the above result back in \eqref{ineqq:main66} give us
\begin{align}\label{ineqq:main67}
    \mathbb{E}_{s,\hat{\lambda}}&\left[\mathfrak{L}(\bm{\omega}^{(k+1)})\right] \leq \mathfrak{L}(\bm{\omega}^{(k)}) - \frac{\eta_{_k}\ell^{(k)}}{4} \mathbb{E}_s\Bigg[\Bigg\Vert\nabla{\mathfrak{L}(\bm{\omega}^{(k)})}\Bigg\Vert^2\Bigg]\nonumber\\
    &+     \sum_{u\in \mathcal{U}}\frac{D_{u}}{D \ell^{(k)}}\frac{{\ell^{(k)}  2 \beta^2\Theta^2 \eta_k^3}\left(\ell^{(k)}\right)\left(\ell^{(k)}-1\right)}{1- 4\eta_k^2\beta^2 \ell^{(k)}\left(\ell^{(k)}-1\right)}\left(1-\frac{{B}_{u}}{D_{u}} \right)  \frac{{(D_{u}-1)}
    \left(\sigma_{u}\right)^2}{D_{u}{B}_{u}}\nonumber  \nonumber\\
    %%%%%%%%%
    &+    \sum_{u\in \mathcal{U}}\frac{D_{u}}{D}\frac{ {4 \eta_k^3 \ell^{(k)} \beta^2 }\left(\ell^{(k)}\right) \left(\ell^{(k)}-1\right)}{{1- 4\eta_k^2\beta^2 \ell^{(k)}\left(\ell^{(k)}-1\right)}}  \mathfrak{X}^{(k)}_u +    \frac{{4 \eta_k^3\ell^{(k)}\beta^2 }\left(\ell^{(k)}\right) \left(\ell^{(k)}-1\right)}{{1- 4\eta_k^2\beta^2 \ell^{(k)}\left(\ell^{(k)}-1\right)}}  \Bigg\Vert  \nabla \mathfrak{L}(\bm{\omega}^{(k)}) \Bigg\Vert^2 \nonumber\\
    & + \frac{\ell^{(k)} |\mathcal{U}|\eta_k^2}{\left(\ell^{(k)}\right)^2} \sum_{u\in \mathcal{U}}q_{u}^{(k)}\frac{8 \Theta^2 \beta^2 \eta_k^2 \left(\ell^{(k)}\right)^2\left(\ell^{(k)}-1\right)}{1- 4\eta_k^2\beta^2 \ell^{(k)}\left(\ell^{(k)}-1\right)} \left(1-\frac{{B}_{u}}{D_{u}} \right)  \frac{{(D_{u}-1)}\left(\sigma_{u}\right)^2}{D_{u}{B}_{u}}\nonumber\\
     & +  \frac{\ell^{(k)} |\mathcal{U}|\eta_k^2}{\left(\ell^{(k)}\right)^2} \sum_{u\in \mathcal{U}}q_{u}^{(k)}\left(\frac{2(\ell^{(k)})^2}{1- 4\eta_k^2\beta^2 \ell^{(k)}\left(\ell^{(k)}-1\right)}  \right) \Bigg\Vert\nabla \mathfrak{L}_u(\bm{\omega}^{(k)})\Bigg\Vert^2\nonumber\\
    &+ \frac{\beta\eta_{_k}^2\left(\ell^{(k)}\right)^2}{2}  \sum_{u\in \mathcal{U}}\frac{ \mathbb{E}_{\hat{\lambda}}\left[\widehat{\lambda}_{u}^{(k)} \right]}{  \ell^{(k)}}\left(1-\frac{{B}_{u}}{D_{u}} \right) \frac{2(D_{u}-1)\Theta^2\left(\sigma_{u}\right)^2}{D_{u}{B}_{u}}\nonumber\\
    &= \mathfrak{L}(\bm{\omega}^{(k)}) - \frac{\eta_{_k}\ell^{(k)}}{4} \mathbb{E}_s\Bigg[\Bigg\Vert\nabla{\mathfrak{L}(\bm{\omega}^{(k)})}\Bigg\Vert^2\Bigg] +     \sum_{u\in \mathcal{U}}\frac{D_{u}}{D \ell^{(k)}}\frac{{\ell^{(k)}  2 \beta^2\Theta^2 \eta_k^3}\left(\ell^{(k)}\right)\left(\ell^{(k)}-1\right)}{1- 4\eta_k^2\beta^2 \ell^{(k)}\left(\ell^{(k)}-1\right)}\left(1-\frac{{B}_{u}}{D_{u}} \right)  \frac{{(D_{u}-1)}
    \left(\sigma_{u}\right)^2}{D_{u}{B}_{u}}\nonumber  \nonumber\\
    %%%%%%%%%
    &+    \sum_{u\in \mathcal{U}}\frac{D_{u}}{D}\frac{ {4 \eta_k^3 \ell^{(k)} \beta^2 }\left(\ell^{(k)}\right) \left(\ell^{(k)}-1\right)}{{1- 4\eta_k^2\beta^2 \ell^{(k)}\left(\ell^{(k)}-1\right)}}  \mathfrak{X}^{(k)}_u +    \frac{{4 \eta_k^3\ell^{(k)}\beta^2 }\left(\ell^{(k)}\right) \left(\ell^{(k)}-1\right)}{{1- 4\eta_k^2\beta^2 \ell^{(k)}\left(\ell^{(k)}-1\right)}}  \Bigg\Vert  \nabla \mathfrak{L}(\bm{\omega}^{(k)}) \Bigg\Vert^2 \nonumber\\
    & + \sum_{u\in \mathcal{U}}\frac{8 q_{u}^{(k)} \ell^{(k)} |\mathcal{U}| \Theta^2 \beta^2 \eta_k^4 \left(\ell^{(k)}-1\right)}{1- 4\eta_k^2\beta^2 \ell^{(k)}\left(\ell^{(k)}-1\right)} \left(1-\frac{{B}_{u}}{D_{u}} \right)  \frac{{(D_{u}-1)}\left(\sigma_{u}\right)^2}{D_{u}{B}_{u}}\nonumber\\
     & +  \sum_{u\in \mathcal{U}}\frac{2 q_{u}^{(k)}\ell^{(k)} |\mathcal{U}|\eta_k^2}{1- 4\eta_k^2\beta^2 \ell^{(k)}\left(\ell^{(k)}-1\right)} \Bigg\Vert\nabla \mathfrak{L}_u(\bm{\omega}^{(k)})\Bigg\Vert^2 + \frac{\beta\eta_{_k}^2\left(\ell^{(k)}\right)^2}{2}  \sum_{u\in \mathcal{U}}\frac{ \mathbb{E}_{\hat{\lambda}}\left[\widehat{\lambda}_{u}^{(k)} \right]}{  \ell^{(k)}}\left(1-\frac{{B}_{u}}{D_{u}} \right) \frac{2(D_{u}-1)\Theta^2\left(\sigma_{u}\right)^2}{D_{u}{B}_{u}}.
\end{align}
Performing algebraic manipulations and gathering terms lead to
\begin{align}\label{ineqq:main67}
    \mathbb{E}_{s,\hat{\lambda}}&\left[\mathfrak{L}(\bm{\omega}^{(k+1)})\right] \leq \mathfrak{L}(\bm{\omega}^{(k)}) - \frac{\eta_{_k}\ell^{(k)}}{4} \mathbb{E}_s\Bigg[\Bigg\Vert\nabla{\mathfrak{L}(\bm{\omega}^{(k)})}\Bigg\Vert^2\Bigg]+\frac{{4 \eta_k^3\ell^{(k)}\beta^2 }\left(\ell^{(k)}\right) \left(\ell^{(k)}-1\right)}{{1- 4\eta_k^2\beta^2 \ell^{(k)}\left(\ell^{(k)}-1\right)}}  \Bigg\Vert  \nabla \mathfrak{L}(\bm{\omega}^{(k)}) \Bigg\Vert^2\nonumber\\
    &+     \sum_{u\in \mathcal{U}}\frac{D_{u}}{D \ell^{(k)}}\frac{{\ell^{(k)}  2 \beta^2\Theta^2 \eta_k^3}\left(\ell^{(k)}\right)\left(\ell^{(k)}-1\right)}{1- 4\eta_k^2\beta^2 \ell^{(k)}\left(\ell^{(k)}-1\right)}\left(1-\frac{{B}_{u}}{D_{u}} \right)  \frac{{(D_{u}-1)}
    \left(\sigma_{u}\right)^2}{D_{u}{B}_{u}}\nonumber  \nonumber\\
    %%%%%%%%%
    &+    \sum_{u\in \mathcal{U}}\frac{D_{u}}{D}\frac{ {4 \eta_k^3 \ell^{(k)} \beta^2 }\left(\ell^{(k)}\right) \left(\ell^{(k)}-1\right)}{{1- 4\eta_k^2\beta^2 \ell^{(k)}\left(\ell^{(k)}-1\right)}}  \mathfrak{X}^{(k)}_u+ \sum_{u\in \mathcal{U}}\frac{8 q_{u}^{(k)} \ell^{(k)} |\mathcal{U}| \Theta^2 \beta^2 \eta_k^4 \left(\ell^{(k)}-1\right)}{1- 4\eta_k^2\beta^2 \ell^{(k)}\left(\ell^{(k)}-1\right)} \left(1-\frac{{B}_{u}}{D_{u}} \right)  \frac{{(D_{u}-1)}\left(\sigma_{u}\right)^2}{D_{u}{B}_{u}}\nonumber\\
     & +  \sum_{u\in \mathcal{U}}\frac{2 q_{u}^{(k)}\ell^{(k)} |\mathcal{U}|\eta_k^2}{1- 4\eta_k^2\beta^2 \ell^{(k)}\left(\ell^{(k)}-1\right)} \Bigg\Vert\nabla \mathfrak{L}_u(\bm{\omega}^{(k)})-\nabla \mathfrak{L}(\bm{\omega}^{(k)})+\nabla \mathfrak{L}(\bm{\omega}^{(k)})\Bigg\Vert^2\nonumber\\
    &+ \frac{\beta\eta_{_k}^2\left(\ell^{(k)}\right)^2}{2}  \sum_{u\in \mathcal{U}}\frac{ \mathbb{E}_{\hat{\lambda}}\left[\widehat{\lambda}_{u}^{(k)} \right]}{  \ell^{(k)}}\left(1-\frac{{B}_{u}}{D_{u}} \right) \frac{2(D_{u}-1)\Theta^2\left(\sigma_{u}\right)^2}{D_{u}{B}_{u}}\nonumber\\
    %%%%%%%%%
    &\overset{(i)}{\le} \mathfrak{L}(\bm{\omega}^{(k)}) - \frac{\eta_{_k}\ell^{(k)}}{4} \mathbb{E}_s\Bigg[\Bigg\Vert\nabla{\mathfrak{L}(\bm{\omega}^{(k)})}\Bigg\Vert^2\Bigg]+\frac{{4 \eta_k^3\ell^{(k)}\beta^2 }\left(\ell^{(k)}\right) \left(\ell^{(k)}-1\right)}{{1- 4\eta_k^2\beta^2 \ell^{(k)}\left(\ell^{(k)}-1\right)}}  \Bigg\Vert  \nabla \mathfrak{L}(\bm{\omega}^{(k)}) \Bigg\Vert^2\nonumber\\
    &+     \sum_{u\in \mathcal{U}}\frac{D_{u}}{D \ell^{(k)}}\frac{{\ell^{(k)}  2 \beta^2\Theta^2 \eta_k^3}\left(\ell^{(k)}\right)\left(\ell^{(k)}-1\right)}{1- 4\eta_k^2\beta^2 \ell^{(k)}\left(\ell^{(k)}-1\right)}\left(1-\frac{{B}_{u}}{D_{u}} \right)  \frac{{(D_{u}-1)}
    \left(\sigma_{u}\right)^2}{D_{u}{B}_{u}}\nonumber  \nonumber\\
    &+    \sum_{u\in \mathcal{U}}\frac{D_{u}}{D}\frac{ {4 \eta_k^3 \ell^{(k)} \beta^2 }\left(\ell^{(k)}\right) \left(\ell^{(k)}-1\right)}{{1- 4\eta_k^2\beta^2 \ell^{(k)}\left(\ell^{(k)}-1\right)}}  \mathfrak{X}^{(k)}_u +  \sum_{u\in \mathcal{U}}\frac{2 q_{u}^{(k)}\ell^{(k)} |\mathcal{U}|\eta_k^2}{1- 4\eta_k^2\beta^2 \ell^{(k)}\left(\ell^{(k)}-1\right)} \Bigg\Vert\nabla \mathfrak{L}_u(\bm{\omega}^{(k)})-\nabla \mathfrak{L}(\bm{\omega}^{(k)})\Bigg\Vert^2\nonumber\\
    & + \sum_{u\in \mathcal{U}}\frac{8 q_{u}^{(k)} \ell^{(k)} |\mathcal{U}| \Theta^2 \beta^2 \eta_k^4 \left(\ell^{(k)}-1\right)}{1- 4\eta_k^2\beta^2 \ell^{(k)}\left(\ell^{(k)}-1\right)} \left(1-\frac{{B}_{u}}{D_{u}} \right)  \frac{{(D_{u}-1)}\left(\sigma_{u}\right)^2}{D_{u}{B}_{u}}\nonumber\\
     & +  \sum_{u\in \mathcal{U}}\frac{2 q_{u}^{(k)}\ell^{(k)} |\mathcal{U}|\eta_k^2}{1- 4\eta_k^2\beta^2 \ell^{(k)}\left(\ell^{(k)}-1\right)} \Bigg\Vert\nabla \mathfrak{L}(\bm{\omega}^{(k)})\Bigg\Vert^2 + \frac{\beta\eta_{_k}^2\left(\ell^{(k)}\right)^2}{2}  \sum_{u\in \mathcal{U}}\frac{ \mathbb{E}_{\hat{\lambda}}\left[\widehat{\lambda}_{u}^{(k)} \right]}{  \ell^{(k)}}\left(1-\frac{{B}_{u}}{D_{u}} \right) \frac{2(D_{u}-1)\Theta^2\left(\sigma_{u}\right)^2}{D_{u}{B}_{u}}
\end{align}
where in inequality $(i)$ we have used the Cauchy-Schwarz inequality $\Vert \mathbf{a}+\mathbf{b} \Vert^2\leq 2 \Vert \mathbf{a} \Vert^2+2\Vert \mathbf{b} \Vert^2$. Doing simplification gives us
\begin{align}\label{ineqq:main6753}
    \mathbb{E}_{s,\hat{\lambda}}&\left[\mathfrak{L}(\bm{\omega}^{(k+1)})\right] \leq \mathfrak{L}(\bm{\omega}^{(k)}) - \frac{\eta_{_k}\ell^{(k)}}{4} \mathbb{E}_s\Bigg[\Bigg\Vert\nabla{\mathfrak{L}(\bm{\omega}^{(k)})}\Bigg\Vert^2\Bigg]+\frac{{4 \eta_k^3\ell^{(k)}\beta^2 }\left(\ell^{(k)}\right) \left(\ell^{(k)}-1\right)}{{1- 4\eta_k^2\beta^2 \ell^{(k)}\left(\ell^{(k)}-1\right)}}  \Bigg\Vert  \nabla \mathfrak{L}(\bm{\omega}^{(k)}) \Bigg\Vert^2\nonumber\\
    &+     \sum_{u\in \mathcal{U}}\frac{D_{u}}{D \ell^{(k)}}\frac{{\ell^{(k)}  2 \beta^2\Theta^2 \eta_k^3}\left(\ell^{(k)}\right)\left(\ell^{(k)}-1\right)}{1- 4\eta_k^2\beta^2 \ell^{(k)}\left(\ell^{(k)}-1\right)}\left(1-\frac{{B}_{u}}{D_{u}} \right)  \frac{{(D_{u}-1)}
    \left(\sigma_{u}\right)^2}{D_{u}{B}_{u}}  \nonumber\\
    &+    \sum_{u\in \mathcal{U}}\frac{D_{u}}{D}\frac{ {4 \eta_k^3 \ell^{(k)} \beta^2 }\left(\ell^{(k)}\right) \left(\ell^{(k)}-1\right)}{{1- 4\eta_k^2\beta^2 \ell^{(k)}\left(\ell^{(k)}-1\right)}}  \mathfrak{X}^{(k)}_u +  \sum_{u\in \mathcal{U}}\frac{2 q_{u}^{(k)}\ell^{(k)} |\mathcal{U}|\eta_k^2}{1- 4\eta_k^2\beta^2 \ell^{(k)}\left(\ell^{(k)}-1\right)} \mathfrak{X}_{u}^{(k)}\nonumber\\
    & + \sum_{u\in \mathcal{U}}\frac{8 q_{u}^{(k)} \ell^{(k)} |\mathcal{U}| \Theta^2 \beta^2 \eta_k^4 \left(\ell^{(k)}-1\right)}{1- 4\eta_k^2\beta^2 \ell^{(k)}\left(\ell^{(k)}-1\right)} \left(1-\frac{{B}_{u}}{D_{u}} \right)  \frac{{(D_{u}-1)}\left(\sigma_{u}\right)^2}{D_{u}{B}_{u}}\nonumber\\
     & +  \sum_{u\in \mathcal{U}}\frac{2 q_{u}^{(k)}\ell^{(k)} |\mathcal{U}|\eta_k^2}{1- 4\eta_k^2\beta^2 \ell^{(k)}\left(\ell^{(k)}-1\right)} G^2 + \frac{\beta\eta_{_k}^2\left(\ell^{(k)}\right)^2}{2}  \sum_{u\in \mathcal{U}}\frac{ \mathbb{E}_{\hat{\lambda}}\left[\widehat{\lambda}_{u}^{(k)} \right]}{  \ell^{(k)}}\left(1-\frac{{B}_{u}}{D_{u}} \right) \frac{2(D_{u}-1)\Theta^2\left(\sigma_{u}\right)^2}{D_{u}{B}_{u}}\nonumber\\
    %%%%%%%
    &\le \mathfrak{L}(\bm{\omega}^{(k)}) + \eta_{_k}\ell^{(k)} \left(\frac{{4 \eta_k^2\beta^2 }\left(\ell^{(k)}\right) \left(\ell^{(k)}-1\right)}{{1- 4\eta_k^2\beta^2 \ell^{(k)}\left(\ell^{(k)}-1\right)}}-\frac{1}{4}\right) \mathbb{E}_s\Bigg[\Bigg\Vert\nabla{\mathfrak{L}(\bm{\omega}^{(k)})}\Bigg\Vert^2\Bigg]\nonumber\\
    &+     \sum_{u\in \mathcal{U}}\frac{D_{u}}{D \ell^{(k)}}\frac{{\ell^{(k)}  2 \beta^2\Theta^2 \eta_k^3}\left(\ell^{(k)}\right)\left(\ell^{(k)}-1\right)}{1- 4\eta_k^2\beta^2 \ell^{(k)}\left(\ell^{(k)}-1\right)}\left(1-\frac{{B}_{u}}{D_{u}} \right)  \frac{{(D_{u}-1)}
    \left(\sigma_{u}\right)^2}{D_{u}{B}_{u}}\nonumber  \nonumber\\
    &+    \sum_{u\in \mathcal{U}}\frac{D_{u}}{D}\frac{ {4 \eta_k^3 \ell^{(k)} \beta^2 }\left(\ell^{(k)}\right) \left(\ell^{(k)}-1\right)+ 2 q_{u}^{(k)}\ell^{(k)} |\mathcal{U}|\eta_k^2}{{1- 4\eta_k^2\beta^2 \ell^{(k)}\left(\ell^{(k)}-1\right)}} \mathfrak{X}_{u}^{(k)}\nonumber\\
    & + \sum_{u\in \mathcal{U}}\frac{8 q_{u}^{(k)} \ell^{(k)} |\mathcal{U}| \Theta^2 \beta^2 \eta_k^4 \left(\ell^{(k)}-1\right)}{1- 4\eta_k^2\beta^2 \ell^{(k)}\left(\ell^{(k)}-1\right)} \left(1-\frac{{B}_{u}}{D_{u}} \right)  \frac{{(D_{u}-1)}\left(\sigma_{u}\right)^2}{D_{u}{B}_{u}}\nonumber\\
     & +  \sum_{u\in \mathcal{U}}\frac{2 q_{u}^{(k)}\ell^{(k)} |\mathcal{U}|\eta_k^2}{1- 4\eta_k^2\beta^2 \ell^{(k)}\left(\ell^{(k)}-1\right)} G^2 + \frac{\beta\eta_{_k}^2\left(\ell^{(k)}\right)^2}{2}  \sum_{u\in \mathcal{U}}\frac{ \mathbb{E}_{\hat{\lambda}}\left[\widehat{\lambda}_{u}^{(k)} \right]}{  \ell^{(k)}}\left(1-\frac{{B}_{u}}{D_{u}} \right) \frac{2(D_{u}-1)\Theta^2\left(\sigma_{u}\right)^2}{D_{u}{B}_{u}}
\end{align}
Assuming
\begin{align}
    \frac{{4 \eta_k^2 \beta^2 }\left(\ell^{(k)}\right) \left(\ell^{(k)}-1\right)}{{1- 4\eta_k^2\beta^2 \ell^{(k)}\left(\ell^{(k)}-1\right)}} <\zeta^{(k)}< \frac{1}{4}
\end{align}
yields
\begin{align}\label{qqeta_cond}
      \eta_k   < \frac{1}{2 \beta}\sqrt{\frac{\zeta^{(k)}}{\left(1+ \zeta^{(k)}\right)  \left(\ell^{(k)}\right) \left(\ell^{(k)}-1\right)}}.
\end{align}
Applying the above condition on \eqref{ineqq:main6753} and performing some algebraic manipulations give us
\begin{equation}
\footnotesize
\begin{aligned}\label{ineqq:main7}
     \mathbb{E}_{s,\hat{\lambda}} \left[\Bigg\Vert  \nabla \mathfrak{L}(\bm{\omega}^{(k)}) \Bigg\Vert^2\right]  &\leq \frac{\mathbb{E}_s \left[\mathfrak{L}(\bm{\omega}^{(k)})\right]- \mathbb{E}_s \left[\mathfrak{L}(\bm{\omega}^{(k+1)})\right]}{\eta_{_k}\ell^{(k)}  \left(\frac{1}{4} - \zeta^{(k)}\right)} +  \frac{1}{\eta_{_k}  \left(\frac{1}{4} - \zeta^{(k)}\right)}\sum_{u\in \mathcal{U}}\frac{D_{u}}{D}\frac{{2 \beta^2\Theta^2 \eta_k^3}\left(\ell^{(k)}-1\right)}{1- 4\eta_k^2\beta^2 \ell^{(k)}\left(\ell^{(k)}-1\right)}\left(1-\frac{{B}_{u}}{D_{u}} \right)  \frac{{(D_{u}-1)}
    \left(\sigma_{u}\right)^2}{D_{u}{B}_{u}}\nonumber  \nonumber\\
    &+    \frac{1}{\eta_{_k}  \left(\frac{1}{4} - \zeta^{(k)}\right)} \sum_{u\in \mathcal{U}}\frac{D_{u}}{D}\frac{ {4 \eta_k^3 \beta^2 }\left(\ell^{(k)}\right) \left(\ell^{(k)}-1\right)+ 2 q_{u}^{(k)}  |\mathcal{U}|\eta_k^2}{{1- 4\eta_k^2\beta^2 \ell^{(k)}\left(\ell^{(k)}-1\right)}} \mathfrak{X}_{u}^{(k)} \nonumber\\
    &+  \frac{1}{\eta_{_k}  \left(\frac{1}{4} - \zeta^{(k)}\right)}\frac{\beta\eta_{_k}^2}{2}  \sum_{u\in \mathcal{U}} \mathbb{E}_{\hat{\lambda}}\left[\widehat{\lambda}_{u}^{(k)} \right]\left(1-\frac{{B}_{u}}{D_{u}} \right) \frac{2(D_{u}-1)\Theta^2\left(\sigma_{u}\right)^2}{D_{u}{B}_{u}}\nonumber\\
    & +  \frac{1}{\eta_{_k}  \left(\frac{1}{4} - \zeta^{(k)}\right)}\sum_{u\in \mathcal{U}}\frac{8 q_{u}^{(k)}  |\mathcal{U}| \Theta^2 \beta^2 \eta_k^4 \left(\ell^{(k)}-1\right)}{1- 4\eta_k^2\beta^2 \ell^{(k)}\left(\ell^{(k)}-1\right)} \left(1-\frac{{B}_{u}}{D_{u}} \right)  \frac{{(D_{u}-1)}\left(\sigma_{u}\right)^2}{D_{u}{B}_{u}} +   \frac{1}{\eta_{_k}  \left(\frac{1}{4} - \zeta^{(k)}\right)}\sum_{u\in \mathcal{U}}\frac{2 q_{u}^{(k)} |\mathcal{U}|\eta_k^2}{1- 4\eta_k^2\beta^2 \ell^{(k)}\left(\ell^{(k)}-1\right)} G^2
\end{aligned}
\end{equation}
Performing some algebraic manipulations and gathering terms gives us
\begin{equation}
\footnotesize
\begin{aligned}\label{ineqq:main78}
     \mathbb{E}_{s,\hat{\lambda}} \left[\Bigg\Vert  \nabla \mathfrak{L}(\bm{\omega}^{(k)}) \Bigg\Vert^2\right]  &\leq \frac{\mathbb{E}_s \left[\mathfrak{L}(\bm{\omega}^{(k)})\right]- \mathbb{E}_s \left[\mathfrak{L}(\bm{\omega}^{(k+1)})\right]}{\eta_{_k}\ell^{(k)}  \left(\frac{1}{4} - \zeta^{(k)}\right)} +  \frac{1}{ \left(\frac{1}{4} - \zeta^{(k)}\right)}\sum_{u\in \mathcal{U}}\frac{D_{u}}{D }\frac{{  2 \beta^2\Theta^2 \eta_k^2} \left(\ell^{(k)}-1\right)}{1- 4\eta_k^2\beta^2 \ell^{(k)}\left(\ell^{(k)}-1\right)}\left(1-\frac{{B}_{u}}{D_{u}} \right)  \frac{{(D_{u}-1)}
    \left(\sigma_{u}\right)^2}{D_{u}{B}_{u}} \\
    &+    \frac{1}{  \left(\frac{1}{4} - \zeta^{(k)}\right)} \sum_{u\in \mathcal{U}}\frac{D_{u}}{D}\frac{ {4 \eta_k^2 \beta^2 }\left(\ell^{(k)}\right) \left(\ell^{(k)}-1\right)+ 2 q_{u}^{(k)} |\mathcal{U}|\eta_k}{{1- 4\eta_k^2\beta^2 \ell^{(k)}\left(\ell^{(k)}-1\right)}} \mathfrak{X}_{u}^{(k)} + \frac{\beta\eta_{_k}}{2 \left(\frac{1}{4} - \zeta^{(k)}\right)}  \sum_{u\in \mathcal{U}}\mathbb{E}_{\hat{\lambda}}\left[\widehat{\lambda}_{u}^{(k)} \right]\left(1-\frac{{B}_{u}}{D_{u}} \right) \frac{2(D_{u}-1)\Theta^2\left(\sigma_{u}\right)^2}{D_{u}{B}_{u}}\\
    & +  \frac{1}{ \left(\frac{1}{4} - \zeta^{(k)}\right)}\sum_{u\in \mathcal{U}}\frac{8 q_{u}^{(k)} |\mathcal{U}| \Theta^2 \beta^2 \eta_k^3 \left(\ell^{(k)}-1\right)}{1- 4\eta_k^2\beta^2 \ell^{(k)}\left(\ell^{(k)}-1\right)} \left(1-\frac{{B}_{u}}{D_{u}} \right)  \frac{{(D_{u}-1)}\left(\sigma_{u}\right)^2}{D_{u}{B}_{u}} +   \frac{1}{\left(\frac{1}{4} - \zeta^{(k)}\right)}\sum_{u\in \mathcal{U}}\frac{2 q_{u}^{(k)} |\mathcal{U}|\eta_k }{1- 4\eta_k^2\beta^2 \ell^{(k)}\left(\ell^{(k)}-1\right)} G^2 
\end{aligned}
\end{equation}
Further, from \eqref{eta_cond_main}, we get
\begin{align}
    \frac{1}{{1- 4\eta_k^2\beta^2 \ell^{(k)}\left(\ell^{(k)}-1\right)}}<\left(\zeta^{(k)}+1\right)
\end{align}
Applying the this condition to \eqref{ineqq:main78} and taking total expectation and averaging across global aggregations lead to
\begin{equation}
\footnotesize
\begin{aligned}\label{ineqq:main79}
     \frac{1}{K} \sum_{k=0}^{K-1}\mathbb{E}_{s,\hat{\lambda}} \left[\Bigg\Vert  \nabla \mathfrak{L}(\bm{\omega}^{(k)}) \Bigg\Vert^2\right]  &\leq \underbrace{\frac{1}{K} \sum_{k=0}^{K-1} \frac{\mathbb{E}_s \left[\mathfrak{L}(\bm{\omega}^{(k)})\right]- \mathbb{E}_s \left[\mathfrak{L}(\bm{\omega}^{(k+1)})\right]}{\eta_{_k}\ell^{(k)}  \left(\frac{1}{4} - \zeta^{(k)}\right)}}_{(a)} +  \frac{1}{K} \sum_{k=0}^{K-1}\mathbb{E}_s \vast[\underbrace{\frac{\left(\zeta^{(k)}+1\right)}{ \left(\frac{1}{4} - \zeta^{(k)}\right)}\sum_{u\in \mathcal{U}}\frac{D_{u}}{D }{  2 \beta^2\Theta^2 \eta_k^2}\left(\ell^{(k)}-1\right)\left(1-\frac{{B}_{u}}{D_{u}} \right)  \frac{{(D_{u}-1)}
    \left(\sigma_{u}\right)^2}{D_{u}{B}_{u}}}_{(b)} \\
    &+    \underbrace{\frac{\left(\zeta^{(k)}+1\right)}{  \left(\frac{1}{4} - \zeta^{(k)}\right)} \sum_{u\in \mathcal{U}}\frac{D_{u}}{D} {4 \eta_k^2 \beta^2 }\left(\ell^{(k)}\right) \left(\ell^{(k)}-1\right)+ 2 q_{u}^{(k)} |\mathcal{U}|\eta_k \mathfrak{X}_{u}^{(k)}}_{(c)} + \underbrace{\frac{\beta\eta_{_k}\ell^{(k)}}{2 \left(\frac{1}{4} - \zeta^{(k)}\right)}  \sum_{u\in \mathcal{U}}\frac{ \mathbb{E}_{\hat{\lambda}}\left[\widehat{\lambda}_{u}^{(k)} \right]}{  \ell^{(k)}}\left(1-\frac{{B}_{u}}{D_{u}} \right) \frac{2(D_{u}-1)\Theta^2\left(\sigma_{u}\right)^2}{D_{u}{B}_{u}}}_{(d)}\\
    & +  \underbrace{\frac{\left(\zeta^{(k)}+1\right)}{ \left(\frac{1}{4} - \zeta^{(k)}\right)}\sum_{u\in \mathcal{U}}8 q_{u}^{(k)} |\mathcal{U}| \Theta^2 \beta^2 \eta_k^3 \left(\ell^{(k)}-1\right) \left(1-\frac{{B}_{u}}{D_{u}} \right)  \frac{{(D_{u}-1)}\left(\sigma_{u}\right)^2}{D_{u}{B}_{u}}}_{(e)} +   \underbrace{\frac{\left(\zeta^{(k)}+1\right)}{\left(\frac{1}{4} - \zeta^{(k)}\right)}\sum_{u\in \mathcal{U}}2 q_{u}^{(k)} |\mathcal{U}|\eta_k  G^2 }_{(f)}\vast]
\end{aligned}
\end{equation}
Assume $\ell_{\mathsf{min}} \le \ell_r^{(k)}\leq \ell_{\mathsf{max}}$  and $\max_k \left\{\zeta^{(k)}\right\} \leq \zeta_{\mathsf{max}}< \frac{1}{4}$ Further, assume 
\begin{equation}\label{eqq:eta_value}
\eta_k = \frac{\alpha}{\sqrt{K}},  
\end{equation}
where $\alpha$ must be chosen such that $\eta_k$ satisfy $ \eta_k   < \min\left\{\frac{1}{2 \beta}\sqrt{\frac{\zeta^{(k)}}{\left(1+ \zeta^{(k)}\right)  \left(\ell^{(k)}\right) \left(\ell^{(k)}-1\right)}}, \frac{1}{2\beta} \right\}$. Moreover, considering that $\mathfrak{L}(\bm{\omega}^{(K)}) \ge \mathfrak{L}^{\star}$, using a telescoping sum over global round iterations we expand term $(a)$ below:
\begin{align}
    (a) \le \frac{1}{K}  \frac{ \mathfrak{L}(\bm{\omega}^{(0)})-  \mathfrak{L}(\bm{\omega}^{(K)})}{\frac{\alpha}{\sqrt{K }}\ell_{\mathsf{min}}   \left(\frac{1}{4} - \zeta_{\mathsf{max}}\right)} \le \frac{ \mathfrak{L}(\bm{\omega}^{(0)})-  \mathfrak{L}^{\star}}{\alpha \sqrt{K }\ell_{\mathsf{min}}   \left(\frac{1}{4} - \zeta_{\mathsf{max}}\right)}
\end{align}
Likewise, we bound term $(b)$ and $(d)$ as follows. Assuming a bounded data sampling noise: $\max_{k,u}\left\{\left(1-\frac{{B}_{u}}{D_{u}} \right)  \frac{{(D_{u}-1)} \left(\sigma_{u}\right)^2}{D_{u}{B}_{u}} \right\} \leq \sigma_{\mathsf{max}}$, we get
\begin{align}
    (b) \le \frac{{  2 \beta^2\Theta^2  \alpha^2 }\left(\ell_{\mathsf{max}}-1\right) \left(\zeta_{\mathsf{max}}+1\right)}{\left(\frac{1}{4} - \zeta_{\mathsf{max}}\right) K} \sigma_{\mathsf{max}}
\end{align}
\begin{align}
    (d)\le \frac{\beta\eta_{_k}\ell^{(k)} 2\Theta^2}{2 \left(\frac{1}{4} - \zeta^{(k)}\right)}  \sum_{u\in \mathcal{U}}\frac{ \mathbb{E}_{\hat{\lambda}}\left[\widehat{\lambda}_{u}^{(k)} \right]}{  \ell^{(k)}}\left(1-\frac{{B}_{u}}{D_{u}} \right) \frac{(D_{u}-1)\left(\sigma_{u}\right)^2}{D_{u}{B}_{u}} \le \frac{\beta\alpha  2\Theta^2}{2 \sqrt{K} \left(\frac{1}{4} - \zeta_{\mathsf{max}}\right)}   \sigma_{\mathsf{max}}
\end{align}
We next bound term $(c)$, $(e)$, and $(f)$. It is trivial to show that $q_{u}^{(k)}= \left(\frac{\beta \ell^{(k)}\left(\mathbb{E}_{\hat{\lambda}}\left[\widehat{\lambda}_{u}^{(k)}\right]D^2-2\mathbb{E}_{\hat{\lambda}}\left[\widehat{\lambda}_{u}^{(k)}\right]D D_{u}+ D^2_{u}\right)+\eta_k D^2}{D^2}\right) \le   2\beta \ell_{\mathsf{max}} +\eta_k $. Furthermore, assume $\max_{k,u}\left\{\mathfrak{X}_{u}^{(k)}\right\}\le \mathfrak{X}_{\mathsf{max}}$. Considering this, we get
\begin{align}
    (c) \le  \frac{ \left({4  \alpha^2 \beta^2 }\left(\ell_{\mathsf{max}}\right) \left(\ell_{\mathsf{max}}-1\right) \mathfrak{X}_{\mathsf{max}} +2|\mathcal{U}| \alpha^2 \mathfrak{X}_{\mathsf{max}}\right) \left(\zeta_{\mathsf{max}}+1\right)}{\left(\frac{1}{4} - \zeta_{\mathsf{max}}\right) K} +  \frac{4\beta \ell_{\mathsf{max}} |\mathcal{U}| \alpha \mathfrak{X}_{\mathsf{max}}\left(\zeta_{\mathsf{max}}+1\right)}{ \left(\frac{1}{4} - \zeta_{\mathsf{max}}\right) \sqrt{K}}
\end{align}
\begin{align}
    (e&)\le  \sum_{u\in \mathcal{U}}\frac{8 q_{u}^{(k)} |\mathcal{U}| \Theta^2 \beta^2 \eta_k^3 \left(\ell^{(k)}-1\right)\left(\zeta^{(k)}+1\right)}{\left(\frac{1}{4} - \zeta^{(k)}\right)} \left(1-\frac{{B}_{u}}{D_{u}} \right)  \frac{{(D_{u}-1)}\left(\sigma_{u}\right)^2}{D_{u}{B}_{u}}\nonumber\\
    &\le  \frac{ (16 \ell_{\mathsf{max}} \Theta^2 \beta^3 ( \alpha^3 |\mathcal{U}|  \left(\ell_{\mathsf{max}}-1\right)\sigma_{\mathsf{max}}\left(\zeta_{\mathsf{max}}+1\right)}{K^{3/2}\left(\frac{1}{4} -  \zeta_{\mathsf{max}}\right)} +\frac{8   \Theta^2 \beta^2 (\alpha^4 |\mathcal{U}| \left(\ell_{\mathsf{max}}-1\right)) \sigma_{\mathsf{max}}\left(\zeta_{\mathsf{max}}+1\right)}{K^2\left(\frac{1}{4} -  \zeta_{\mathsf{max}}\right)} 
\end{align}

\begin{align}
    (f)&\le \frac{\left(\zeta^{(k)}+1\right)}{\left(\frac{1}{4} - \zeta^{(k)}\right)}\sum_{u\in \mathcal{U}}2 (q_{u}^{(k)}) |\mathcal{U}|\eta_k  G^2 \nonumber\\
    &\le  \frac{ 4\beta \ell_{\mathsf{max}}|\mathcal{U}| \alpha  G^2\left(\zeta_{\mathsf{max}}+1\right)}{\sqrt{K}\left(\frac{1}{4} -  \zeta_{\mathsf{max}}\right)} +\frac{2|\mathcal{U}|  \alpha^2   G^2 \left(\zeta_{\mathsf{max}}+1\right)}{K\left(\frac{1}{4} -  \zeta_{\mathsf{max}}\right)}
\end{align}
Considering the above upper bounds, \eqref{ineqq:main79} can be further bound as follows:
\begin{equation}
\footnotesize
\begin{aligned}\label{ineqq:main76}
     \frac{1}{K} \sum_{k=0}^{K-1}\mathbb{E}_{s,\hat{\lambda}} \left[\Bigg\Vert  \nabla \mathfrak{L}(\bm{\omega}^{(k)}) \Bigg\Vert^2\right]  &\leq \frac{ \mathfrak{L}(\bm{\omega}^{(0)})-  \mathfrak{L}^{\star}}{ \sqrt{K}\sqrt{|\mathcal{U}|} \alpha\ell_{\mathsf{min}}   \left(\frac{1}{4} - \zeta_{\mathsf{max}}\right)} + \frac{{  2 \beta^2\Theta^2  \alpha^2 }\left(\ell_{\mathsf{max}}-1\right)\left(\zeta_{\mathsf{max}}+1\right)}{K\left(\frac{1}{4} - \zeta_{\mathsf{max}}\right) } \sigma_{\mathsf{max}}\\
     &+ \frac{ \left({4  \alpha^2 \beta^2 }\left(\ell_{\mathsf{max}}\right) \left(\ell_{\mathsf{max}}-1\right) \mathfrak{X}_{\mathsf{max}} +2|\mathcal{U}| \alpha^2 \mathfrak{X}_{\mathsf{max}}\right)\left(\zeta_{\mathsf{max}}+1\right)}{ K\left(\frac{1}{4} - \zeta_{\mathsf{max}}\right)} +  \frac{4\beta \ell_{\mathsf{max}} |\mathcal{U}| \alpha \mathfrak{X}_{\mathsf{max}}\left(\zeta_{\mathsf{max}}+1\right)}{\sqrt{K} \left(\frac{1}{4} - \zeta_{\mathsf{max}}\right) }\\
     &+ \frac{ (16 \ell_{\mathsf{max}} \Theta^2 \beta^3 ( \alpha^3 |\mathcal{U}|  \left(\ell_{\mathsf{max}}-1\right)\sigma_{\mathsf{max}}\left(\zeta_{\mathsf{max}}+1\right)}{K^{3/2}\left(\frac{1}{4} -  \zeta_{\mathsf{max}}\right)} +\frac{8   \Theta^2 \beta^2 (\alpha^4 |\mathcal{U}| \left(\ell_{\mathsf{max}}-1\right)) \sigma_{\mathsf{max}}\left(\zeta_{\mathsf{max}}+1\right)}{K^2\left(\frac{1}{4} -  \zeta_{\mathsf{max}}\right)} \\
     &+ \frac{ 4\beta \ell_{\mathsf{max}}|\mathcal{U}| \alpha  G^2 \left(\zeta_{\mathsf{max}}+1\right)}{\sqrt{K}\left(\frac{1}{4} -  \zeta_{\mathsf{max}}\right)} +\frac{2|\mathcal{U}|  \alpha^2   G^2 \left(\zeta_{\mathsf{max}}+1\right)}{K\left(\frac{1}{4} -  \zeta_{\mathsf{max}}\right)}+ \frac{\beta\alpha  2\Theta^2}{\sqrt{K} 2 \left(\frac{1}{4} - \zeta_{\mathsf{max}}\right)}   \sigma_{\mathsf{max}} = O\left(\frac{1}{\sqrt{K}}\right)\\
     &=O\left(\frac{1}{\sqrt{K}}\right)+ O\left(\frac{1}{{K}}\right)+O\left(\frac{1}{{K}}\right)+O\left(\frac{1}{\sqrt{K}}\right)+O\left(\frac{1}{{K}^{3/2}}\right)+O\left(\frac{1}{{K}^2}\right)+O\left(\frac{1}{\sqrt{K}}\right)+O\left(\frac{1}{{K}}\right)+O\left(\frac{1}{\sqrt{K}}\right) = O\left(\frac{1}{\sqrt{K}}\right)
\end{aligned}
\end{equation}
Moreover, taking limit from both hand sides of \eqref{ineqq:main76} when $K\rightarrow\infty$, all of the terms in the right hand side in the \eqref{ineqq:main76} become $0$. Consequently, we have 
\begin{align}
    &\lim_{K\rightarrow \infty}\frac{1}{K} \sum_{k=0}^{K-1}\mathbb{E}\left[\left\Vert\nabla{\mathfrak{L}(\bm{\omega}^{(k)})}\right\Vert^2\right]=0,
\end{align}
which guarantees the convergence of $\mathbb{X}$L and concludes the proof of Corollary~\ref{cor:main}.

% \include{Rebuttal}
% \bibliography{BIB}
\end{document}